\documentclass[lettersize,journal]{IEEEtran}
\IEEEoverridecommandlockouts
\usepackage{cite}
\usepackage{amsmath,amssymb,amsfonts,mathtools}
\usepackage[linesnumbered,noend,ruled,vlined]{algorithm2e}
\usepackage{graphicx}
\usepackage{textcomp}
\usepackage{xcolor}
\usepackage{tensor}
\usepackage{float}
\usepackage{nomencl}
\makenomenclature
\usepackage{subcaption}
\usepackage[OT1]{fontenc} 
\usepackage{multirow}
\usepackage{gensymb}
\usepackage{flushend}

\newlength{\limage}
\newlength{\rimage}
\newlength{\mimage}
\newlength{\aimage}
\newlength{\bimage}
\newlength{\cimage}
\newlength{\dimage}
\newlength{\eimage}
\newlength{\timage}
\newlength{\ttimage}

\newlength{\textfloatsepsave} 

\def\BibTeX{{\rm B\kern-.05em{\sc i\kern-.025em b}\kern-.08em
    T\kern-.1667em\lower.7ex\hbox{E}\kern-.125emX}}

\graphicspath{{./figs/}}
\DeclareGraphicsExtensions{.png,.jpg,.eps,.pdf}
\title{Omni-swarm: A Decentralized Omnidirectional Visual-Inertial-UWB State Estimation System for Aerial Swarms}

\author{Hao Xu, Yichen Zhang, Boyu Zhou, Luqi Wang, Xinjie Yao, Guotao Meng, Shaojie Shen
\thanks{
\textit{(Corresponding author: Hao Xu.)}}
\thanks{All authors are with the Department of Electronic and Computer Engineering, Hong Kong University of Science and Technology, Hong Kong, China.
{\tt\small $\{$hxubc, yzhangec, bzhouai, lxwang, gmeng, xyaoab$\}$@connect.ust.hk, eeshaojie@ust.hk}

}
}

\begin{document}
\maketitle

\begin{abstract}
Decentralized state estimation is one of the most fundamental components of autonomous aerial swarm systems in GPS-denied areas yet it still remains a highly challenging research topic. Omni-swarm, a decentralized omnidirectional visual-inertial-UWB state estimation system for aerial swarms, is proposed in this paper to address this research niche. To solve the issues of observability, complicated initialization, insufficient accuracy, and lack of global consistency, we introduce an omnidirectional perception front-end in Omni-swarm. It consists of stereo wide-FoV cameras and ultra-wideband sensors, visual-inertial odometry, multi-drone map-based localization, and visual drone tracking algorithms. The measurements from the front-end are fused with graph-based optimization in the back-end. The proposed method achieves centimeter-level relative state estimation accuracy while guaranteeing global consistency in the aerial swarm, as evidenced by the experimental results. Moreover, supported by Omni-swarm, inter-drone collision avoidance can be accomplished without any external devices, demonstrating the potential of Omni-swarm as the foundation of autonomous aerial swarms.
\end{abstract}


\section{Introduction}\label{sect:intro}
\IEEEPARstart{F}{or} any aerial robotics system, the estimation of states, including positions and attitudes, is crucial. 
The estimation system lays a solid foundations for higher-level functions, such as path planning \cite{zhou2019robust} and mapping \cite{wang2019surfel}. 
The state estimation problems for single-drone systems are currently well-addressed through approaches such as visual-inertial odometry (VIO) \cite{qin2018vins,murORB2,Geneva2020ICRA} and LiDAR odometry\cite{zhang2014loam,lin2020loam}.
However, when we look beyond a single-drone to multiple drones working as an aerial swarm, the problem becomes much more complicated.
In a swarm, each drone needs to estimate its ego state and also obtain the relative poses of other drones.

To date, the vast majority of aerial swarm researchers have adopted external devices, such as motion capture systems \cite{preiss2017crazyswarm}, ultra-wideband (UWB) systems with anchors\cite{ledergerber2015robot} and GPS \cite{jaimes2008approach}, and RTK-GPS\cite{moon2016outdoor} systems, to provide state estimations, which  significantly limits the application of aerial swarms in the real world. 
Although motion capture systems and UWB modules with anchors can work in indoor environments with decent accuracy, they are centralized systems and require bulky external devices, meaning that they are susceptible to losing the central devices and are challenging to deploy.
A requirement of a practical swarm application is simple deployment, which such approaches typically fail to meet.

\begin{figure}[t]
    \centering
    \includegraphics[width=0.8\linewidth]{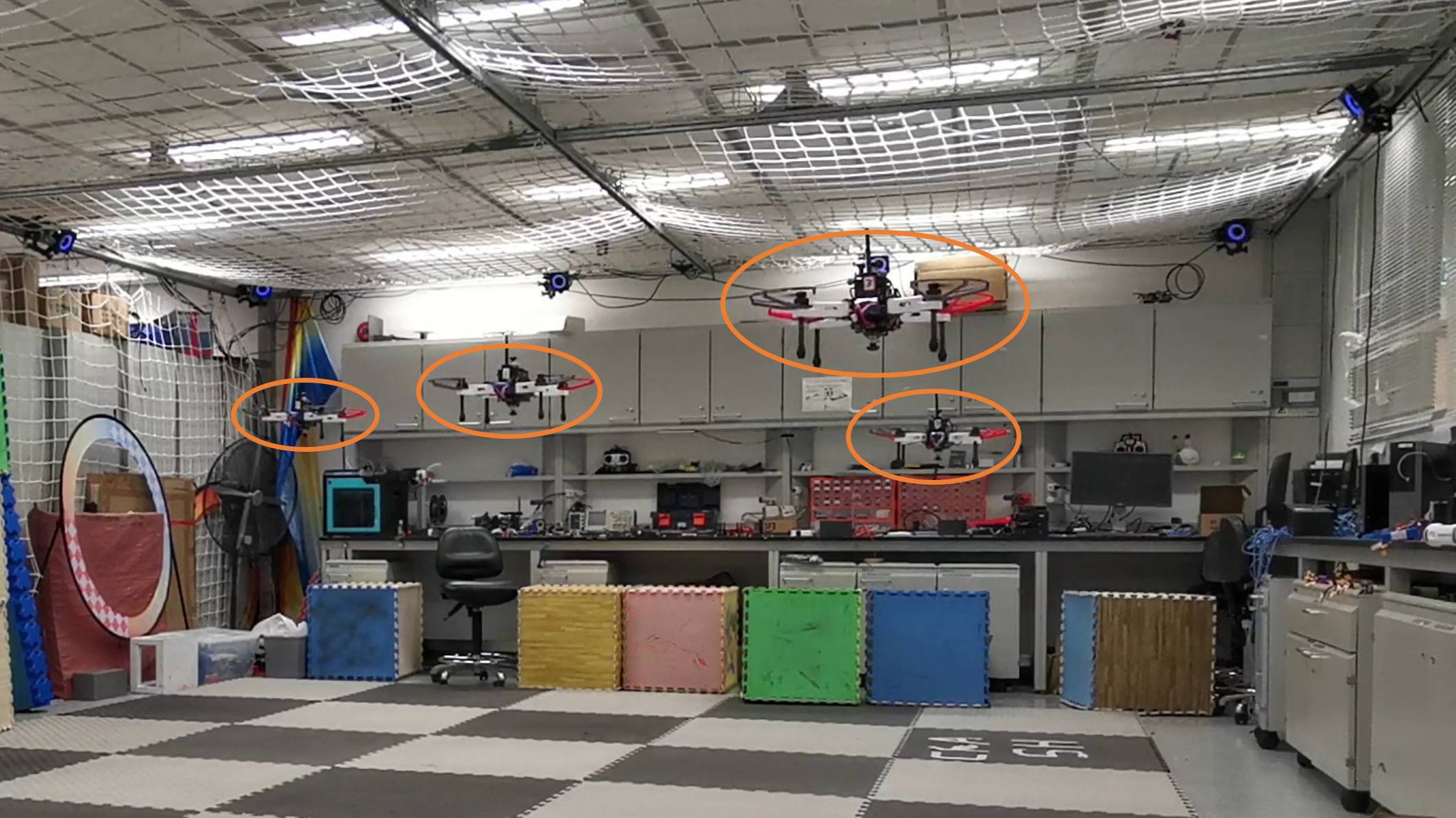}
\caption{
    \small{Indoor aerial swarm formation flight with four drones. The customized
    drone platforms are circled.}}\label{fig:SWARM}
    \vspace{-0.6cm}
\end{figure}

The decentralized scheme of swarm robots is becoming popular in swarm robotics research \cite{petravcek2021large, smith2018distributed, petravcek2021large, zhu2021cooperative, zhu2021distributed, choudhary2017distributed, lajoie2020door} because of its significant advantages.
A robot swarm with this scheme does not require all robots to have stable communication with a central computer, which makes it more flexible in real-world environments where communication is limited.
Additionally, each robot can act largely independently from the remainder of the team, making the whole system more fault-tolerant to  single-point failure.
To build a fully autonomous decentralized multi-robot system (multi-aerial-robot swarm, also known as an aerial swarm), a key problem is how to achieve relative state estimation in a decentralized fashion.
The primary motivation of this paper is to solve this fundamental problem with additional global consistency of the estimated states, laying a solid foundation for a decentralized aerial swarm.

Recently, researchers have started to develop  approaches to perform decentralized relative state estimation on aerial swarms.
One of the most straightforward ideas is to utilize visual object detection to detect the drones in the aerial swarm to estimate the relative state \cite{xu2020decentralized,guo2017ultra,guo2019ultra, ziegler2021distributed}.
Fusing distance measurements from the UWB and odometry (usually VIO) \cite{xu2020decentralized,walter2019uvdar,nguyen2019vision,guo2017ultra,guo2019ultra,ziegler2021distributed} is another viable approach, along with estimating relative states from common environment features captured by the drones \cite{piasco2016collaborative,achtelik2011collaborative} or extracting relative states from sparse maps built by the aerial swarms \cite{cunningham2010ddf,cunningham2013ddf,choudhary2017distributed,lajoie2020door}.

However,  the practicability of these methods is limited by some serious issues, as follows,
\begin{enumerate}
    \item Observability issue caused by a restricted field of view (FoV).
    For visual-drone-detection-based methods\cite{walter2019uvdar,nguyen2019vision,xu2020decentralized}, the relative states are observable only when others drones are in the drone's FoV. 
    \item Complicated initialization. UWB-odometry fusion methods \cite{guo2017ultra,guo2019ultra,xu2020decentralized, ziegler2021distributed} require large motions to initialize the system. The complex initialization procedure may cause severe safety issues and even crashes. 
    \item Insufficient accuracy. The estimated position errors in previous works  \cite{guo2017ultra,guo2019ultra, ziegler2021distributed} are generally around 20 cm--50 cm, meaning that these swarm systems can scarcely be adopted for use in confined indoor spaces or close formation scenarios. 
    \item Lack of global consistency. For all the current relative state estimation methods, the estimated poses drift and the ego state estimation cause consistency issues.
    Global consistency becomes especially important when we expect to build global maps based on state estimation.
\end{enumerate}

To address the challenges, in this paper, we extend our previous method\cite{xu2020decentralized} and propose \textbf{Omni-swarm: a decentralized omnidirectional visual-inertial-UWB state estimation system for aerial swarms}, which combines the advantages of  the UWB-odometry fusion method\cite{xu2020decentralized,walter2019uvdar,nguyen2019vision,guo2017ultra,guo2019ultra,ziegler2021distributed}, visual-object-detection-based methods\cite{walter2019uvdar,nguyen2019vision,xu2020decentralized}, and map-based methods \cite{piasco2016collaborative,achtelik2011collaborative, luqi2018collaborative, michael2014collaborative,choudhary2017distributed,lajoie2020door}.

The most important contribution to address the aforementioned issues is the introduction of an omnidirectional perception front-end.
The front-end includes the hardware capturing omnidirectional visual information and the algorithms processing this information. 
Specifically, we use two fisheye cameras with a wide-FoV (up to $235^\circ$) to achieve omnidirectional observation of the surrounding area.
We develop VINS-Fisheye as the ego-motion estimator of the front-end, which is a VIO state estimator using the measurements from
stereo wide-FoV fisheye cameras for ego-motion estimation.
Multi-drone map-based localization (MDML) based on real-time-generated sparse maps is further introduced to ensure the global consistency of state estimation and to achieve fast initialization.
Finally, a visual drone tracking module detects and tracks the other drones to provide accurate relative pose estimation of the tracked targets.
In addition to omnidirectional visual information, we also use UWB sensors to measure the relative distance between drones. This measurement can also be considered as omnidirectional.

As the back-end of the state estimation, we adopt the graph-based optimization method, which fuses the measurements from the front-end in real-time to estimate the states of the swarm with high accuracy.
In the back-end, we adopt the state-of-the-art outlier rejection method to reduce the errors from the front-end to achieve high-accuracy and robustness.

Omni-swarm is designed to be decentralized, it runs on each drone's onboard computer individually instead of using a central server.
Differing from previous work on relative-state estimation, which only work in a line-of-sight situation\cite{xu2020decentralized,walter2019uvdar,guo2017ultra,guo2019ultra, ziegler2021distributed}, Omni-swarm can estimate the state of the swarm when another drone is in non-line-of-sight if the same place has been visited. 
This capability is given by MDML.
Another advantage  over previous works \cite{xu2020decentralized,walter2019uvdar,guo2017ultra,guo2019ultra, ziegler2021distributed} that MDML brings the global consistency; the long-term drifting of the ego state is eliminated by the map-based localization and thus guarantee global consistency of the estimation results.
Omni-swarm also inherits the high accuracy of our previous work\cite{xu2020decentralized}, which has a high relative localization accuracy compared to other related works\cite{walter2019uvdar,guo2017ultra,guo2019ultra, ziegler2021distributed}.

In addition, the complex initialization problem is solved in Omni-swarm by introducing various system initialization methods, including map-based initialization and visual drone detection tracking initialization.
Another Omni-swarm improvement that helps with practical applications is its plug-and-play feature,
based on our newly introduced initialization methods, Omni-swarm allows the temporary joining or exiting of drones.
Finally, Omni-swarm has redundancy in the information shared among the swarm and computations performed by the drones, which brings robustness to possible temporary signal loss and partial sensor failures. 

To verify the above features of Omni-swarm, we perform comprehensive experiments in simulation and real-world experiments.
Moreover, we design an inter-drone collision avoidance experiment to verify Omni-swarm under realistic conditions.
The main contributions of this paper are:
\begin{enumerate}
    \item Omni-swarm, a decentralized omnidirectional visual-inertial-UWB swarm state estimation system. 
    Extensive experiments are conducted to validate Omni-swarm.
    \item Open-source releases of the software and the custom datasets have been made public\footnote{https://github.com/HKUST-Aerial-Robotics/Omni-swarm}.
\end{enumerate}

In our previous work \cite{xu2020decentralized}, we proposed a two-stage visual-inertial-UWB fusion method for relative state estimation.
The method has the advantages of both visual-object-detection-based relative state estimation\cite{walter2019uvdar,nguyen2019vision} and UWB-odometry fusion relative state estimation \cite{guo2017ultra,guo2019ultra}.
However, global consistency is absent. 
Also, the method still suffers from the same complicated initialization issue and the observability issue caused by the restricted FoV, as with other related methods. 
Although the previous method works in a non-line-of-sight case, the relative estimation accuracy will deteriorate. 
In extreme situations, the relative state can become unobservable. 
For example, drones fly parallelly side-by-side.
In this paper, the initialization and global consistency issues are addressed by introducing multi-drone map-based localization, and the observability is fixed by using omnidirectional cameras.

In this paper, related works are discussed in Sect. \ref{sect:related}. Omni-swarm is briefly introduced in Sect. \ref{sect:overview}, and a clear definition of the state estimation problem is defined in Sect.\ref{subsec:def}. 
The front-end of Omni-swarm is presented in Sect. \ref{sect:front-end}, while the back-end is introduced in Sect. \ref{sect:back-end}. 
Experimental results are discussed in Sect. \ref{sect:experiment}. 
Finally, we conclude the paper in Sect. \ref{sect:con} and introduce potential future works. 

\section{Related Works}\label{sect:related}
\subsection{Visual inertial odometry on Aerial Swarm}

In order to overcome the state estimation issues in multiple environments, including GPS-denied areas, visual-inertial simultaneous localization and mapping (visual-inertial SLAM) \cite{campos2021orb, mur2017visual, usenko2019visual} and visual-inertial odometry (VIO) \cite{qin2018vins,Geneva2020ICRA, qin2019a, bloesch2015robust} are widely adopted on single drone systems.
These visual inertial systems fuse the visual images and the inertial measurement unit (IMU) data together to estimate ego-motion without requiring external devices. 
Although VIO systems can be effortlessly adopted in various environments without cumbersome external devices and they are directly utilized by researchers on swarms for formation flights \cite{weinstein2018visual, lusk2020distributed}, the state estimations suffer from severe drift, which can be detrimental to multi-drone systems. 
Specifically, the different drifts of individual drones in an aerial swarm induce distinct position estimations at the same location, possibly causing fatal crashes in collaborative missions in the absence of other correction methods. 

\subsection{Relative Swarm State Estimation}

One of the most effective methods to compensate for the drift of VIO in aerial swarms is to incorporate the relative state estimation between the drones. 
The existing relative state estimation methods can generally be divided into the following three categories:
1) UWB-odometry fusion relative state estimation methods \cite{guo2017ultra,guo2019ultra,xu2020decentralized, nguyen2019persistently, ziegler2021distributed},
2) visual-object-detection-based relative state estimation methods \cite{walter2019uvdar,nguyen2019vision,xu2020decentralized}, and
3) environment-feature-based relative state estimation methods \cite{piasco2016collaborative,achtelik2011collaborative}.

However, all of these methods have drawbacks in real application scenarios.
The previous UWB-odometry based methods\cite{guo2017ultra,guo2019ultra, ziegler2021distributed} fuse the ego-motion estimated by VIO and UWB distance measurements to achieve relative state estimation.
These methods can only estimate relative  localization in line-of-sight situations and provide merely a meter to decimeter level of accuracy.
Even in line-of-sight situations, the relative localization can become unobservable in some cases, e.g., parallel flight tasks \cite{nguyen2019distance}.
Furthermore, the initialization requires a certain amount of motion, usually several meters.
Both of these drawbacks substantially limit the practicability of the UWB-visual fusion methods in feature-rich narrow environments.
A viable idea  to address observability issue was proposed by Nguyen et al. \cite{nguyen2019distance}. 
Briefly, a drone in the swarm keeps an irregular motion to guarantee observability of the relative state estimation, which may present safety issues and limit the aerial swarm's cooperation.
In Omni-swarm, the observability issue can be solved by visual drone tracking and map-based localization while the formation is flying.
In addition, by introducing map-based localization, we can achieve fast initialization without motion.

Visual-object-detection-based methods  \cite{walter2019uvdar,nguyen2019vision,xu2020decentralized} are capable of delivering centimeter-level accuracy. 
However, the accuracy of these methods in a swarm of drones is highly dependent on the distance the drones are from each other.
Data association from the detection results to the corresponding drone ID is also an issue when all the drones appear identical.
A coupled-probabilistic-data-association-filter (CPDAF)-based approach is proposed in \cite{nguyen2019vision} to associate the detection result and estimate the relative state.
Nevertheless, the visibility requirement between the drones and the limited FoV of their cameras restrict the swarm formation and the distance between drones for stable state estimation.
In this paper, the FoV issue is addressed by introducing an omnidirectional front-end.
Due to the UWB measurements, the proposed method can still estimate the relative state when the  distances between the drones are too far for them to detect each other. 

Finally, all the aforementioned methods are limited to relative localization, and global consistency of the estimated states  cannot be guaranteed. 
In this paper, global consistency is guaranteed by introducing map-based localization. 
This will also help us with global mapping in our future work.

\subsection{Environmental-Features-based Method and Collaborative Simultaneous Localization and Mapping}

Environment-feature-based methods \cite{piasco2016collaborative,achtelik2011collaborative} rely on multiple drones' common environmental texture features to estimate the relative poses. 
The methods require sufficient overlapped features between the view of the drones, limiting the swarm formation and heading, while only working in feature-rich environments.

Collaborative simultaneous localization and mapping (CSLAM) methods \cite{luqi2018collaborative, michael2014collaborative, cunningham2010ddf,cunningham2013ddf,choudhary2017distributed,lajoie2020door, zhu2021distributed, zhu2021cooperative} focus on sparse and dense mapping utilizing the sensors on the multi-robot system, and can estimate the states of aerial swarms.
DDF-SAM \cite{cunningham2010ddf,cunningham2013ddf} presents a landmark-based back-end implementation without providing the front-end and data association between features among the swarm. 
Choudhary et al. \cite{choudhary2017distributed} utilized objects as landmarks, requiring known objects and their 3D models in the scene, which limits the method's real-world practicability.
DOOR-SLAM \cite{lajoie2020door,luqi2018collaborative, michael2014collaborative} achieves localization of robots using VIO incorporating loop closure detection (also known as map-based localization in this paper).
\cite{luqi2018collaborative, michael2014collaborative, lajoie2020door, choudhary2017distributed} estimate the state with the pose graph optimization (PGO).
However, the accuracy of relative localization is crucial for cooperation inside the aerial swarm, and this has not been featured in these papers.
Generally speaking, the PGO methods are not accurate because they do not directly use the features for relative localization.
Zhu et. al. \cite{zhu2021distributed} propose a distributed visual-inertial fusion for cooperative localization and reach centimeter-level accuracy. 
In \cite{zhu2021distributed} the environmental features are tightly coupled in the cooperative localization.

Due to the nature of map-based localization, the above CSLAM methods are limited to feature-rich environments and are also limited by the camera FoV.
In this paper, with an omnidirectional front-end and UWB measurements, Omni-swarm will not be limited by the FoV and can still estimate relative state without rich common environmental features.

\begin{figure*}[ht!]
    \centering
    \vspace{-0.5cm}
    \includegraphics[width=0.7\linewidth]{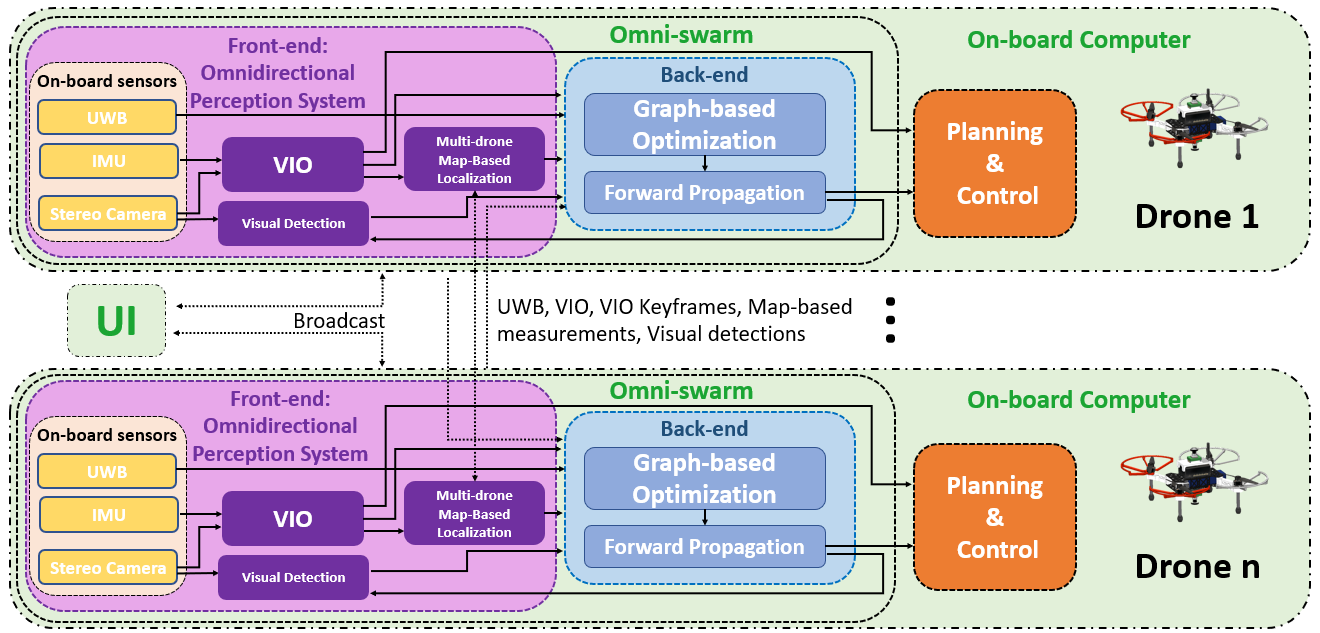}
    \caption{\small{The system architecture of Omni-swarm. The data from the onboard sensors are processed, and then broadcast to all the other drones. 
    The swarm state estimation on each of the onboard computers collects both onboard and broadcast information, including the relative distance from the UWB modules, the VIO, map-based measurements, and the detection results, and performs optimization prediction to obtain real-time relative state estimations. 
    The estimation results are then sent back to facilitate the matching procedure of detection and tracking and meanwhile serve the planning and control. The ground station obtains the drones' real-time information to monitor the flight status and concurrently sends the commands to the drones. All the communications between the devices are through UWB broadcast.}}
    \vspace{-0.6cm}
    \label{fig:structure}
\end{figure*}

\subsection{Visual-inertial-UWB Fusion with Global Consistency}
Some of the works mentioned as being UWB-odometry fusion \cite{guo2019ultra, xu2020decentralized, ziegler2021distributed} can be considered to be visual-inertial-UWB fusion (also known as visual-inertial-range fusion, or VIR fusion). 
However, global consistency is absent in these works.
Unlike these relative localization methods, a class of works  \cite{guo2019ultra, xu2020decentralized, ziegler2021distributed} uses VIR fusion to achieve global localization by placing a fixed UWB anchor in the environment.
With the help of this anchor, these methods can effectively eliminate the drift of odometry.
The drawback of these methods is that they require additional infrastructure to be placed on the ground and they maintain global consistency only when drones are in the line of the sight of the anchors, limiting their real-world practical value.
In this paper, the proposed Omni-swarm guarantees global consistency without any external infrastructure, which is more flexible in real-world applications.

\section{SYSTEM OVERVIEW}\label{sect:overview}
\subsection{Notation}\label{notation}
To aid understanding of the proposed system, the notations are defined below.

\nomenclature[1]{$\hat{(\cdot)}$}{The estimated state.}

\nomenclature[2]{$\mathbf{z}_{(\cdot)}^t $}{The measurement data at time t.}

\nomenclature[3]{$\tensor*[^{b_k}]{(\cdot)}{^t_i}$}{State of drone $i$ in drone \textit{k}'s body frame. For simplicity, the pose in the body frame is defined as a 4-DoF pose, i.e., $\tensor*[^{b_k}]{(\cdot)}{^t_i}= (\tensor*[^{v_k}]{\mathbf{P}}{_{k}^t})^{-1} \tensor*[^{v_k}]{(\cdot)}{^t_i}$.}

\nomenclature[4]{$\tensor*[^{v_k}]{(\cdot)}{^t_i}$}{State of drone $i$ in drone $i$'s local frame.}

\nomenclature[5]{$\tensor*[^{v_k}]{\mathbf{P}}{_{i}^t}$}{Equal to $\begin{bmatrix}
\mathbf{R}_z( \tensor*[^{v_k}]{\psi}{^t_{i}}) & \tensor*[^{v_k}]{\mathbf{X}}{_{i}^t} \\
0  &  1
\end{bmatrix} $. The pose of drone $i$ in drone \textit{k}'s local frame at time $t$. For simplicity, the notation of $ \tensor*[]{\mathbf{P}}{_k^t}$ represents $\tensor*[^{v_k}]{\mathbf{P}}{_k^t}$. $\tensor*[^{v_k}]{\mathbf{R}}{_z}(\tensor*[^{v_k}]{\psi}{^t_i})$ represents the rotation matrix rotated over the z axis with angle $\tensor*[^{v_k}]{\psi}{^t_{i}} = (\tensor*[^{v_k}]{\mathbf{R}}{^t_i})_\psi$.}

\nomenclature[6]{$\tensor*[^{v_k}]{\mathbf{T}}{_{i}^t}$}{Equal to $\begin{bmatrix}
    \tensor*[^{v_k}]{\mathbf{R}}{^t_i} & \tensor*[^{v_k}]{\mathbf{X}}{_{i}^t} \\
    0  &  1
    \end{bmatrix} $. The 6-DoF pose of drone $i$ in drone \textit{k}'s local frame at time $t$. $\tensor*[^{v_k}]{\mathbf{R}}{^t_i}$ represents the rotation matrix}
    
\nomenclature[7]{$\tensor*[]{\mathbf{\tilde{P}}}{_k^t}, \tensor*[]{\mathbf{\tilde{T}}}{_k^t}$}{The 4-DoF and 6-DoF pose, respectively, of  drone \textit{k} in its local frame, as estimated by VIO, which drifts through time. This pose is initialized to an identity matrix after the drone start-up, which also gives the origin and the axis of the local frame.}

\nomenclature[8]{$\tensor*[^{v_k}]{\mathbf{X}}{_{i}^t}$}{Equal to $\left[\tensor*[^{v_k}]{x}{_i^t}, \tensor*[^{v_k}]{y}{_i^t}, \tensor*[^{v_k}]{z}{_i^t}\right]^T$. The translation part of $\tensor*[^{v_k}]{\mathbf{P}}{_{i}^t}$.}

\nomenclature[9]{$\tensor*[]{\mathbf{\delta P}}{_i^t}$}{The transformation matrix from time $t-1$ to $t$ of drone $i$ from the VIO result, i.e., $\tensor*[]{\mathbf{P}}{_i^t} = \tensor*[]{\mathbf{P}}{_i^{t-1}}\tensor*[]{\mathbf{\delta P}}{_i^t} $.}

\nomenclature[10]{$d_{i,j}^t $}{Distance between drone $i$ and drone $j$ at time $t$.}

% \nomenclature[11]{$ {\mathbf z_D}_{k\rightarrow j}^{t_0}$}{The relative pose measurement of the detected drone $i$ in the body frame for drone $k$.}

\nomenclature[12]{$\mathcal{F}_k^t$}{The keyframe of the drone $k$ at time $t$, which contains the 4D pose $\tensor*[^{v_k}]{\mathbf{\hat P}}{_{i}^t}$ to be estimated, a few virtual camera keyframes $\tensor*[^c]{\mathcal{K}}{_k^t}$ and other essential information of the drone.}

\nomenclature[13]{$\tensor*[^c]{\mathcal{K}}{_k^t}$}{The keyframe of drone $k$'s virtual camera $c$ at time $t$, which contains the global descriptor, local features, virtual camera's extrinsic and other essential information of the drone. The virtual camera $c$ is cropped from the raw fisheye camera.}

\nomenclature[14]{$\mathcal{SF}_k^t$}{Equal to $\left[\mathcal{F}_1^t \ \mathcal{F}_2^t \ ...\  \mathcal{F}_n^t \right].$ The swarm keyframe of the drone $k$ at time $t$, which contains $n$ keyframes.}

\nomenclature[15]{$\mathcal{G}_k$}{The graph built on drone $k$ for state estimation.}

\nomenclature[16]{$(\cdot)_R$}{The rotation part of the transformation matrix.}

\nomenclature[17]{$(\cdot)_P$}{The corresponding 4-DoF transformation matrix.}

\nomenclature[18]{$(\cdot)_T$}{The corresponding 6-DoF transformation matrix.}

\nomenclature[19]{$(\cdot)_X$}{The translation part of the transformation matrix.}

\nomenclature[20]{$(\cdot)_\psi$}{The yaw angle of the rotation matrix.}
\nomenclature[21]{$(\mathcal{F})_f$}{The global descriptor of the keyframe $\mathcal{F}$.}
\nomenclature[22]{$(\mathcal{K})_\mathcal{F}$}{The corresponding keyframe $\mathcal{F}_k^t$ of virtual camera keyframe $\mathcal{K}$.}
\nomenclature[23]{$(\mathcal{F})_{lf}$}{The local descriptors of the features of the keyframe $\mathcal{F}$.}
\nomenclature[24]{$\left\Vert(\cdot)\right\Vert$}{Euclidean norm of $(\cdot)$ if $(\cdot)$ is a vector or matrix; otherwise if $\cdot$ is a set, $\left\Vert(\cdot)\right\Vert$ is its size.}
\nomenclature[25]{$\left\Vert(\cdot)\right\Vert_\Sigma$}{Mahalanobis norm of $\cdot$.}
\nomenclature[26]{$\mathcal{D}$}{The set of all existing drones, including the currently unavailable drones due to loss of communication, user poweroff and accident.}
\nomenclature[27]{$\mathcal{D}^k_a$}{The set of all available drones for drone $k$.}
\nomenclature[28]{$\mathcal{D}^k_e$}{The set of all estimated drones of drone $k$'s state estimation.}
\nomenclature[29]{$\mathcal{D}^k_u$}{The set of all uninitialized drones of drone $k$'s state estimation, where $\mathcal{D}^k_u = \mathcal{D}^k_a - \mathcal{D}^k_e$.}
\nomenclature[30]{$D_i$}{The $i$ th drone.}

\printnomenclature

Suppose our aerial swarm contains up to $n$ drones.
A drone with ID $i$, will be denoted as drone $i$ or $D_i$, where $i\in \mathcal{D},\ \ \mathcal{D}=\{1, 2, 3, ... n\}$.
Although Omni-swarm runs independently on each drone, to simplify the discussion, we discuss drone $k$'s Omni-swarm by default in the following unless otherwise stated.
We say a drone $i$ is available to drone $k$ when it satisfies following conditions: 1) drone $i$'s Omni-swarm and ego-motion state estimation (VIO) work properly; 2) drone $i$ and drone $k$ have a stable network connection.
The set of all available drones for drone $k$ is denoted as $\mathcal{D}^k_a$, which also contains the drone itself.

\subsection{State Estimation Problem of Aerial Swarm}\label{subsec:def}
For an aerial swarm that contains a maximum of $n$ homogeneous drones, the state estimation problem can be represented as follows.
For every drone $k\in \mathcal{D}$, estimate the 6-DOF pose $\tensor*[^{v_k}]{\mathbf{T}}{_i^t}$ for every drone $i\in \mathcal{D}^k_a$ at time $t$ in drone $k$'s local frame.
The state estimation problem for drone $k$ can be split into two parts:
\begin{enumerate}
    \item Estimating the ego-motion state of drone $k$ in a local frame, i.e., $\tensor*[]{\mathbf{\hat T}}{_k^t}$.
    \item Estimating the state of any other arbitrary drone $i$, i.e., $\tensor*[^{v_k}]{\mathbf{\hat T}}{_i^t}$, and 4-DoF relative state $\tensor*[^{b_k}]{\mathbf{\hat P}}{_i^t}$.
\end{enumerate}

The VIO utilizes the gravitational acceleration measured by the IMU to help extract the roll and pitch angles in the attitude.
Because the gravity acceleration is consistent among drones, with the estimation of the 4-DoF pose $\tensor*[^{v_k}]{\mathbf{\hat P}}{_i^t}$ and relative pose $\tensor*[^{b_k}]{\mathbf{\hat P}}{_i^t}$, we are able to combine the drone's own VIO $\tensor*[]{\mathbf{\tilde T}}{_i^t}$  to obtain the 6-DoF pose:
\begin{equation}\label{eq:4to6}
    \tensor*[^{v_k}]{\mathbf{\hat T}}{_i^t} = \begin{bmatrix}
        \mathbf{R}_z\left((\tensor*[^{v_k}]{\mathbf{\hat P}}{_i^t})_\psi-(\tensor*[]{\mathbf{\tilde T}}{_i^t})_\psi\right)\tensor*[]{\mathbf{\tilde R}}{_i^t} && (\tensor*[^{v_k}]{\mathbf{\hat P}}{_i^t})_X \\
        0 && 1
    \end{bmatrix},
\end{equation}
where $\mathbf{R}_z\left((\tensor*[^{v_k}]{\mathbf{\hat P}}{_i^t})_\psi-(\tensor*[]{\mathbf{\tilde T}}{_i^t})_\psi\right)$ eliminates the yaw drift of the rotation  $\tensor*[]{\mathbf{\tilde R}}{_i^t}$ estimated by the VIO, where $\mathbf{R}_z(\psi)$ is the rotation matrix rotated over the z axis with angle $\psi$.

\subsection{Global Consistency of the State Estimation}\label{subsec:consis}
In SLAM research\cite{qin2018relocalization, qiu2017model}, global consistency of state estimation represents that the estimate results are drift-free, e.g., the $\tensor*[^{v_k}]{\mathbf{\hat P}}{_i^t}$ do not drift along with the robot move.

In our previous work \cite{xu2020decentralized}, we focused on estimating  $\tensor*[^{b_k}]{\mathbf{\hat P}}{_i^t}$, which is the relative state estimation for the aerial swarm.
The ego-motion was directly estimated by the VIO; i.e., $\tensor*[]{\mathbf{\hat T}}{_k^t} = \tensor*[]{\mathbf{\tilde T}}{_k^t}$ was assumed. 
However, the VIO is concerned with local accuracy and suffers from long-term drifting.
In this paper, by adopting the map-based localization method, we can estimate the state of the aerial swarm with guaranteed global consistency,
which means both $\tensor*[^{v_k}]{\mathbf{\hat T}}{_k^t}$ and $\tensor*[^{b_k}]{\mathbf{\hat P}}{_i^t}$ are estimated by the proposed state estimator for the aerial swarm.

\subsection{Observability and Initialization}
For all state estimation systems, initialization is critical, especially for Omni-swarm.
The necessary condition for initializing the state estimation is that observability must be satisfied.
Initialization with improper states and insufficient observability are likely to direct Omni-swarm to incorrect state estimation.

Compared to previous works, a major highlight of the system proposed in this paper is the extended observability, which is important for its practical application.
First, ego-motion estimation, namely the VIO in our system, is always assumed to be available for drones in an aerial swarm since it is the fundamental module for stable flight.
Therefore, the Omni-swarm observability problem is focused on the relative pose.
The observability of drone $i$'s states estimated by drone $k$ can be classified into two levels: 1) 3-DoF observable: The position $\tensor*[^{v_k}]{\mathbf{X}}{_{i}^t}$ is observable, 
and 2) 6-DoF observable: The 6-DoF pose $\tensor*[^{v_k}]{\mathbf{T}}{_{i}^t}$. When the 4-DoF pose $\tensor*[^{v_k}]{\mathbf{P}}{_{i}^t}$ and VIO $\tensor*[]{\mathbf{\tilde T}}{_i^t}$ are observable, $\tensor*[^{v_k}]{\mathbf{T}}{_{i}^t}$ will also be observable, following Eq. (\ref{eq:4to6}).

However, the observability of each drone in a swarm running Omni-swarm may be different. 
Thus, to provide the ability of plug-and-play, we track the observability of each available drone in the swarm and only initialize and estimate the state of observable drones.
We use set $\mathcal{D}_u^k$ to describe all the uninitialized drones and the set $\mathcal{D}_e^k$ to describe all the initialized (estimated) drones.
The details of the observability and initialization of Omni-swarm will be stated in Sect. \ref{sect:obser} and Sect. \ref{sect:init}.

\subsection{System Architecture}
As shown in Fig. \ref{fig:structure}, our proposed method is divided into a front-end and a back-end, and Omni-swarm independently runs on each drone in the swarm.
In the front-end, the raw measurements are pre-processed by an ego-motion estimator (VIO), visual drone tracking module (VDT), and multi-drone map-based localization module (MDML). 
In the back-end, graph-based optimization is utilized for state estimation. 
After this, we propagate the real-time state of the aerial swarm with the latest VIO and the estimated states.

\subsection{Graph-based Optimization}

\begin{figure*}[ht]
    \centering
    \settowidth\timage{\includegraphics[height=6cm]{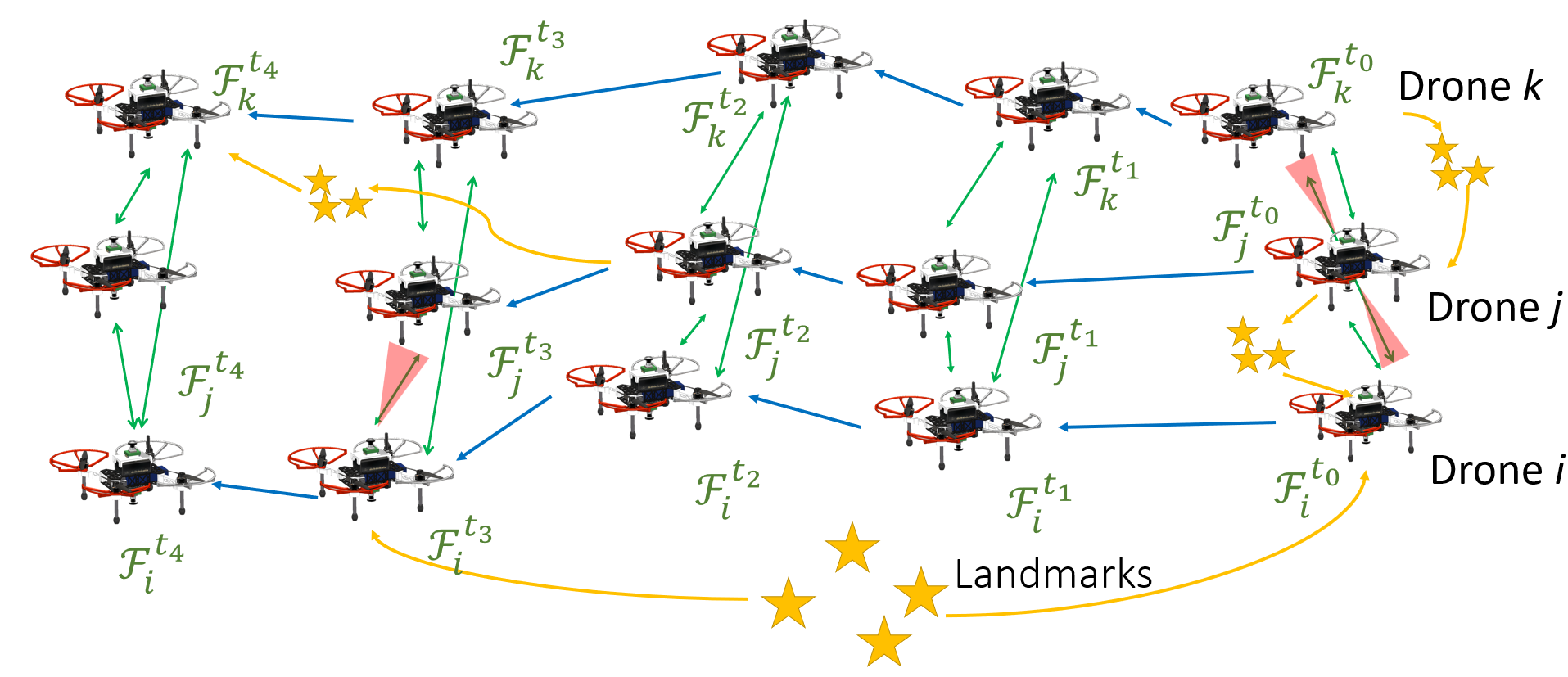}}
    \settowidth\ttimage{\includegraphics[height=6cm]{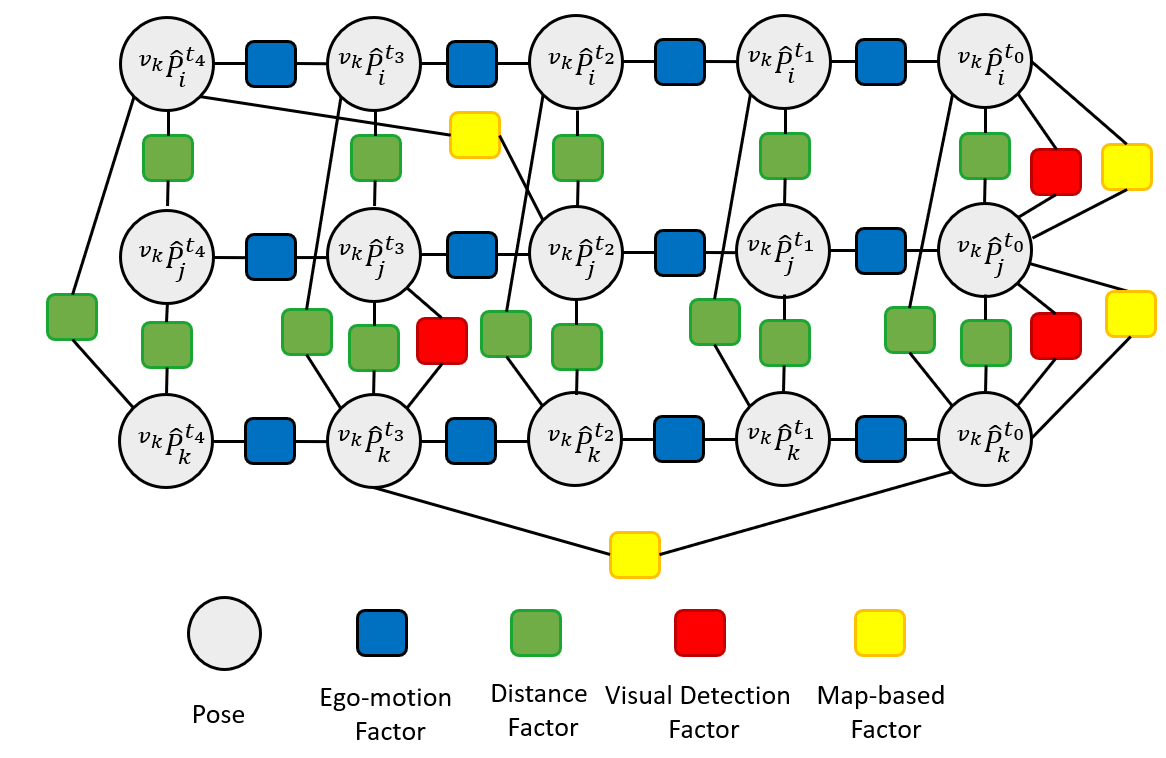}}
    \resizebox{0.9\textwidth}{!}{
        \begin{tabular}{p{\timage}p{\ttimage}}
    \includegraphics[height=6cm]{measurements_5.PNG}\newline
    \vspace{-0.5cm}
    \subcaption{}\label{fig:measurements}
        &   \includegraphics[height=6cm]{factor_graph.png}\newline
        \vspace{-0.5cm}
        \subcaption{}\label{fig:factor_graph}
    \end{tabular}
    }
    \vspace{-0.5cm}
    \caption{\small{
        A demonstration of measurements and corresponding factor graph in the graph-based optimization.
        a) Measurements involved in the graph-based optimization, including the UWB measurements (green), the VIO measurements (blue arrow) and the map-based factor  (yellow).
        The yellow stars represent the landmarks of the sparse map, and the map-based factors are detected from these landmarks.
        b) Factor graph for swarm state estimation. 
        The poses are connected by ego-motion factors, distance factors, visual detection factors and map-based factors.
        }}\label{fig:graph}
    \vspace{-0.6cm}
\end{figure*}

We adopt a 4-DoF graph-based optimization for the swarm state estimation in the back-end of Omni-swarm.
Suppose $\mathcal{SF}^{t}$ denotes the swarm keyframe, which contains the keyframes $\left\{ \mathcal{F}_1^t \ ... \ \mathcal{F}_n^t \right\}$ of $n$ drones at time $t$.

The graph $\mathcal{G}$ contains $m$ swarm keyframes, the keyframes $\left\{ \mathcal{F}_1^t \ ... \ \mathcal{F}_n^{t_m} \right\}$ of these swarm keyframes serve as the vertices of $\mathcal{G}$ and the measurements serve as edges to connect these keyframes:
$$
\begin{aligned}
    \mathcal{G}=\{ \mathcal{SF}^{1}, \ \mathcal{SF}^{2}, & \ \mathcal{SF}^{3} ... \ \mathcal{SF}^{m}, \\
    &  ...\ {\mathbf{z}_\mathcal{L}}_{i\rightarrow j}^{t_0 \rightarrow t_1} ...\ {\mathbf{z}_D}_{k\rightarrow l}^{t_2}... \ {{z}_d}_{p,q}^{t_3} \}.
\end{aligned}
$$
An illustration of $\mathcal{G}$ can be found in Fig. \ref{fig:measurements}.

In our graph-based optimization, we use maximum a posteriori (MAP) inference for the factor graph based on non-linear least-squares optimization\cite{dellaert2017factor} for swarm state estimation.
A factor graph $\mathcal{G}_f$ is constructed from graph $\mathcal{G}$ in our back-end after the outlier rejection module filters out the outlier measurements in the graph $\mathcal{G}$.
As shown in Fig. \ref{fig:factor_graph}, the poses of the vertices in $\mathcal{G}$ serve as the variables in $\mathcal{G}_f$, and the variables are connected by four types of factors, which are built from the measurements of $\mathcal{G}$:
\begin{itemize}
    \item Ego-motion factor: Each pose is connected to the previous pose of the same drone by an ego-motion factor, which represents the 4-DoF relative pose  $\mathbf{z}_{\mathbf{\delta P}_{j}}^t$ from the previous keyframe.
    This factor smoothes the state estimation and provides local accuracy.
    \item Map-based factor: The poses are connected by the corresponding map-based factors, which
    represent the 4-DoF relative poses  ${\mathbf{z}_\mathcal{L}}_{i\rightarrow j}^{t_0 \rightarrow t_1}$ from the keyframe $\mathcal{F}_i^{t_0}$ to $\mathcal{F}_j^{t_1}$ to ensure global consistency and provide relative state estimation. 
    \item Distance factor: The poses at one timestamp are connected to each other by distance factors, which are the distance measurement $z^t_{d_{i,j}}$ measured by UWB for relative state estimation. 
    \item Visual detection factor:  The poses are connected by a visual detection factor ${\mathbf z_D}_{k\rightarrow j}^{t_0}$ if one drone successfully detects another drone in the visual drone tracking module. 
    This factor stands for accurate relative state estimation. 
\end{itemize}

In the following discussions, we will first introduce these measurements and factors and then give the optimization problem for state estimation from the factor graph $\mathcal{G}_f$ .

\section{Front-end: Omnidirectional perception system and Measurement Modeling}\label{sect:front-end}
\begin{figure*}[ht]

    \centering
    \settowidth\limage{\includegraphics[height=4cm]{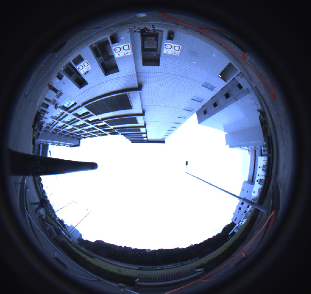}}
    \settowidth\limage{\includegraphics[height=4cm]{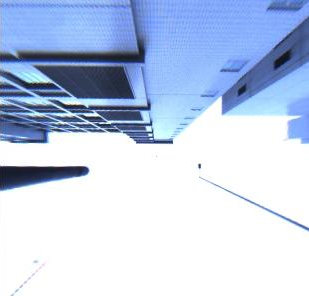}}
    \settowidth\rimage{\includegraphics[height=4cm]{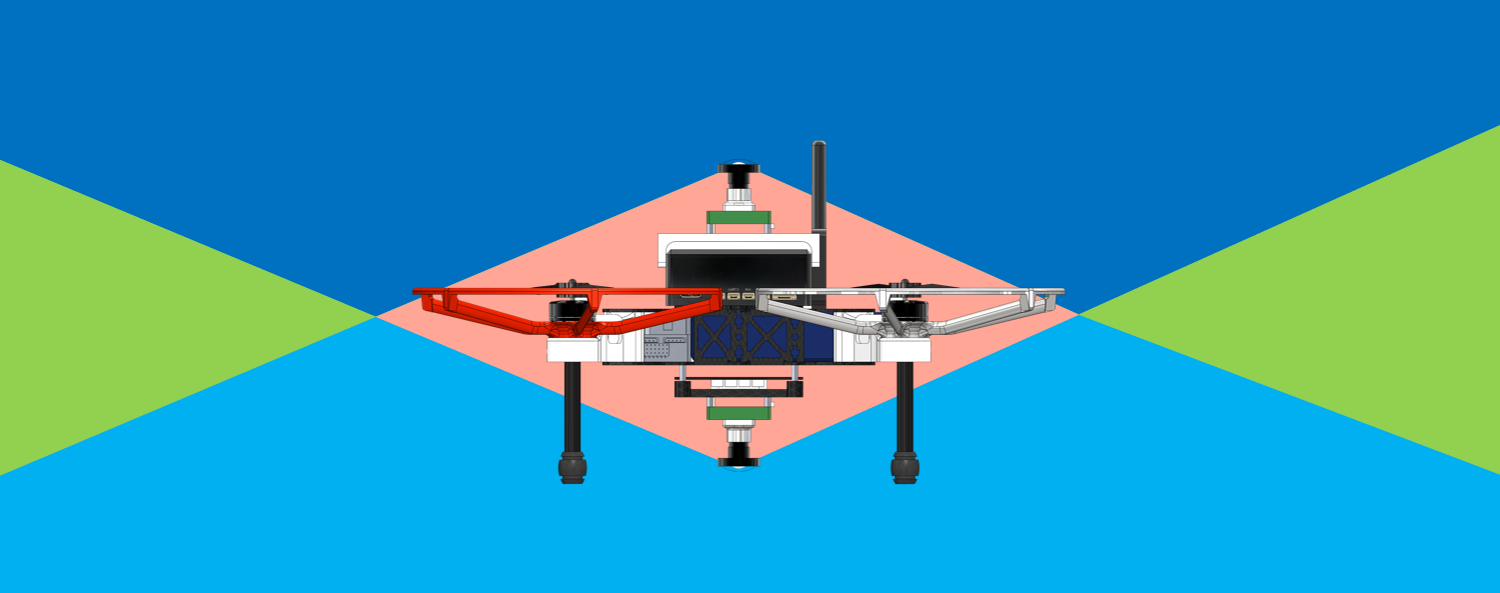}}
    \resizebox{0.8\textwidth}{!}{
        \begin{tabular}{p{\limage}p{\limage}p{\rimage}}
    \includegraphics[height=4cm]{raw}\newline
    \vspace{-0.5cm}
    \subcaption{}\label{fig:fisheye_raw}
        &   \includegraphics[height=4cm]{top}\newline
        \vspace{-0.5cm}
        \subcaption{}\label{fig:fisheye_top}
        &   \includegraphics[height=4cm]{fov_fisheye.PNG}\newline
        \vspace{-0.5cm}
        \subcaption{}\label{fig:fov_fisheye}
    \end{tabular}
    }
    \begin{subfigure}{0.8\textwidth}
        \centering
        \includegraphics[width=1.0\linewidth]{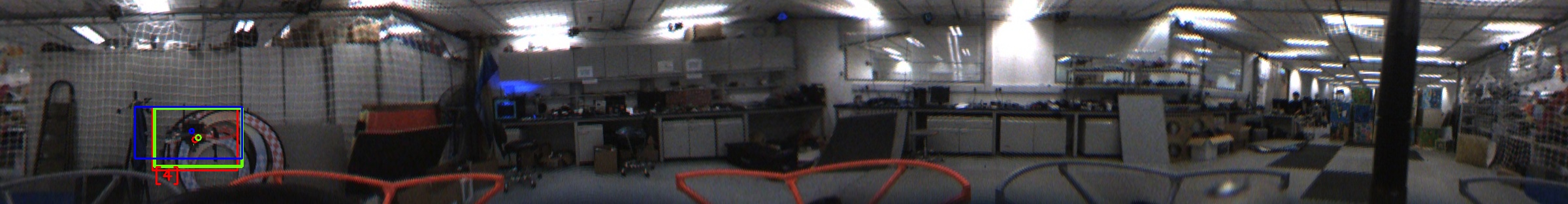}
        \caption{}\label{fig:fisheye_flatten}
    \end{subfigure}
    \vspace{-0.3cm}
    \caption{\small{(a): The raw image of the fisheye camera. 
    (b): The extracted distortion-free upward-facing view. 
    (c): This figure illuminates the available FoV for Omni-swarm. 
    The blue and green areas denote the fisheye cameras' coverage, which is omnidirectional. 
    The pink area denotes the dead zone. 
    The green area denotes the stereo coverage, which is utilized for map-based localization. 
    (d): The concatenation of distortion-free images.
    The red bounding box and circle are extracted by the visual object detection.
    The green bounding  box and circle are tracked by the visual object tracking.
    The blue  bounding box and circle are reprojected estimated states of the drone.
    The circles are the center of the bounding box.
    }}
    \vspace{-0.5cm}
\end{figure*}

\subsection{Omnidirectional Visual Inertial Odometry}
To achieve omnidirectional visual perception, Omni-swarm adopts two 235-degree FoV fisheye cameras to cover all the surrounding directions, as shown in Fig. \ref{fig:fov_fisheye}. 
Based on previous works on omnidirectional VIO \cite{qin2018vins,gao2017dual,gao2020autonomous}, we develop VINS-Fisheye\footnote{https://github.com/HKUST-Aerial-Robotics/VINS-Fisheye}, which is an omnidirectional visual-inertial navigation system derived from VINS-Fusion \footnote{https://github.com/HKUST-Aerial-Robotics/VINS-Fusion}.
VINS-Fisheye uses an IMU and previously mentioned stereo fisheye cameras to estimate ego-motion.

Because of the massive distortion of the fisheye cameras, it is hard to directly apply the existing visual algorithms to the raw image data produced by them. 
As an alternative, we reproject the raw image captured by a fisheye camera into five distortion-free images for later algorithms, which follows the pipeline proposed in \cite{gao2017dual} and \cite{gao2020autonomous}. 
An example of a raw image and processed distortion-free images is shown in Fig. \ref{fig:fisheye_flatten}.
After reprojecting the raw fisheye images, VIO will be extracted based on these distortion-free images to provide real-time local pose and velocity estimation.
Due to the long-term drifting of the VIO, instead of directly fusing the original odometry, we fuse the 4-DoF relative pose extracted from  the VIO in the back-end, which can be modeled as
\begin{equation}\label{eq:delta_p}
    \mathbf{z}_{\mathbf{\delta P}_{i}}^t =(\tensor*[]{\mathbf{\tilde P}}{_i^{t-1}})^{-1}(\tensor*[]{\mathbf{\tilde P}}{_i^t}) =  (\tensor*[]{\mathbf{P}}{_i^{t-1}})^{-1}(\tensor*[]{\mathbf{P}}{_i^t}) + \mathbf{n}_{vio},
\end{equation}
where the noise of the relative pose is assumed as Gaussian, $\mathbf{n}_{vio} \sim \mathcal{N}(0,\,\mathbf{\sigma}_{vio}^{2})$.

The VIO keyframes contain the distortion-free images generated from the fisheye stereo cameras and the external parameters of the camera, and the real-time pose estimation is utilized for further processing to avoid redundant computation.

\subsection{Visual Drone Tracking Module}\label{sect:visual_detection}
\setlength{\textfloatsep}{2pt}

\begin{algorithm}[h]
    \SetAlgoLined
    \KwData{Set of estimated drones $\mathcal{D}_e$, set of visual target files $\mathcal{V}_{TF}$}
    \KwIn{$\mathcal{I}$, the set of distortion-free images.}
    \KwOut{$\mathcal{Z}_D$, the set of visual target measurements.}
    \SetKwProg{Fn}{Function}{}{end function}
    \SetKwFunction{Match}{\textbf{Match}}
    \SetKwFunction{FImageProcess}{\textbf{VisualTargetTracking}}
    \SetKwFunction{FYolo}{\textbf{YOLO}}
    \SetKwFunction{FGNN}{\textbf{Hungarian}}
    \SetKwFunction{FMOSSE}{\textbf{MOSSETrackers}}
    \SetKwFunction{FNewMOSSE}{\textbf{NewMOSSE}}
    \SetKwFunction{FONES}{\textbf{ONES}}
    \SetKwFunction{FCreateVTF}{\textbf{NewVTF}}
    \SetKwFunction{FCreateVTFAny}{\textbf{NewAnonVTF}}
    \SetKwFunction{FPoseEst}{\textbf{PoseEstimation}}
    \Fn{\FImageProcess{$\mathcal{I}$}} {
        $\mathcal{V}_{TC} \leftarrow $ \FYolo{$\mathcal{I}$}

        %Apply tracker first
        $\mathcal{V}_{TF}  \leftarrow $ \FMOSSE{$\mathcal{I}, \mathcal{V}_{TF}$}

        %Match detection with tracker
        % $\mathbf{C}  \leftarrow$ $(1+\tau_e)*$\FONES{$\Vert\mathcal{V}_{TC}\Vert + \Vert\mathcal{V}_{TF}\Vert+ \Vert \mathcal{D}_e \Vert$} \label{alg:init_c} \\
        \For{${V_{TC}}_i \in \mathcal{V}_{TC}$} {
            \For{${V_{TF}}_j \in \mathcal{V}_{TF}$} {
                $\mathbf{C}_{i,j} \leftarrow c_{i,j}({V_{TC}}_i, {V_{TF}}_i)$
            }
            \For{$j \in \mathcal{D}_{e}$} {
                $\mathbf{C}_{i,j+\Vert\mathcal{V}_{TF}\Vert} \leftarrow c_{i,j}({V_{TC}}_i, D_i)$
            }
        }
        $\mathcal{M} \leftarrow$ \FGNN{$\mathbf{C}$}  \label{alg:gnn}  \\
        
        ${\mathcal{V}_{TF}}^u \leftarrow \varnothing$\\
        \For{${V_{TC}}_i \in \mathcal{V}_{TC}$} {
            \If {${V_{TC}}_i \in \mathcal{M}$} {
                $j \leftarrow \mathcal{M}(i)$\\
                \If{$j < \Vert\mathcal{V}_{TF}\Vert$} {
                    ${V_{TF}}_j.T \leftarrow$\FNewMOSSE{$\mathcal{I}, {V_{TC}}_i.\mathbf{B}$} \label{alg:replace}
                }
                \Else {
                    % $V_{TF}^* \leftarrow$ \\
                    % $\mathcal{V}_{TF} \leftarrow \mathcal{V}_{TF} \cup V_{TF}^* $\\
                    ${\mathcal{V}_{TF}}^u \leftarrow  {\mathcal{V}_{TF}}^u \cup\ $\FCreateVTF{${V_{TC}}_i, D_i$}  \label{alg:new1}
                }
            } \Else {
                % $V_{TF}^* \leftarrow$ \FCreateVTFAny{${V_{TC}}_i$} \label{alg:new2}\\
                % $\mathcal{V}_{TF} \leftarrow \mathcal{V}_{TF} \cup V_{TF}^* $ \\
                ${\mathcal{V}_{TF}}^u \leftarrow  {\mathcal{V}_{TF}}^u \cup $ \FCreateVTFAny{${V_{TC}}_i$} \label{alg:new2}
            }
        }

        $\mathcal{Z}_D \leftarrow \varnothing$\\
        \For{${V_{TF}}_i \in {\mathcal{V}_{TF}}^u $} {
            $\mathcal{Z}_D \leftarrow \mathcal{Z}_D \cup $ \FPoseEst{${V_{TF}}_i$}
        }

        ${\mathcal{V}_{TF}} \leftarrow  {\mathcal{V}_{TF}} \cup {\mathcal{V}_{TF}}^u$

        \Return  $\mathcal{Z}_D$
    }
    \caption{Visual Target Tracking Module}\label{alg:visual_tracker}
\end{algorithm}

In Omni-swarm, we introduce a visual drone tracking (VDT) module for visual tracking and relative pose estimation of other drones in the swarm.
VDT uses visual object tracking techniques to keep tracking all the drone targets it detects.
A drone target tracked by VDT is recorded as a visual target file ${V_{TF}}_{i^*}$ in the module, where $i^*\in\{1^*, 2^*, 3^* ... \}$ is its ID inside VDT, and ${V_{TF}}_{i^*}.d$ is the drone that tracks this visual target file.
A visual target file ${V_{TF}}_{i^*}$ can be anonymous, i.e., its drone ID is unknown, or identified, i.e., its drone ID has been successfully associated.

Alg. \ref{alg:visual_tracker} demonstrates the algorithm of VDT, which will be described in detail in this section.
VDT starts by detecting the drones in the image using a visual object detector. 
A drone detected by this detector creates a visual target candidate ${V_{TC}}_{i^*}$,
which may create a visual target file or update a visual target file in subsequent algorithms.
VDT will attempt to associate the visual target candidates with the drones observed by the state estimation and visual target files to identify its drone ID and VDT ID.
However, this process may not be successful, either because the system is not properly initialized or because the ${V_{TC}}_{i^*}$ is a false target.
Visual target candidates that are not correctly matched to a drone will create anonymous visual target files in the subsequent algorithms.
Finally, we use a 6-DoF pose estimator for relative pose estimation of these visual target files. 

\subsubsection{Visual Drone Detection}
We adopt YOLOv4-tiny\cite{yolov3,bochkovskiy2020yolov4}, one of the state-of-the-art visual object detection approaches based on a convolutional neural network (CNN), for detecting the 2D bounding boxes of the drones on the distortion-free images extracted from raw fisheye images. 
In practice, limited to the computational resources, only the upward-facing camera is used.
The network is trained with our custom data to efficiently detect our custom drones.
A demonstration of the detected bounding boxes is shown in Fig. \ref{fig:fisheye_flatten}.

As the \textbf{YOLO} function in Alg. \ref{alg:visual_tracker} shows, the result of the visual object tracking is a set of target candidates $\mathcal{V}_{TC}=\{ {V_{TC}}_{1^*}, {V_{TC}}_{2^*}\ ..\}$.
A target candidate  ${V_{TC}}_{i^*}$ detected by the visual object detector can be represented by a bounding box $\mathbf{B}_{i^*}=[\mathbf{B}_{i^*}^L, \mathbf{B}_{i^*}^R]$, where $\mathbf{B}_{i^*}^L$ is the coordinates of the left-upper corner of bounding box and  $ \mathbf{B}_{i^*}^R$ is the coordinates of the right-bottom corner.

\subsubsection{Visual Object Tracking of Drones}
In VDT,  we use MOSSE\cite{bolme2010visual} as the object tracking technique to keep track of visual target files.
MOSSE is a visual tracking approach that combines robustness and efficiency.
The images $\mathcal{I}$ extracted from the original image by VINS-Fisheye are also utilized for visual object tracking, as the function \textbf{MOSSETrackers} in Alg. \ref{alg:visual_tracker}. 

If a new visual target file ${V_{TC}}_{i^*}$ is successfully associated with an existing visual target file  ${V_{TF}}_{j^*}$, then the visual tracker of  ${V_{TF}}_{j^*}$ will be replaced by a new visual object tracker initialized by  $\mathbf{B}_{i^*}$ (line 15 of Alg. \ref{alg:visual_tracker}).
Otherwise, we use $\mathbf{B}_{i^*}$ to create a new visual target file in the function \textbf{NewVTF} and function \textbf{NewAnonVTF}. 

In practice, since visual object detection consumes far more computational resources than visual object tracking, we run the former at a lower frequency (1 Hz in practice) than the latter (10 Hz in practice).

\subsubsection{Data Association}\label{sect:ass}
For a homogeneous swarm, one issue is the association between the detected targets and the drones observed by state estimation, but  this is essential for subsequent state estimation.
The problem of data association for visual drone detection in Omni-swarm is divided into two sub-problems.
The first is to initialize the state estimation, and the second is to perform global-nearest-neighbor (GNN) matching on the target based on the existing estimation. 
The state estimation of Omni-swarm can be initialized by a variety of methods, which will be described in detail later in Sect. \ref{sect:init}, including an approach using only anonymous visual target files.
On the other hand, GNN has been widely adopted in multi-target tracking algorithms \cite{betke2016data, konstantinova2003study, sinha2012track} for data association.

Here, we propose a unified framework to associate the visual target candidates to visual target files and to the estimated drones. The latter two are collectively referred to as objects.
This method does not require the system to be fully initialized, considering that it also has the responsibility of tracking of anonymous targets.
We project all the drones estimated by Omni-swarm onto the image plane, as shown by the blue bounding box in Fig. \ref{fig:fisheye_flatten}.
These projected bounding boxes are treated the same as the bounding box of the visual target files (the green bounding box in Fig. \ref{fig:fisheye_flatten}), as the objects in GNN.
The corresponding cost between candidates (the red bounding box in Fig. \ref{fig:fisheye_flatten}) and objects is defined by
\begin{equation}\label{eq:dist}
        c_{i^*, j} = 
        \ \left\{
            \begin{aligned}
                &1 - o_{i^*,j}, &  o_{i^*,j} > 0\\
                &+\infty, & o_{i^*,j} = 0,
            \end{aligned}
        \right.
\end{equation}
where $o_{i^*,j}$ is defined as the overlap of the two bounding boxes, which is defined as
$$o_{i^*,j} = A^i_{i^*,j} / max(A_{j}, A_{i^*}),$$
where $A^i_{i^*,j}$ is the intersection area of bounding box $\mathbf{B}_{i^*}$ and $\mathbf{B}_{j}$, and $A_{j}$ and $A_{i^*}$ are the areas of two the bounding boxes.

With the corresponding cost defined here, we create a corresponding cost matrix $\mathbf{C}$ of the candidates and objects.
In addition, we add dummy objects and dummy candidates with  a $+\infty$ distance to all normal candidates and objects to represent missing detections and false alarms.
Finally, Omni-swarm uses the Hungarian algorithm\cite{kuhn1955hungarian, bertsekas1992forward} to solve the GNN problem defined by Betke et al.\cite{betke2016data} with the corresponding cost matrix $\mathbf{C}$.
Alg. \ref{alg:visual_tracker}, line \ref{alg:gnn}, shows this procedure.
The output of the  Hungarian algorithm is a corresponding dictionary in which the candidates assigned to dummy objects are not present.

After the Hungarian algorithm is performed, for each candidate ${V_{TF}}_{i*}$ , if ${V_{TF}}_{i*}$ is assigned to a visual target file, the old visual target files' visual object tracker will be updated in Alg. \ref{alg:visual_tracker}, line \ref{alg:replace}.
If a ${V_{TF}}_{i*}$ has been assigned to a drone observed by state estimation, it will create a named visual target; otherwise, it creates a new anonymous visual target file. 
This is shown in Alg. \ref{alg:visual_tracker}, line \ref{alg:new1} and line \ref{alg:new2}.

\subsubsection{6-DoF Pose Estimator}\label{pose_est}
The VDT adopts a CNN-based 6-Dof pose estimator \cite{pavlakos20176, pavliv2021tracking} to extract the accurate 6-DoF relative pose estimation of the visual target files as the function \textbf{PoseEstimation}.
Specifically, this approach performs relative pose estimation by using CNNs to extract semantic feature points of objects in images and build optimization problems. 
The network is trained with the distortion-free images collected from the onboard fisheye cameras for efficiently detecting our custom drones.
Before solving the full-perspective pose estimation problem, an additional outlier rejection on features using EPnP with RANSAC  \cite{lepetit2009epnp} is performed.

The 6-DoF relative poses extracted by VDT from drone $k$ to drone $j$ will be adopted as visual detection measurements, which can be modeled as
\begin{equation}\label{eq:D}
    {\mathbf z_D}_{k\rightarrow j}^{t_0} = (\tensor*[^{v_i}]{\mathbf{P}}{_k^{t_0}})^{-1}(\tensor*[^{v_i}]{\mathbf{P}}{_j^{t_0}}) + \mathbf{n}_D,
\end{equation}
where this relationship  stands with any reference frame $v_i$ and $\mathbf{n}_D$ is Gaussian noise.
We call this result the visual detection measurement.
We only perform pose estimation on newly created or updated VTFs to save computational resources.

\subsection{Multi-drone Map-based Localization Module}
Similar to the approach used in VINS-Fisheye, we also perform visual object detection on distortion-free images.
The multi-drone map-based localization (MDML) module performs relative localization and eliminates the drifting of the VIO by identifying the locations visited by all the drones of the aerial swarm. 
Sparse maps, which contain landmarks and keyframes, are simultaneously generated by local and remote measurements on every drone. 
Beyond optimizing the sparse maps, we utilize them to extract relative poses among drones in the swarm by a loop closure detection procedure. 
The MDML module is decentralized,  running on each drone separately.

\begin{algorithm}[t]
    \SetAlgoLined
    \KwData{Local Visual Database $\mathcal{D}_l$, Remote Visual Database $\mathcal{D}_r$, Local Drone $k$}
    \KwIn{$\mathcal{F}_j^{t_1}$}
    \SetKwProg{Fn}{Function}{}{end function}
    \SetKwFunction{FLOOP}{\textbf{LOOP\_DETECTION}}
    \SetKwFunction{FQUERY}{\textbf{KF\_QUERY}}
    \SetKwFunction{FPnP}{\textbf{PNP\_RANSAC}}
    \SetKwFunction{FBF}{\textbf{BF\_MATCHER}}
    \SetKwFunction{FGCK}{\textbf{G\_CHECK}}
    \SetKwFunction{FKNN}{\textbf{KNN\_SEARCH}}
    \SetKwFunction{FADD}{\textbf{ADD}}
    \Fn{\FQUERY{$\mathcal{F}_j^{t_1}$, $\mathcal{D}$}} {
        $\mathcal{C} \leftarrow \varnothing$ \\
        % $\mathcal{D}_{l_n},\mathcal{D}_{r_n} \leftarrow \mathcal{D}_l, \mathcal{D}_r$\\
        \For{$\tensor*[^c]{\mathcal{K}}{_j^{t_1}} \in \mathcal{F}_j^{t_1}$} {
            \If{$j=k$} {
                $\mathcal{C}\leftarrow \mathcal{C} \ \cup \ $\FKNN{$\tensor*[^c]{\mathcal{K}}{_j^{t_1}}, \mathcal{D}_r, \tau_{fl}$} \\
            }
            $\mathcal{C}\leftarrow \mathcal{C} \ \cup \ $\FKNN{$\tensor*[^c]{\mathcal{K}}{_j^{t_1}}, \mathcal{D}_l, \tau_{fl}$} \\
            
        }

        \FADD{${\mathcal{K}}{_j^{t_1}}, j=k, \mathcal{D}_l, \mathcal{D}_r$}\\
        $\mathcal{K} \leftarrow \operatorname*{argmin}_{(\tensor*[^c]{\mathcal{K}}{}){_f}}{\left\Vert (\tensor*[^c]{\mathcal{K}}{}){_f} - (\tensor*[^c]{\mathcal{K}}{_j^{t_1}})_f \right\Vert }$ \\
        \Return $\mathcal{K}.{\mathcal{F}}$
    }
    
    \Fn{\FGCK{$ ^{V_j}\hat{\mathbf{R}}_i^{t_0}, \tilde{\mathbf{R}}_i^{t_0}, \tilde{\mathbf{R}}_j^{t_1}$}} {
    $\delta \hat{\mathbf{R}}^{t_0 \rightarrow t_1}_{i \rightarrow j} \leftarrow ( ^{V_j}\hat{\mathbf{R}}_i^{t_0})^{-1} \tilde{\mathbf{R}}_i^{t_0}$\\
    $ ^{V_i}\hat{\mathbf R}^{t_1}_{j}\leftarrow\tilde{\mathbf{R}}_i^{t_0} \delta \hat{\mathbf{R}}^{t_0 \rightarrow t_1}_{i \rightarrow j}$ \\
    $\delta\psi \leftarrow (\tilde{\mathbf{R}}_j^{t_1})_\psi - ( ^{V_i}\hat{\mathbf R}^{t_1}_{j})_\psi$ \\
    $\Delta \mathbf{R} \leftarrow (\mathbf{R}_z(\delta\psi)\   ^{V_i}\hat{\mathbf R}^{t_1}_{j})^T\  \tilde{\mathbf{R}}_j^{t_1}$ \\
    \Return $\left\Vert \Delta \mathbf{R} \right\Vert > \tau_{rot}$
    }
    
    \Fn{\FLOOP{$\mathcal{F}_j^{t_1}$}} {
    $\mathcal{F}_i^{t_0}\leftarrow$\FQUERY{$\mathcal{F}_j^{t_1}$, $\mathcal{D}$} \\
    \If {$\mathcal{F}_i^{t_0} \neq \emptyset$} {
    ${\mathcal{P}_{2d}}^{t_0}_i, {\mathcal{P}_{3d}}^{t_1}_j\leftarrow$\FBF{$\mathcal{F}_i^{t_0}$, $\mathcal{F}_j^{t_1}$} \\
    $inliers, ^{V_j}{\hat{\mathbf T}}_i^{t_0}\leftarrow$\FPnP{${\mathcal{P}_{2d}}^{t_0}_i$, ${\mathcal{P}_{3d}}^{t_1}_j$} \\
    \If{$inliers  \geq \tau_{in}$} {
        \If{\FGCK{$(\tensor*[^{V_j}]{\hat{\mathbf{T}}}{_i^{t_0}})_R,(\mathcal{F}_i^{t_0})_R,(\mathcal{F}_j^{t_1})_R$}} {
            ${\mathbf{z}_\mathcal{L}}_{i\rightarrow j}^{t_0 \rightarrow t_1} \leftarrow\left((\tensor*[^{V_j}]{\hat{\mathbf T}}{_i^{t_0}})^{-1} \tilde{\mathbf T}_j^{t_1}\right)_P$ \\
            \Return ${\mathbf{z}_\mathcal{L}}_{i\rightarrow j}^{t_0 \rightarrow t_1}$
        }
    }
    }
    \Return $\emptyset$
    }
    
    \caption{Multi-drone Map-based Localization Algorithm for Drone $k$}\label{alg:loop_detection}
\end{algorithm}

\subsubsection{Multi-drone map-based Localization Procedure}\label{sect:loop_detection}

\begin{figure}[ht]
    \centering
    \includegraphics[width=1.0\linewidth]{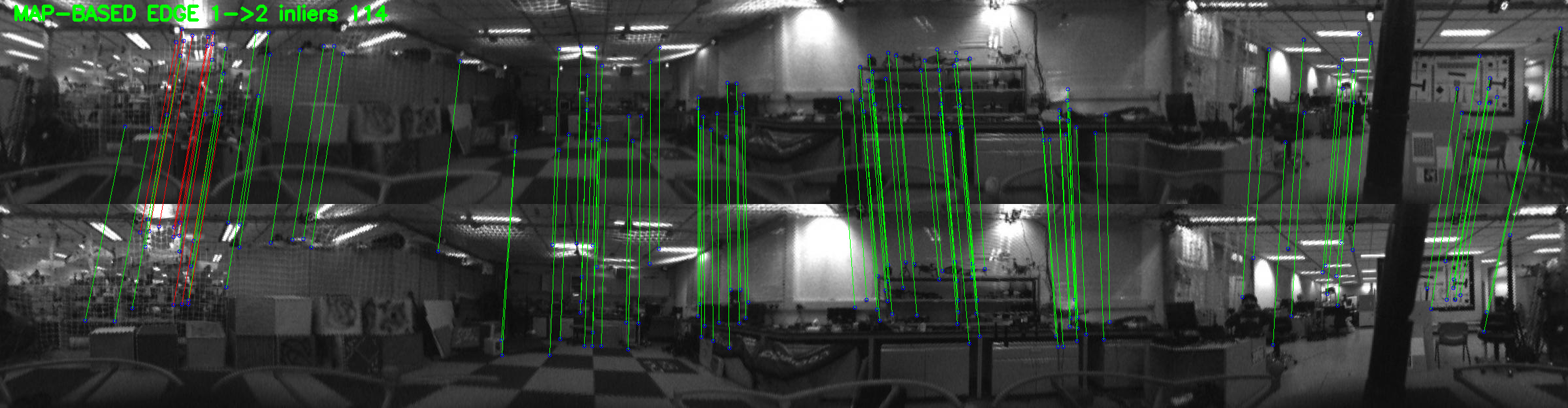}
    \caption{\small{
        A demonstration of detected map-based measurements between different drones.
        Green lines denote the inlier landmarks correspondences, and red lines denote the outliers.
    }}\label{fig:ex1_drone2}

\end{figure}

When the MDML module receives the VIO keyframes, it uses MobileNetVLAD \cite{arandjelovic2016netvlad, sarlin2019coarse} to extract the global features and uses SuperPoint \cite{detone2018superpoint} to extract the landmarks and the corresponding descriptor.
Correspondences between landmarks from the upward-facing and downward-facing cameras are established by performing feature matching, and the matched landmarks are triangulated for estimating their 3D positions in the local frame.
The global descriptor and landmarks, together with the odometry and extrinsic, are packed into keyframe $\mathcal{F}^t_i$, which will be broadcast to the entire swarm later.
It is worth noting that we broadcast only the feature points with the 3D position successfully estimated (some of the features will fail to establish correspondence) to reduce the amount of communication.

To store and retrieve these keyframes, we build databases based on Faiss\cite{johnson2019billion}, which is a vector similarity retrieval database. 
The keyframes $\mathcal{F}^{t_1}_j$, which are indexed by the corresponding global descriptors, are saved in the databases as maps.
After the module receives a keyframe remotely or locally, loop closure detection is adopted to extract the relative pose. 
There are two separate visual databases on each drone; a remote database  $\mathcal{D}_r$ that stores keyframes from remote drones, and  a local database $\mathcal{D}_l$ that stores keyframes from the local drone.
Extracting the map-based measurement for a pair of keyframes from $\mathcal{D}_r$ is avoided to save computational resources since all the extracted map-based measurements are broadcast to the whole swarm.

The procedure of loop closure detection is shown in Alg. \ref{alg:loop_detection}.
Suppose a drone $k$ receives a keyframe $\mathcal{F}^{t_1}_j$ from drone $j$, where if $j=k$, the keyframe is generated by the drone itself. 
The most similar keyframe $\mathcal{F}^{t_0}_i$ to $\mathcal{F}^{t_1}_j$ in the databases is retrieved by function \textbf{KF\_QUERY}.
We use function \textbf{KNN\_SEARCH} to retrieve the $K$ nearest-neighbor of the descriptors from the Faiss database, where $K$ is set to 5 in practice.
When the search is completed, the new keyframes are added to the local or remote database by function \textbf{ADD}.
Finally, function \textbf{KF\_QUERY} returns the corresponding keyframe $\mathcal{K}.{\mathcal{F}}$ of the nearest-neighbor distortion-frame image keyframe $\mathcal{K}$.

\subsubsection{Relative pose extraction}
Once the keyframe $\mathcal{F}_i^{t_0}$ is returned from the visual database, we establish the 2D-3D matches from the landmarks of $\mathcal{F}_i^{t_0}$ to landmarks of the $\mathcal{F}_j^{t_1}$ by using the landmark descriptors with a brute-force matcher \textbf{BF\_MATCHER}.
The brute-force matcher finds the correspondences of features with a minimum L2 distance, and uses a cross check to reduce outlier correspondence\footnote{A more detailed introduction to the brute-force matcher can be found at https://docs.opencv.org/4.x/dc/dc3/tutorial\_py\_matcher.html}.
Either keyframe retrieved from the database or brute-force matching may bring abnormal results.
To remove the outliers, the map-based measurement is verified with two methods in the map-based localization module.
\begin{itemize}
    \item Homography test \cite{zhang19963d} and Perspective-n-point (PnP) test with RANSAC \cite{lepetit2009epnp} between the 2D features of the incoming keyframe and the 3D positions of the queried keyframe. These tests are peformed in the function \textbf{PNP\_RANSAC}.
    \item Geometric test peformed by \textbf{G\_CHECK}. In this test, the consistency of gravity is checked with the 6-DoF pose extracted by PnP RANSAC.
\end{itemize}
If enough inliers are found in \textbf{PNP\_RANSAC} and the geometric test is passed, the map-based measurement ${\mathbf z_{\mathcal{L}}}_{i\rightarrow j}^{t_0 \rightarrow t_1}$ is considered valid.
In addition, ${\mathbf z_{\mathcal{L}}}_{i\rightarrow j}^{t_0 \rightarrow t_1}$ can be modeled as:

\begin{equation}\label{eq:L}
    {\mathbf z_\mathcal{L}}_{i\rightarrow j}^{t_0 \rightarrow t_1} = (\tensor*[^{v_k}]{\mathbf{P}}{_i^{t_0}})^{-1}(\tensor*[^{v_k}]{\mathbf{P}}{_j^{t_1}}) + \mathbf{n}_{\mathcal{L}},
\end{equation}
where this relationship stands with any reference frame $v_k$ and $\mathbf{n}_{\mathcal{L}}$ is Gaussian noise.

When $i \neq j$, the map-based measurement ${\mathbf z_{\mathcal{L}}}_{i\rightarrow j}^{t_0 \rightarrow t_1}$ provides sufficient observability of the relative pose of drone $i$ and drone $j$. 
This makes map-based measurements essential in the observability verification and initialization of state estimation.
If $t_0 \neq t_1 $, the map-based measurement represents the relative pose of the drones that visit the same place at different times, which  eliminates the accumulated drifting error of the VIO.

\subsection{UWB Measurement}
The distance measurements from the UWB module can be modeled as
\begin{equation}\label{eq:uwb}
    \mathbf{z}_{d_{i,j}}^t=\left \Vert \tensor*[^{v_k}]{\mathbf{X}}{_{i}^t} - \tensor*[^{v_k}]{\mathbf{X}}{_{i}^t} \right \Vert_2 + \mathbf{n}_d,
\end{equation}
where $\mathbf{n}_d \sim \mathcal{N}(0,\,\sigma_{d}^{2})$ is the Gaussian noise of the distance measurement.
% In practice, $\sigma_{d}$ is near 3-5 cm. 
The installation length of the antenna relative to the IMU is ignored in our model. 
% Considering the scale of the drone, the installation length of the antenna relative to the IMU can be ignored in our model. 

\section{Back-End: Graph-based Optimization for State Estimation}\label{sect:back-end}
\begin{figure}[ht]
    \centering
    % \begin{subfigure}{\textwidth}
    \centering
    \includegraphics[width=1.0\linewidth]{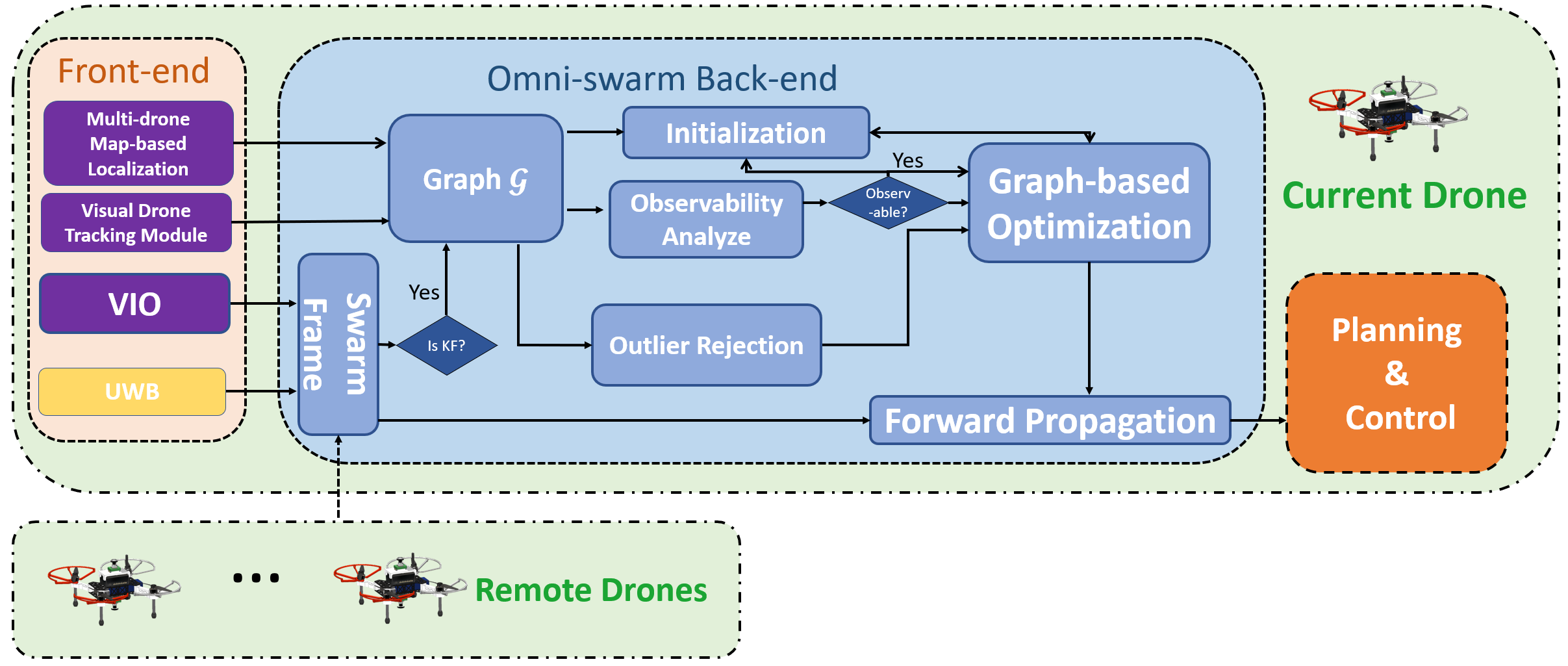}
    \caption{\small{The structure of the back-end of Omni-swarm.
    The swarm frames $\mathcal{SF}^t$ composed of VIO and UWB measurements are first judged to be swarm keyframes, and the non-keyframes are only utilized for forward propagation, while the swarm frames that qualify as swarm keyframes are added to the graph. 
    The system will evaluate whether the measurements in the graph meet the observability requirements. 
    When the requirements are met, the system solves the graph-based optimization for swarm state estimation. 
    Finally, we use the optimization results to perform forward propagation to obtain the real-time state of the swarm. }}\label{fig:estimation}
    \vspace{-0.5cm}
\end{figure}

\subsection{Swarm Keyframe Limitation}\label{sect:branddel}
Similar to our previous work \cite{xu2020decentralized}, a swarm frame $\mathcal{SF}^t$ is judged to be a swarm keyframe is by the motion of the drones in this frame relative to the last swarm keyframe, or a new drone is discovered in  $\mathcal{SF}^t$.
To limit the computational resources utilized by the back-end, we keep the numbers of swarm keyframes in $\mathcal{G}$ within a preset maximum number $m_{max}$.
In contrast to the sliding window of the swarm keyframes employed in our previous work \cite{xu2020decentralized}, here we use a random deletion mechanism to achieve better global consistency; 
once the graph's swarm keyframe in the graph grows to more than $m_{max}$, we randomly delete one swarm keyframe. 
The intuition of this approach is that the earlier swarm keyframes in $\mathcal{G}$ have less impact on the relative estimation.
With this approach, the earlier swarm keyframes in $\mathcal{G}$ are not immediately discarded but become more sparse, which allows us to use early swarm keyframes for optimization without affecting the computational speed due to too many keyframes.
This leads to less drifting and better global consistency.
In practice, $m_{max}$ is set to 100 as a tradeoff between the performance and accuracy.
Taken together, compared to the previous sliding window method, this method does not affect the accuracy of relative estimation, but obtains better global consistency, which is verified in Sect. \ref{sect:randdel}.

\subsection{Outlier Rejection}\label{sect:outlier}
One challenge for achieving robust swarm state estimation is the variety of factors that can cause measurement outliers.
Although we perform several outlier rejections in the front-end, some outliers will always be transferred to the back-end.
Here we introduce the outlier rejection techniques adopted by the back-end of Omni-swarm.

\subsubsection{Unified outlier rejection for visual target measurements and map-based measurements}
Pairwise-consistency measurement set maximization (PCM), proposed by Mangelson et al. \cite{mangelson2018pairwise}, is the state-of-the-art outlier rejection technique for loop closure (also called map-based measurements in the paper) and has been verified on real-world SLAM systems  \cite{lajoie2020door} \cite{rosinol2020kimera}.
In Omni-swarm, we introduce a unified outlier rejection module for the visual target measurements and the map-based measurements (collectively referred to as relative pose measurements) based on PCM.
This method is applicable to visual detection measurement because the visual target measurements are also 6-DoF relative pose measurements.
Our PCM module combines the inter-drone and intra-drone PCM outlier rejection methods \cite{lajoie2020door, rosinol2020kimera, mangelson2018pairwise}.

The core of the PCM outlier rejection is consistency graphs\cite{mangelson2018pairwise}.
A consistency graph is a graph  $\mathbf{G}^{PCM}_{i, j} = \{V^{PCM}_{i,j}, \mathcal{E}^{PCM}_{i,j}\}$, where vertex $v\in V^{PCM}_{i,j}$ represents a relative pose measurement between drone $i$ and drone $j$, and each edge $e\in \mathcal{E}^{PCM}_{i,j}$ represents the consistency of the measurements.
Only the consistent pairs-of-vertices are connected with an edge. 
We maintain a set of consistency graphs $\left\{\mathbf{G}^{PCM}_{i, j}| i\in \mathcal{D}, j \in \mathcal{D} \right\}$ of each pair of drones. 
When $i\neq j$,  $\mathbf{G}^{PCM}_{i, j}$ denotes the inter-drone measurements, and vice versa for the intra-drone measurements.

Prior to graph-based optimization, we will first update the consistency graphs $\left\{\mathbf{G}^{PCM}_{i, j}| i\in \mathcal{D}, j \in \mathcal{D} \right\}$ with the newly received relative pose measurements in $\mathcal{G}$. 
Similar to \cite{rosinol2020kimera}, the consistency graph is updated incrementally to save the computational power.
For every updated $\mathbf{G}^{PCM}_{i, j}$, we adopt the fast maximum clique method proposed by Pattabiraman et al.\cite{pattabiraman2015fast} to solve the PCM problem defined in \cite{mangelson2018pairwise} and thus determine the inner measurements.
Each drone only solves the PCM problem with itself involved to save computational power and shares the inlier measurements with the other drones in the aerial swarm.

\subsubsection{Outlier Rejection for Distance Measurements}

In our experiments, we find that the UWB generates significant outliers when two drones have a large relative elevation angle. 
This is because of the occlusion of the drones' airframe, and we decide to flag out the measurement $z^t_{d_{i,j}}$, which satisfies
\begin{equation}
    \left\vert \arcsin\left( z^t_{d_{i,j}}/\left(\tensor*[^{v_k}]{z}{_i^t} - \tensor*[^{v_k}]{z}{_j^t}\right)\right) \right\vert > \tau_{ele},
\end{equation}
where $\tensor*[^{v_k}]{z}{_i^t}$ and $\tensor*[^{v_k}]{z}{_j^t}$ is the estimated height of drone $i$ and $j$, respectively, and $\tau_{ele}$ is the elevation threshold, which is set to 37$^\circ$ in practice.
$\tau_{ele}$ is chosen regarding the angle at which the drone fuselage obscures the UWB.
Finally, we utilize the residual defined in Eq. (\ref{eq:uwb_res}) without the Huber loss to test if the distance measurement is consistent with the current estimation, and $\mathbf{r}_d(\mathbf{z}_{d_{ij}}^t, \mathcal{X}_k)>\tau_d$ will be flagged as an outlier, where $\tau_d$ is the distance outlier threshold, which is set to 0.3 m in practice.
The expected error of UWB measurements is no more than 10 cm, and we choose it three times as $\tau_d$ to ensure that the correct values are not overly filtered out. 
When $\tau_d$ is too large, the outlier will interfere with the estimate estimation, and when $\tau_d$ is too small, the correct measurement will be rejected due to inaccurate initialization.

\subsection{Optimization Problem}\label{sect:opti}
After the outlier rejection, the factor graph $\mathcal{G}_f$ will be constructed from the graph $\mathcal{G}$ with poses as its variables and inlier measurements as the factors.
Once we set up the factor graph, a non-linear least-squares optimization problem is built to solve the maximum a posteriori (MAP) inference of the factor graph \cite{dellaert2017factor} for swarm state estimation.
For a drone $k$, the full state vector $\mathcal{X}_k$ of the swarm state estimation problem is defined as:
\vspace{-0.2cm}
\begin{equation}
\begin{aligned}
    [ ^{v_k}\mathbf{\hat X}_{0}^{t_0 T}, & \ ^{v_k}\mathbf{\hat \psi}_{0}^{t_0} \ ... \ ^{v_k}\mathbf{\hat X}_{0}^{t_{m - 1} T},\ ^{v_k}\mathbf{\hat \psi}_{0}^{t_{m - 1}},\\
    &...\ \ ^{v_k}\mathbf{\hat X}_{1}^{t_0 T},\ ^{v_k}\mathbf{\hat \psi}_{1}^{t_0} \ ... \ ^{v_k}\mathbf{\hat X}_{n-1}^{t_{m - 1} T},\ ^{v_k}\mathbf{\hat \psi}_{n-1}^{t_{m - 1}}]^T,
\end{aligned}
\end{equation}
where $\left[^{v_k}\mathbf{\hat X}_{i}^{t T},\ ^{v_k}\mathbf{\hat \psi}_{i}^{t}\right]^T$ is the state vector of 4-DoF pose $^{v_k}\mathbf{\hat P}_{i}^{t}$, $n$ is the number of drones in the swarm system, and $m$ is the number of keyframes in the graph.
The non-linear least-squares optimization problem for MAP inference is expressed as the following formulation:
\begin{equation}\label{eq:opti_eq}
\begin{aligned}
    \min_{\mathcal{X}_k} \Bigg\{
    &\sum_{(i,t)\in\mathcal{S}} \left\Vert \mathbf{r}_\mathcal{RP} \left(\mathbf{z}_{\mathbf{\delta P_{i}}}^t, \mathcal{X}_k \right) \right\Vert_\Sigma^2 + \\
    &\sum_{(i,j,t)\in\mathcal{U}} \rho\left( \left\Vert  \mathbf{r}_d(\mathbf{z}_{d_{ij}}^t, \mathcal{X}_k) \right\Vert^2_\Sigma \right) \\ 
    +&\sum_{(i,j,t)\in\mathcal{VD}} \rho\left(\left\Vert \mathbf{r}_\mathcal{RP}({\mathbf{z}_D}_{i\rightarrow j}^{t}, \mathcal{X}_k)\right\Vert_\Sigma^2 \right)  \\
    +&\sum_{\mathcal{L}_{k\rightarrow j}^{t_0 \rightarrow t_1}\in\mathcal{L}} 
    \rho\left(\left\Vert \mathbf{r}_\mathcal{RP}({\mathbf{z}_\mathcal{L}}_{i\rightarrow j}^{t_0 \rightarrow t_1}, \mathcal{X}_k)\right\Vert_\Sigma^2 \right) 
\Bigg\},
\end{aligned}
\end{equation}
where $\mathcal{S}$ is the set of all odometry factors; $\mathcal{VD}$ is the set of all visual detection factors,
$\mathcal{U}$ is the set of all distance factors,
$\mathcal{L}$ is the set of map-based factors,
$ \mathbf{r}_d(\mathbf{z}_{d_{ij}}^t, \mathcal{X}_k)$ is the residual of the distance factor,
where $\mathbf{r}_\mathcal{RP}(\cdot, \mathcal{X}_k)$ represents the residual of the relative pose, which is applicabe to the ego-motion factors, map-based factors and visual detection factors.
$\mathbf{r}_{\mathcal{RP}} \left(\mathbf{z}_{\mathbf{\delta P_{i}}}^t, \mathcal{X}_k \right)$ represents the residual of the ego-motion factor, ensuring the local consistency of drone $i$'s state,
$\mathbf{r}_\mathcal{RP}({\mathbf{z}_D}_{i\rightarrow j}^{t}, \mathcal{X}_k)$ is the residual of the visual detection factor, where $(i,j)$ presents the drone $j$ detected by drone $i$,
$\mathbf{r}_\mathcal{RP}({\mathbf{z}_\mathcal{L}}_{i\rightarrow j}^{t_0 \rightarrow t_1}, \mathcal{X}_k)$ represents the residual of the map-based factor, which ensures the global consistency and observability of the relative state.
Since some outlier measurements may be generated in distance measurements, map-based factor measurements, and visual detection measurements, we adopt the Huber norm  $\rho(s)$ \cite{huber1992robust}  to reduce the effect of possible outlier factors.

According to the measurement models stated in (\ref{eq:delta_p}),(\ref{eq:D}) and (\ref{eq:L}), the residual of the relative pose is defined as
\begin{equation} \label{eq:ego_res}
    \mathbf{r}_\mathcal{RP}({\mathbf{z}_\mathcal{RP}}_{i\rightarrow j}^{t_0 \rightarrow t_1}, \mathcal{X}_k) = \left({\mathbf{z}_\mathcal{RP}}_{i\rightarrow j}^{t_0 \rightarrow t_1} \right)^{-1} \left( ( ^{v_k}\mathbf{\hat P}_{i}^{t_0})^{-1} { }^{v_k}\mathbf{\hat P}_{j}^{t_1}\right),
\end{equation}
where $\mathbf{z}_\mathcal{RP}$ represents the relative pose measurements including the ego-motion measurement, map-based measurement, and visual target measurement.
Referring to the measurement model stated in Eq. (\ref{eq:uwb}), the residual of the distance is
\begin{equation}\label{eq:uwb_res}
\mathbf{r}_d(\mathbf{z}_{d_{ij}}^t, \mathcal{X}_k) = \mathbf{z}_{d_{ij}}^t - \left\Vert  ^{v_k}\mathbf{\hat x}_{i}^t -  ^{v_k}\mathbf{\hat x}_{j}^t \right \Vert.
\end{equation}

Finally, the Ceres Solver, an open-source C++ library developed by Google for modeling and solving non-linear least-squares optimization problems\cite{ceres-solver}, is adopted to solve the optimization problem with a trust region method using a sparse normal Cholesky decomposition is chosen as the optimization algorithm.
Meanwhile, since the planner or controller may require high-frequency real-time poses, the poses $\tensor*[^{v_k}]{\mathbf{\hat P}}{_i^t}$ and $\tensor*[^{v_k}]{\mathbf{\hat T}}{_i^t}$ are forward propagated based on the latest IMU propagated VIO $\tensor*[]{\mathbf{\tilde P}}{_i^t}$ and $\tensor*[]{\mathbf{\tilde T}}{_i^t_1}$ to predict the swarm state at a higher rate of 100 Hz following (\ref{eq:4to6}) and
\begin{equation}\label{eq:forward_pro}
    \tensor*[^{v_k}]{\mathbf{\hat P}}{_i^{t_1}} = \tensor*[^{v_k}]{\mathbf{\hat P}}{_i^{t}}(\tensor*[]{\mathbf{\tilde P}}{_i^{t}})^{-1} \tensor*[]{\mathbf{\tilde P}}{_i^{t_1}}.
\end{equation}
% VIO has good local accuracy, so the propagation does not lose the state estimation accuracy.

\subsection{Observability Analysis}\label{sect:obser}
\begin{figure}[ht]
    \centering
    \settowidth\limage{\includegraphics[height=2cm]{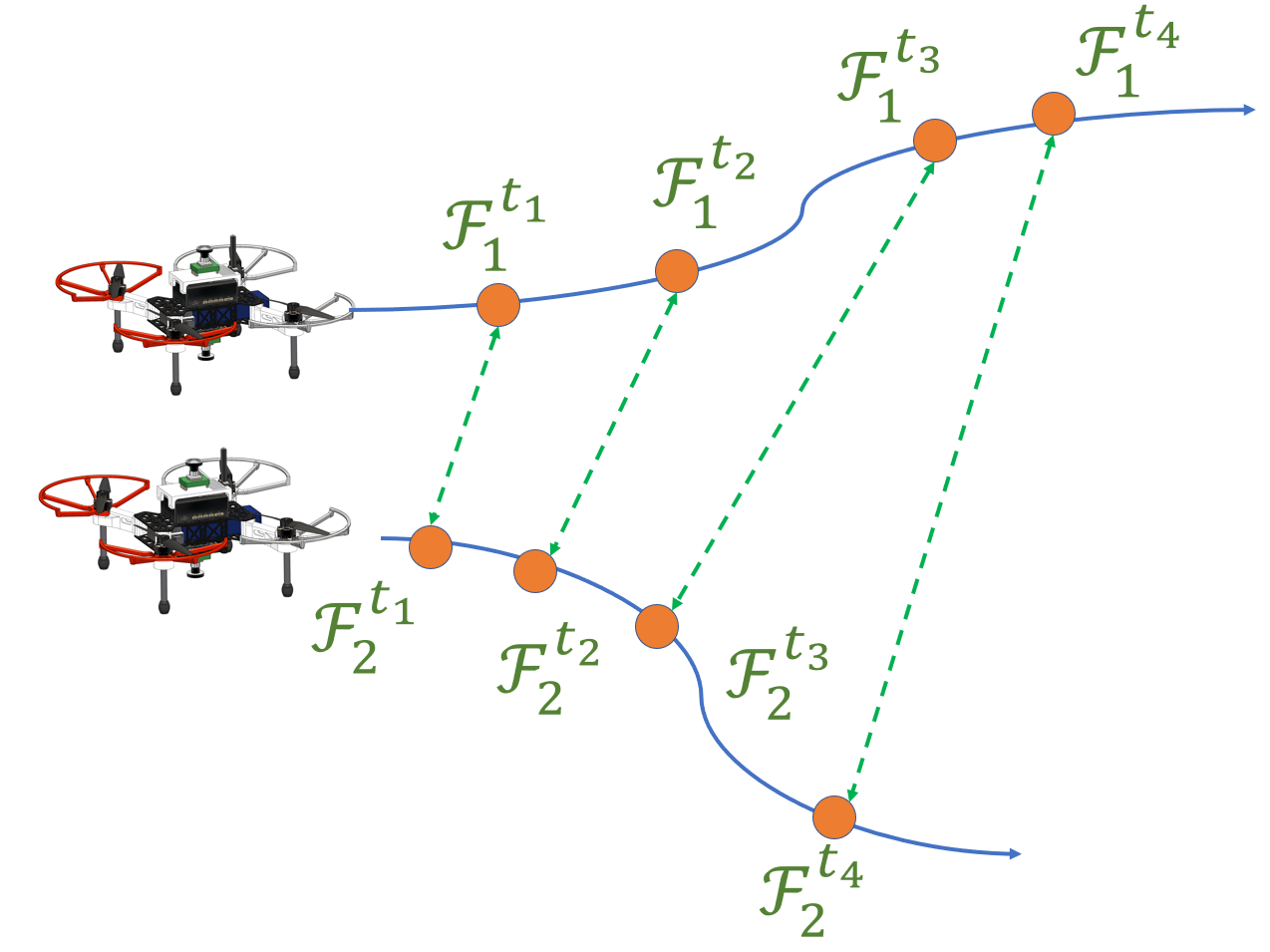}}
    \settowidth\mimage{\includegraphics[height=2cm]{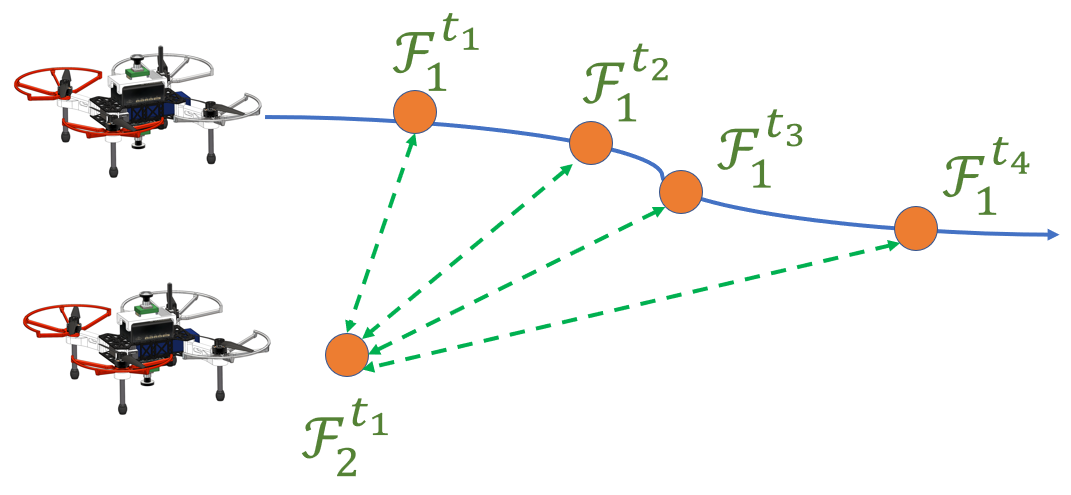}}
    \settowidth\rimage{\includegraphics[height=2cm]{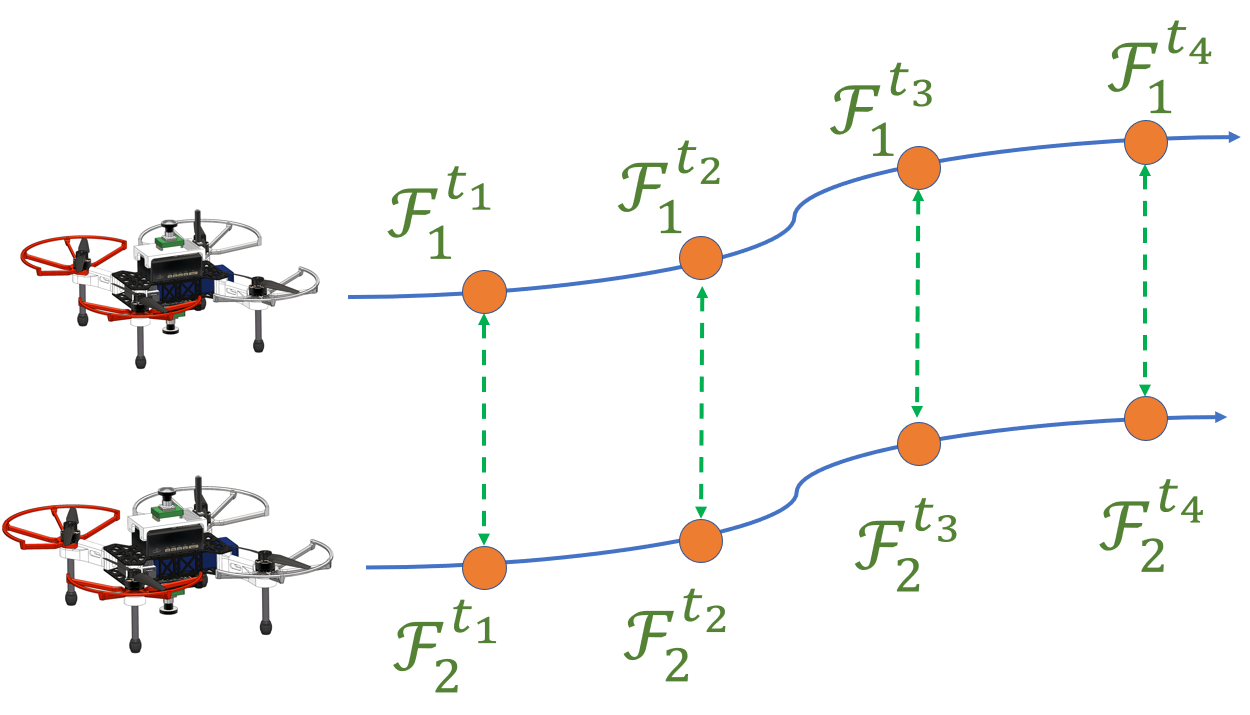}}
    \resizebox{1.0\linewidth}{!}{
        \begin{tabular}{p{\limage}p{\mimage}p{\rimage}}
        \includegraphics[height=2cm]{obs_fly.PNG}
        \subcaption{}\label{fig:obs_a}
        \vspace{-0.5cm}
            &   \includegraphics[height=2cm]{obs_fixed.PNG}
            \subcaption{}\label{fig:obs_b}
        \vspace{-0.5cm}
        &   \includegraphics[height=2cm]{obs_parall.PNG}
        \subcaption{}\label{fig:obs_c}
        \vspace{-0.5cm}
        \end{tabular}
    }

    \caption{\small{
        Observability of UWB-odometry fusion: 
        a) Two drones fly separately and the flight direction is not parallel.
        At this time, the two drones are observable to each other due to the presence of motion.
        b) Drone 1 is flying and drone 2 is in place. 
        At this point, drone 1 can observe the position of drone 2, but the yaw of 2 is not observable. Drone 1 is not observable from drone 2.
        c) The two drones are flying parallel to each other and at this point, the two drones are unobservable to each other.}}\label{fig:obs}
        \vspace{-0.2cm}
\end{figure}

\begin{table}[ht]
    \centering
    \caption{Observability with typical measurements combinations in different scenarios.
    T in the table means the measurement exists.
    F in the table means the measurement is absent.
    T/F in the table indicates that the existence of the measurement does not change the observability.}
    \label{tab:observability}
    \setlength\tabcolsep{2pt}
    \resizebox{\columnwidth}{!}{

    \begin{tabular}{c c c c c c c c}
        \hline
        Typical Scenarios & \begin{tabular}[c]{@{}c@{}}Motion\\ $k$ \end{tabular}& \begin{tabular}[c]{@{}c@{}}Motion\\ $i$ \end{tabular}& UWB & \begin{tabular}[c]{@{}c@{}}Detection\\$k \leftrightarrow i$\end{tabular}  &  \begin{tabular}[c]{@{}c@{}}Map-based\\$k \leftrightarrow i$\end{tabular} & \begin{tabular}[c]{@{}c@{}}Observability\\$k \leftrightarrow i$\end{tabular} \\ \hline
        Feature-rich &  T/F &  T/F &  T/F &  T/F &  T & 6-DoF  \\ \hline
        \multirow{4}{*}{Feature-poor} 
        & F & F & T/F & F & F & No \\
        & T/F & T/F & T/F & T & T/F & 6-DoF  \\
         & T & F & T & F & F & 3-DoF \\
         & T & T & T & T/F & T/F & 6-DoF\\ \hline
        \end{tabular}
    }
\end{table}

One of the key features of this system is the fusion of several sensors with completely different characteristics, and this multi-sensor fusion can effectively improve the observability in various environments. 
In feature-rich scenarios  (usually narrow indoor environment), the rich features allow us to take full advantage of the map-based localization for state estimation.
The inter-drone map-based measurements and visual detection measurements provide direct 6-DoF observability between drones, as the first and third rows in Table \ref{tab:observability} show.

However, map-based localization is challenging to perform in feature-poor scenarios,  e.g., open environments such as stadiums, lawns, and farmland, where there are few features, and visual detection measurements are limited by the visual detection range.
In such scenarios, UWB-odometry fusion can be utilized for state estimation instead.
The necessary condition for the observability of the relative state estimated by UWB-odometry fusion is the presence of relative motion, i.e., an aerial swarm stationary on the ground or flying in parallel does not satisfy this condition. 
As shown in Fig. \ref{fig:obs}, when drone 1 in the factor graph has sufficient motion and drone 2 is fixed, drone 1 can estimate the position of drone 2.
However, the yaw angle of drone 2 cannot be estimated.
Only when both drone 1 and 2 in the graph have motion and there is relative motion can drone 1 and 2 estimate each other's 6-DoF pose with UWB-odometry fusion.
In practice, this brings a considerable level of annoyance.
For example, aerial swarms are likely to remain in fixed-formation flight for long periods, in which case relative motion does not exist.
In Omni-swarm, this issue is solved by fusing the relative pose given by visual drone tracking.

In summary, Table \ref{tab:observability} shows the observability of Omni-swarm in various scenarios.
We find that, by incorporating multiple sensors and measurements, Omni-swarm is capable of maintaining observability in various complex environments.

\subsection{Initialization}\label{sect:init}
Omni-swarm independently tracks the observability of each drone in the $\mathcal{D}_u^k$ for initialization.
Depending on the observability, as given in Table \ref{tab:observability}, Omni-swarm has various ways of performing initialization:
1) UWB-odometry initialization with sufficient motion;
2) map-based measurements for initialization;
and 3) anonymous visual detection measurements for initialization. 
A special case is drone $k$ itself, whose state is initialized by its VIO.

\subsubsection{Map-based measurements for initialization}\label{sect:init_map}
One of the biggest advantages of map-based localization is that only one measurement is needed to provide sufficient observability, as shown in the first row of Table \ref{tab:observability}.
In this case, when a map-based measurement ${\mathbf{z}_\mathcal{L}}_{i\rightarrow j}^{t_0\rightarrow t_1}$, where $i \in \mathcal{D}_e^k$ is an estimated drone and $j\in\mathcal{D}_u^k$ is an uninitialized drone, is detected by the map-based localization module, the system immediately initializes the drone, and we use this map-based measurement and the ego-motion of drone $i$ to initialize its poses in $\mathcal{G}$:
$$
\tensor*[^{v_k}]{\mathbf{\hat P}}{_{j}^t} \leftarrow \tensor*[^{v_k}]{\mathbf{\hat P}}{_{i}^{t_0}}
\ {\mathbf{z}_\mathcal{L}}_{i\rightarrow j}^{t_0\rightarrow t_1}
\ \left( (\mathbf{\tilde P}_{j}^{t_1})^{-1}
\ \mathbf{\tilde P}_{j}^{t}\right),
$$
where $\tensor*[^{v_k}]{\mathbf{\hat P}}{_{j}^t}$ is the state to be initialized in $\mathcal{G}$.

\subsubsection{Anonymous Visual Detection Measurements for Initialization}\label{sect:init_det}
Similar to map-based measurements, the visual detection measurements also contain 6-DoF information, which provides sufficient observability to initialize the state estimation, as shown in the third row of Table \ref{tab:observability}.
The difference is that the target ID of these visual detection measurements is anonymous when the system is not fully initialized.
Therefore, we need to associate the anonymous IDs with the IDs of the uninitialized drones $\mathcal{D}_u^k$ first, and then initialize the state estimates of the matched drone using the same method as in Sect. \ref{sect:init_map}.

The data association method with anonymous visual detection measurements in Omni-swarm is a well-pruned depth-first search (DFS) \cite{zhou1993multitarget, nguyen2019vision}.
This method was first proposed by Zhou et al.\cite{zhou1993multitarget} for multi-target tracking and was extended by Nguyen et al. \cite{nguyen2019vision} to the case where multiple homogeneous drones in an aerial swarm detect each other with anonymous relative measurements.
We do not expand the details of this DFS algorithm here due to the space limitation.

\subsubsection{UWB-odometry for Initialization}
Unlike the previous two cases, initialization with the UWB-odometry requires a combination of measurements in multiple frames to provide sufficient observability, as shown in the fourth and fifth rows of Table \ref{tab:observability}.
The initialization condition for drone $i$ is that there exists enough motion of drone $i$ in the swarm keyframes where both ego-motion and UWB measurements of drone $i$ and drone $k$ exist, i.e., drone $i$ has at least 3-DoF observability by drone $k$.
In practice, we require the motion of drone $i$ to be larger than a bounding box of $[1.5, 1.5, 0.8]$ m.
% Here we choose three initializations because it is a tradeoff between performance and avoiding local optima. 
This bounding box is chosen so that the system has a sufficient baseline length.

In contrast to the previous cases, we cannot obtain an initial estimate of the states directly from the UWB-odometry measurement. 
The initialization with UWB-odometry fusion is accomplished by solving Eq. (\ref{eq:opti_eq}).
To avoid the problem of the local optimal in this process, we will set three different random values for initializing the state of drone $i$ and choose the one with the lowest cost as result of the optimization.
Specifically, the poses of the other drones are initialized with the values of x, y positions randomly distributed in $[-5, 5]$ m and z position randomly distributed in $[-1, 1]$ m.
The initialized value of yaw is equal to the yaw estimated by VIO.
This scale corresponds to the possible distribution of an aerial swarm of 10-20 drones in the takeoff state.
This procedure can be accomplished within within 15 ms on a PC and 100 ms on the onboard computer. 

A point worth noting is that this UWB-odometry fusion cannot avoid the unobservability in the case in Fig. \ref{fig:obs}c; i.e., if the flight paths of the two drones are parallel during the UWB-odometry initialization, the method may lead to incorrect results.
Nevertheless, with other measurements in the Omni-swarm, this problem is very unlikely to occur. 
This is one advantage of Omni-swarm over other UWB-odometry fusion methods.

\subsubsection{Initialization Procedure}

When a new drone starts up and starts sending its measurements, Omni-swarm adds it to the set of available drones $\mathcal{D}_a^k$ and set of uninitialized drones $\mathcal{D}_u^k$. 
The observability condition shown in Table \ref{tab:observability} of each drone in $\mathcal{D}_u^k$ is checked before each optimization.
Once a drone $i$ in $\mathcal{D}_u^k$ is found to satisfy at least 3-DoF observability, the initialization approaches corresponding to the available measurements will be adopted to initialize its states.

\section{System Implementation}
To fully demonstrate the Omni-swarm method proposed in this paper in real-world experiments, we present an aerial swarm system containing drones, algorithms, software management, a communication network, and 3D user interface.
We also develop an aerial swarm trajectory planning method for verifying the practical value of Omni-swarm in inter-drone collision avoidance experiments.

\subsection{Aerial Platform}

\begin{figure}[t]
    \centering
    % \begin{subfigure}{\textwidth}
    \centering
    \includegraphics[width=0.7\linewidth]{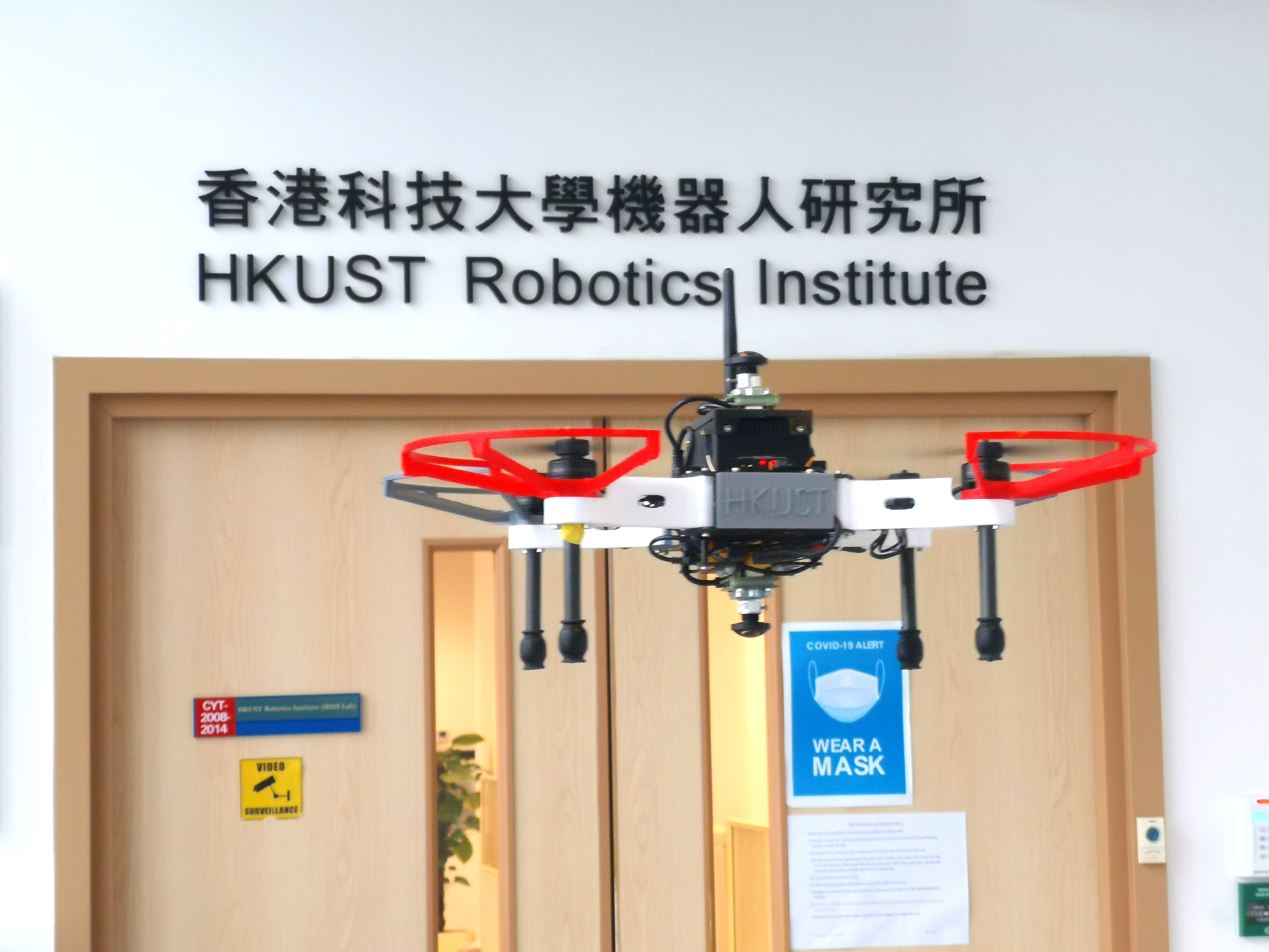}
    \caption{\small{One of the aerial platforms in the swarm system, which is equipped with stereo fisheye cameras, a DJI N3 flight controller, a Nooploop UWB module, and a DJI Manifold2-G onboard computer with an Nvidia TX-2 chip.}}\label{fig:drone}
    \vspace{-0.5cm}
\end{figure}

Our aerial swarm contains a few homogeneous custom drones, as shown in Fig. \ref{fig:drone}. 
The drone is equipped with a DJI N3 flight controller, two Pointgrey cameras with fisheye lenses, a NoopLoop UWB module, a DJI Manifold 2-G onboard computer with an Nvidia TX2 Module, and a WiFi module for communication.
The front-end of VINS-Fisheye is accelerated with CUDA, and the various CNNs used for state estimation are accelerated with TensorRT.
On the TX2, the TensorRT accelerated CNNs are two times faster than using TensorFlow and PyTorch and it takes up less memory.
The computational performance will be detailed in Sect. \ref{sect:timing}

To facilitate the distribution of all the algorithms, including state estimation and planning, we divide the software running on the onboard computer into two layers: the boot layer running on the operating system and the algorithm layer running in the docker container.
Omni-swarm together with the trajectory planning method, is deployed by a docker image and individually runs on each drone.
The docker images are distributed through our wireless ad hoc network and each drone automatically pulls updates to improve the development efficiency. 
An additional computer serves as the maintenance server of the aerial swarm to facilitate the distribution of the docker image, which is not essential for flight.
After we update the code, we may quickly get every drone's docker image up to date by simply booting up the drone within the operating range of the maintenance server's wireless ad hoc network.

\setlength{\textfloatsep}{\textfloatsepsave}% Recover \textfloatsep

\subsection{Frequency and Synchronization}\label{sect:freq}
Different measurements have different frequencies, which generally depend on the nature of the sensor and its computing power. 
In practice, we set the acquisition frequency of the camera to 20 Hz (for collecting datasets) or 16 Hz (for inter-drone collision avoidance experiments), IMU to 400 Hz and UWB to 100 Hz for distance measurements.
VIO uses the IMU to predict the current ego-motion, so its output is also 400 Hz.

Timestamp synchronization is another vexing problem in robot swarms.
The UWB module completes mutual timestamp synchronization during ranging and communication, and sends the synchronized timestamps to the onboard computer via the serial port. 
Therefore, we choose the UWB timestamp as the time reference to obtain a swarm-wide synchronized timestamp.

To cope with the different frequencies and the time differences between different measurements, we take the timestamps of UWB measurements and convert swarm keyframes and all the other measurements to these timestamps.
For VIO, this conversion process is achieved by finding the nearest UWB timestamp.
Since VIO has 400 Hz output, the error caused by the difference between the VIO timestamp and the nearest UWB timestamp is small.
For visual target measurement and map-based measurement, we use ego-motion to convert their relative pose by
\begin{equation} \label{eq:cvt_t}
    {\mathbf{z}_\mathcal{RP}}_{i\rightarrow j}^{t_a\rightarrow t_b} \leftarrow 
    \ \left( (\mathbf{\tilde P}_{i}^{t_a})^{-1}
    \ \mathbf{\tilde P}_{i}^{t_0}\right)
    \ {\mathbf{z}_\mathcal{RP}}_{i\rightarrow j}^{t_0\rightarrow t_1}
    \ \left( (\mathbf{\tilde P}_{j}^{t_1})^{-1}
    \ \mathbf{\tilde P}_{j}^{t_b}\right),
\end{equation}
where ${\mathbf{z}_\mathcal{RP}}_{i\rightarrow j}^{t_0\rightarrow t_1}$ is the relative pose measurement, $t_a$ and $t_b$ are the timestamps of swarm keyframes, which are nearest to $t_0$ and $t_1$.
Eq. (\ref{eq:cvt_t}) converts the ${\mathbf{z}_\mathcal{RP}}_{i\rightarrow j}^{t_0\rightarrow t_1}$ to a relative pose measurement between timestamp $t_a$ and $t_b$.
The accuracy loss caused by this conversion process is also negligible due to the good local accuracy of VIO.

Finally, Omni-swarm's graph-based optimization is solved at a speed of 1 Hz and is forward propagated through VIO at 100 Hz.
Table \ref{tab:freq} shows the computational frequency of the main components in Omni-swarm.
Compared to on a PC, we reduce the computational frequency of the Omni-swarm on the onboard computer so that Omni-swarm has sufficient real-time performance in real-world applications.
This is shown in  the bottom row of Table \ref{tab:freq}.
In addition, on the onboard computer, the position control frequency is 50 Hz.
\begin{table}[]
    \centering
    \caption{\small{The computational frequency of each component on the PC and onboard computer.
            In the table, VINSF and VINSB is the frontend and back-end of VINS-Fisheye, respectively.
            Det. is the visual detection (YOLO) in VDT, and Trk. is the visual tracking (MOSSE). 
        Desc. is the descriptors extraction (NetVLAD and SuperPoint) in MDML.
        LoopDet. is the loop closure detection in MDML, including database retrieval and brute-force matching.
        Opti. is the graph-based optimization.
        $n$ in the table is the drone number of the swarm. The units in this table are Hz.}}
    \label{tab:freq}
    \resizebox{1.0\linewidth}{!}{
    \begin{tabular}{c|ccccccc}
    \hline
    Plat & VINSF & VINSB & Det. & Trk. & Desc. & Loop. Det. & Opti. \\ \hline
    PC & 20 & 10 & 2 & 20 & 0.3 & $0.3*n$ & 10 \\ \hline
    Onboard & 16 & 8 & 1 & 10 & 0.3 & $0.3*n$ & 1 \\ \hline
    \end{tabular}}
    \vspace{-0.2cm}
\end{table}

\subsection{Redundant Computations}
The computations involved in the Omni-swarm can be classified into two types according to whether they are computed redundantly or not:
1) Distributed computations: VIO, visual target tracking, and map-based localization,  which are not duplicated computed on different drones.
% 2) Partial redundant computations: PCM-based outlier rejection, these tasks may be redundant compute by a pair of drones twice.
2) Redundant computations: Graph-based optimization of (Eq. (\ref{eq:opti_eq})) is redundant when computed on different drones.
We choose to solve this optimization on each drone individually to allow the state estimation to be robust to temporary network failure or single-point failure.
It also avoids the communication overhead induced by distributed optimization methods \cite{choudhary2017distributed, ziegler2021distributed}.

\subsection{Network Setup}
The latency, bandwidth, interference immunity and working range of the communication directly determine if the proposed method will work well in practice.
In the real-world experiments, the drones need to exchange image descriptors and flight paths, requiring a much larger communication bandwidth.
In this case, we use a wireless ad hoc network based on the independent basic service set (IBSS)  configuration of IEEE 802.11 standard\cite{crow1997ieee} to meet the communication requirements, which cover the task execution system and maintenance requirements.
One advantage of the wireless ad hoc network is its decentralized nature. 
No central server or router is required for deploying the wireless ad hoc network.
With a 5.2 Ghz wireless ad hoc network, our communication network can cover a sufficient area in our experiments and provide high bandwidth and low latency communication services.
In our laboratory environment, this network reached a 4.9 MB/s bandwidth at a distance of 22.4 m.

We use lightweight communications and marshalling (LCM)\cite{huang2010lcm} to achieve efficient data encoding and communication broadcast in the ad hoc wireless network.
The communication broadcast is limited to one-hop neighbors, covering all the swarm members in our current experiment.
Some of the transmissions with lower bandwidth requirements are simultaneously backed up by UWB.
While this network architecture is not part of the proposed state estimation, it accelerates our development and debugging process.

\subsection{Communication Bandwidth Requirements}
\begin{table}[]
    \centering
    \caption{\small{The broadcasting bandwidth requirements of Omni-swarm for each drone.
    Odom \& UWB is the odometry and UWB distances measurements.
    Keyframe represents the keyframe information for map-based localization.
    Inliers are the broadcast PCM result for each other drone in the swarm.}}
    \label{tab:band}
    \begin{tabular}{c|ccc}
    \hline
    Item        & Freq     & Size   & Bandwidth \\ \hline
    Odom \& Dis & 100  Hz  & 49 Bytes & 4.90   \\ \hline
    Keyframe    & 0.3  Hz  & 180 kB  & 54.0 kB/s  \\ \hline
    Inliers     & 1    Hz  & 552 Bytes   & 0.47 kB/s  \\ \hline
    Sum         & \multicolumn{3}{c}{59.4 kB/s}   \\ \hline
    \end{tabular}
    \vspace{-0.2cm}
\end{table}

\begin{table}[]
    \centering
    \caption{\small{Packet loss rate when there are different numbers of drones in the field.}
    }
    \label{tab:loss}
    \begin{tabular}{c|ccc}
    \hline
    Number & 2 & 3 & 4 \\ \hline
    Avg. Loss. & 3.5\% & 4.7\% & 3.9\% \\ \hline
    \end{tabular}
    \vspace{-0.5cm}
\end{table}

Table \ref{tab:band}  summarizes the major bandwidth of data broadcasting to the swarm by Omni-swarm for each drone, where keyframe takes the biggest proportion.
% We do not list some negligible communications in Table \ref{tab:band}.
The major bandwidth is consumed on broadcasting landmark descriptors generated by the SuperPoint.
In Omni-swarm, we compress the landmark descriptors (averaging 696 per keyframe) from 256 dimensions to 64 dimensions using PCA dimensionality reduction, which has no significant impact on the feature matching but reduces the communication amount by 534.5 kB (three times to current) per keyframe.
We send each landmark descriptor separately in broadcasting the keyframe for map-based localization. 
Losing some of landmark descriptors does not affect the loop closure detection.

Table \ref{tab:loss} shows the packet loss rate with different numbers of drones in the field in the actual test. 
This result indicates that the packet loss rate does not change significantly with the current number of drones, which shows that Omni-swarm is scalable in the network.
According to the available bandwidth in the lab environment (4.9 MB/s), the ideal limitation of the size of the aerial swarm given by the ad hoc network is 83.

\subsection{Trajectory Planning for Swarm}\label{sect:planning}
To demonstrate the capability of our state estimation method, we conduct fully autonomous formation flights and inter-drone collision avoidance tests.
In these flights, trajectories of the drones are generated by the method extended from Fast-Planner \cite{zhou2019robust}, a real-time trajectory planner based on kinodynamic search and gradient-based optimization.  

The original Fast-Planner \cite{zhou2019robust} assumes a static environment and does not consider the interaction with other agents. Therefore, we adapt it to achieve decentralized collision avoidance.
Among all the drones, trajectories are shared through the wireless ad hoc network.
Each drone checks whether any conflict exists in its own trajectory and that of others.
Whenever a conflict is found, it replans a new trajectory immediately.
The replanning starts by searching a safe and dynamically feasible initial path, in which motion primitives conflicting with other drones' trajectories are pruned to avoid an inter-drone collision.
Collision avoidance is achieved by properly designing the collision penalty function in the trajectory optimization, following recent literature \cite{zhou2021ego, zhou2021raptor}. If a collision happens, the penalty term grows rapidly, which quickly dominates the overall optimization objective function. 
The process is repeated for each drone independently until it reaches the designated goal.
In this process, all computations, including trajectory planning, state estimation, and flight control, are done on the onboard computer.

\section{Experiments}\label{sect:experiment}
To validate Omni-swarm's feasibility, accuracy, and practical value, we run a series of experiments: 
1) timing and scalability of Omni-swarm on a PC and TX2 onboard computer;
2) experiments to verify the features of Omni-swarm;
3) accuracy comparison with the ground truth on a custom dataset with various other approaches;
4) verification of the global consistency on an outdoor dataset;
5) real-time performance verification in an inter-drone collision avoidance experiment.

\subsection{Scalability}
It should be clear that the computation effort scale of the Omni-swarm's front-end is not significant with the number of drones. This is because most components in the front-end are running at a fixed frequency, which is not relevant to the scale of the drone.
There are two special cases: 1) VDT's pose estimation is computed separately for each detected drone. 
It is related to the number of drones in the FoV. 
In practice, it is unlikely that there will be too many drones in the FoV, so the scalability issue here is not severe.
2) Loop closure detection requires the Faiss database to retrieve the keyframe. 
% The larger the scale of the aerial swarm,  the larger the number of keyframes in the databases. 
Faiss has the complexity of $O(k\ log(l/w))$ for KNN searching \cite{JDH17}, where $k=5$ is the selection number each time,  $w=32$ is a parameter of Faiss, $l$ is the size of the keyframes in the database, which can be considered as linear to the swarm scale.
In practice, the computation time that grows logarithmically with the scale of the swarm does not introduce scalability difficulties.

Next, we discuss the scalability of the Omni-swarm's back-end. 
Here we assume that measurements are consistent between any two drones.
Two main computational components are involved in the back-end: 
1) PCM outlier rejection:
The drone only needs to compute the PCM problem with another drone using relative pose measurements making the complexity of this algorithm $O(n)$,  where $n$ is the swarm scale. 
2) Graph-based optimization:  The scale of Problem (\ref{eq:opti_eq}) grows quadratically with the swarm scale, so the computational effort of graph-based optimization is $O(n^2)$
\begin{figure}[t]
    \centering
    \centering
    \includegraphics[width=0.8\linewidth]{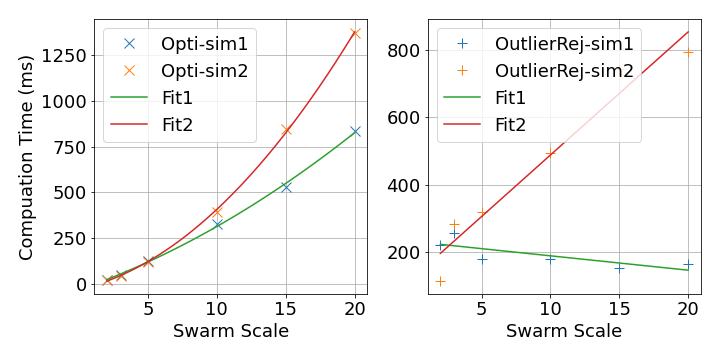}
    \vspace{-0.3cm}
    \caption{\small{
        A demonstration of the growth of computational time with different swarm scales in the simulation of two scenarios.
        The left figure shows the computation time of graph-based optimization (symbol 'x' ) and fitted quadratic splines.
        The right figure shows the computation time of PCM outlier rejection (symbol '+' ) and fitted lines.
        Suffix 1 and 2 in the legends correspond to Scenario 1 and Scenario 2.
    }}\label{fig:scale}
    \vspace{-0.6cm}
\end{figure}

To verify the conclusions here, we use simulations to simulate the complexity of the back-end computation. Aerial swarms of scales from 2 to 20 are set up in the simulations and are tested under different scenarios: 1) The drones in the swarm are lined up, each drone can only generate relative pose measurements with several neighboring drones, but all drones have distance measurements between them.
2) The drones in the swarm are closely grouped in a grid formation, and all drones can generate relative pose measurements with each other.

The results of the simulation are shown in Fig. \ref{fig:scale}. 
It can be seen that the quadratic curves perfectly fit the computation for graph-based optimization in both scenarios. 
While in Scenario1, the average computation time of PCM outlier rejection is not significantly related to the swarm scale because the consistency graph that needs to be computed for each drone in Scenario 1 only depends on the number of neighbors (2-4). 
In Scenario2, PCM outlier rejection takes linearly increasing time because all drones have relative pose measurements with each other.
% The data shown in Fig. \ref{fig:scale} is consistent with our analysis above.

\subsection{Features Verification}
\subsubsection{Fast initialization and plug-and-play}\label{sect:plugandplay}
In this experiment, drone 1 and 2 are firstly booted up in the field, with drone 1 carrying a battery in poor condition to simulate a real emergency.
Omni-swarm is properly initialized after the program is loaded and the two drones take off.
Drone 3 is then placed in the field and powered on, at which point drone 1 is automatically landed by the onboard computer's control system due to its lack of power.
This action does not affect Omni-swarm's estimation, and the program on drone 3 is then successfully loaded and correctly estimated by drone 1.
Drone 3 establishes a correct estimation of 1 and 2 at the same time.
Next, we launch drone 3 and then remove the battery of drone 1 and remove it from the field.
We then add drone 4 in the field, and after it is estimated by drone 3, we launch drone 4 and let drone 2 go around drones 3 and 4  to the back of the field with the inter-drone collision avoidance algorithm.
Finally, we land drone 2 and shut it down, and the state estimation of the remaining drones 3 and 4 works normally.

\subsubsection{Non-line-of-sight condition flight experiment}
In this experiment, the two drones are separated by a short wall at the beginning. 
Then, the two drones take-off and fly around the wall anti-clockwise. 
During the flight, the drones are not in each others' line-of-sight. 
We disable the UWB measurement in Omni-swarm since the UWB modules can still measure distances crossing the foam obstacles between the drones.
Once a drone visits the other drone's starting place, Omni-swarm immediately initializes, and the state of the two drones is estimated correctly. 
This experiment verifies that the proposed method works even in totally non-line-of-sight cases if the same location has been revisited. 

Recording of these two experiments can be found in our video\footnote{https://www.youtube.com/watch?v=SMtJUkKoza4}.

\subsection{Evaluation on Datasets}
\begin{figure*}[ht]

    \centering
    \settowidth\limage{\includegraphics[height=3cm]{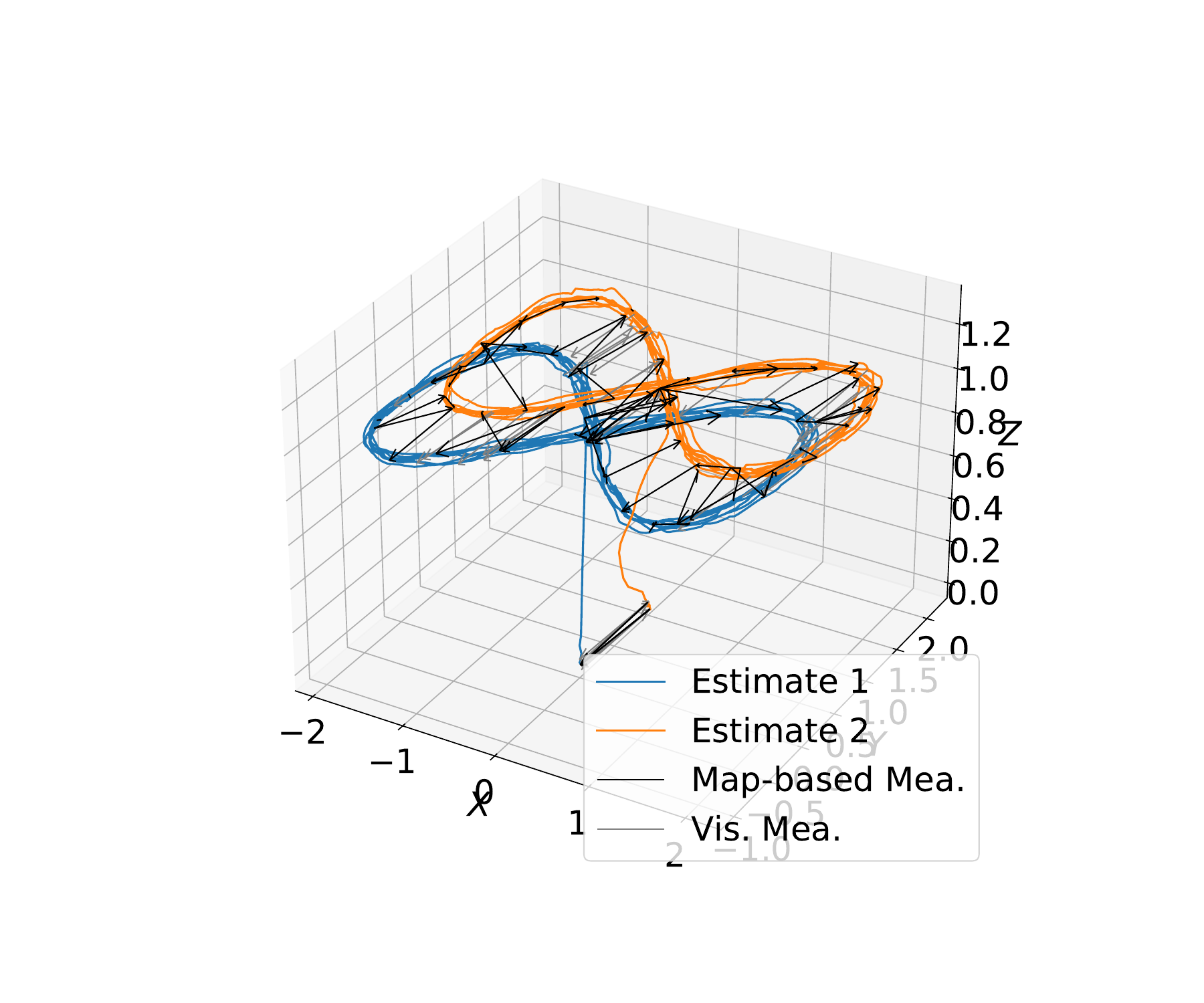}}
    \settowidth\mimage{\includegraphics[height=3cm]{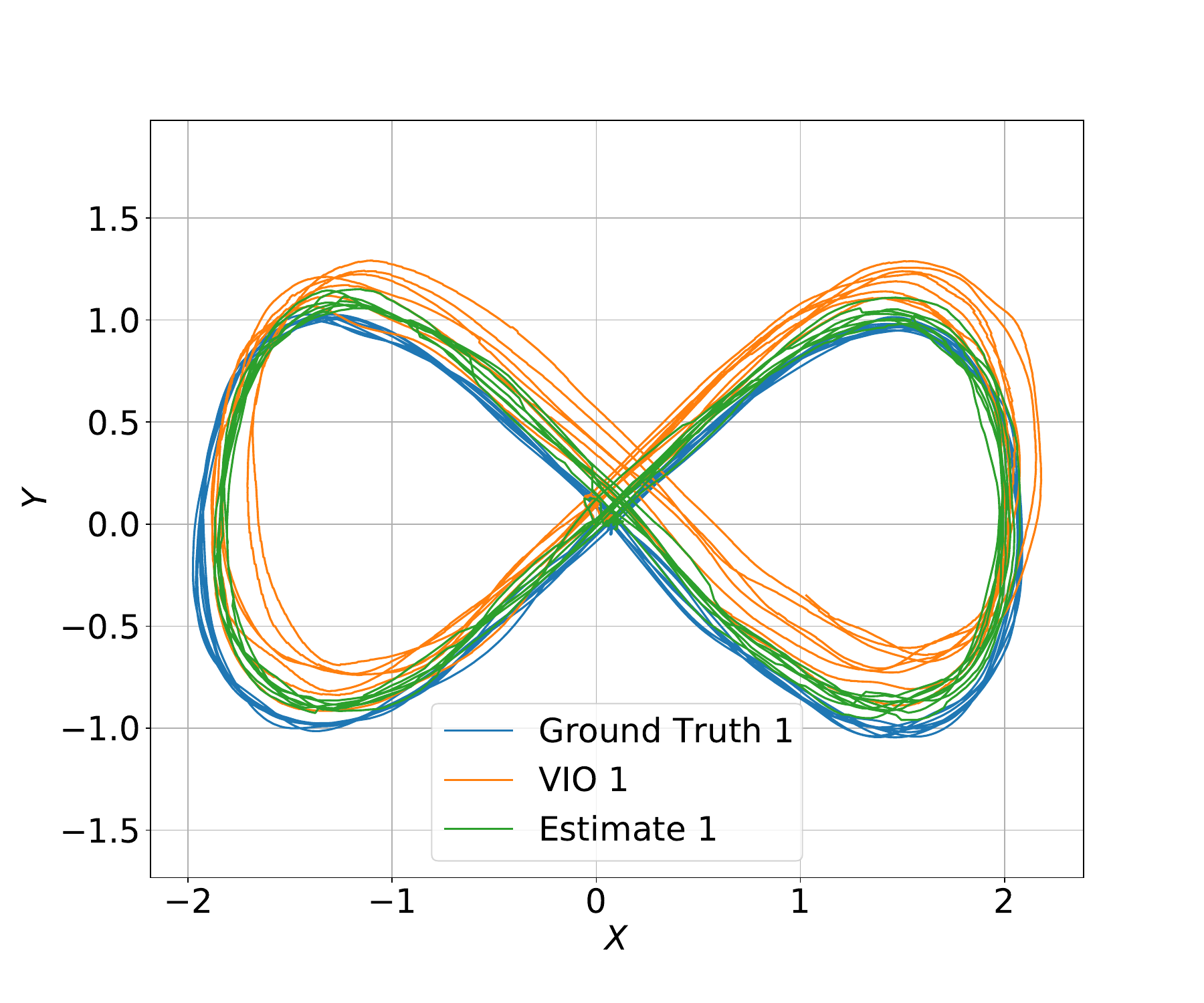}}
    \settowidth\rimage{\includegraphics[height=3cm]{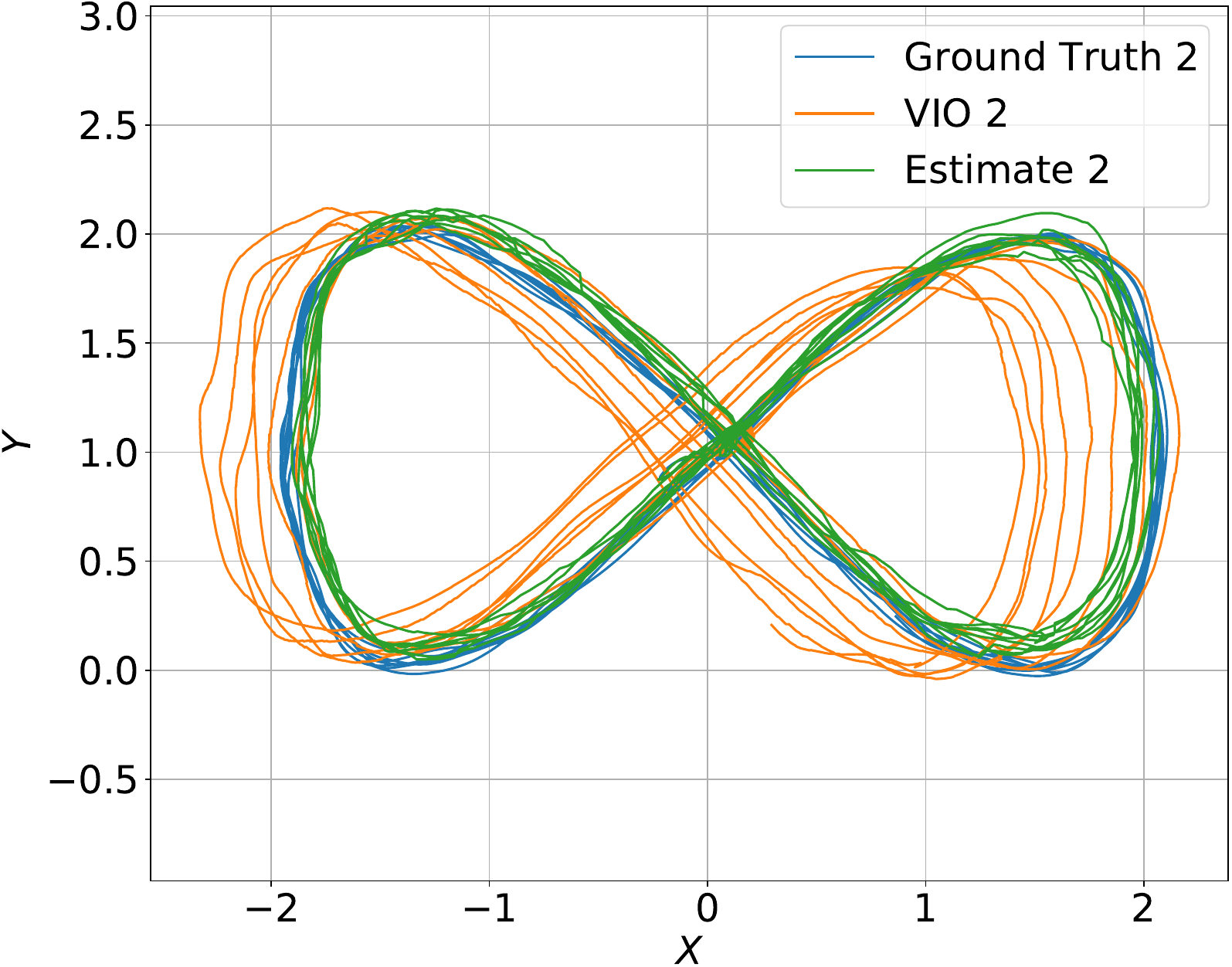}}
    \resizebox{0.7\linewidth}{!}{
        \begin{tabular}{p{\limage}p{\mimage}p{\rimage}}
    \includegraphics[height=3cm]{Traj2}\newline
    \subcaption{}\label{fig:ex1_traj}
    \vspace{-0.5cm}
        &   \includegraphics[height=3cm]{fusedvsgt2d_1}\newline
    \subcaption{}\label{fig:ex1_com}
    \vspace{-0.5cm}
    &   \includegraphics[height=3cm]{fusedvsgt2d_4}\newline
    \subcaption{}\label{fig:ex1_com2}
    \vspace{-0.5cm}
    \end{tabular}
    }
    \caption{\small{ The estimated trajectories of the two drones in an indoor dataset. (a): The black arrows are the map-based measurements. The gray arrows are the visual detection measurements. Only part of the measurements are shown in this figure for better visualization. (b)-(c): The comparison of the ground truth trajectories (blue), the VIO trajectories (orange), and the estimated trajectories (green). The VIO trajectories drift away from the ground truth trajectories, while the estimated trajectories closely follow the ground truth.
    }}\label{fig:ex1}
    \vspace{-0.2cm}
\end{figure*}

We collect indoor and outdoor datasets for accuracy comparison. 
The datasets include the raw sensor data and ground truth given by a motion capture system in the indoor laboratory and outdoor environment without ground truth.
The indoor datasets include:
1) two double-drone-fixed-heading formation flight datasets;
2) a double-drone formation flight dataset with the drones heading towards the velocity direction; 
3) a double-drone-random-target flight dataset.
The details of the datasets are shown in Table \ref{tab:datasets}.
\begin{table}[]
    \centering
    \caption{\small{
        The datasets we use for accuracy comparison.
        Traj. Len. is the trajectory length of different drones in the aerial swarm.
        Takeoff time is the average time for aircraft takeoff in different datasets.
        }}\label{tab:datasets}
    \begin{tabular}{c|c|c|c|c|c}
    \hline
    Dataset              & Scale & \begin{tabular}[c]{@{}c@{}}Avg. Traj. \\ Len. (m)\end{tabular} & \begin{tabular}[c]{@{}c@{}}Takeoff \\ Time (s)\end{tabular} & Type       & Envir.  \\ \hline\hline
    \textbf{Parallel1}   & 2     & 93.0                                                           & 21.5                                                        & Real-world & Indoor  \\ \hline
    \textbf{Parallel2}   & 2     & 106.4                                                          & 41.3                                                        & Real-world & Indoor  \\ \hline
    \textbf{Parallel3}   & 2     & 130.7                                                          & 39.1                                                        & Real-world & Indoor  \\ \hline
    \textbf{RandFlight}  & 2     & 45.5                                                           & 57.8                                                        & Real-world & Indoor  \\ \hline
    \textbf{Outdoor1}    & 2     & 237.2                                                          & 31                                                          & Real-world & Outdoor \\ \hline
    \end{tabular}
    \vspace{-0.5cm}
\end{table}

\subsubsection{Metric Definitions} \label{sect:metrics}
In this paper, two metrics are involved, absolute trajectory error (ATE) and relative error (RE), which are used to measure the global consistency and relative estimation accuracy, respectively, of Omni-swarm.
The definition of ATE used in this paper is the same as that in \cite{zhang2018tutorial}.
A point worth noting is that the drift of each trajectory and the relative state estimation error between trajectories estimated by the state estimation are included in the ATE, i.e., the ATE measures the global consistency and the relative localization accuracy of the swarm.

The RE is used to characterize the relative state estimation accuracy, which is defined as 
\begin{equation}
    \begin{aligned}
    RE_{rot_k}^i &= \left(\frac{1}{N}\sum_{t=0}^{N-1} \Vert \angle\left(\tensor*[^{v_k}]{\mathbf{\hat R}}{_{k}^t}^T \tensor*[^{v_k}]{\mathbf{\hat R}}{_{i}^t}\right) \Vert^2 \right)^{\frac{1}{2}} \\
    RE_{pos_k}^i &= \left(\frac{1}{N}\sum_{t=0}^{N-1} \Vert \left((\tensor*[^{v_k}]{\mathbf{\hat R}}{_{k}^t}^T (\tensor*[^{v_k}]{\mathbf{\hat T}}{_{i}^t}- \tensor*[^{v_k}]{\mathbf{\hat T}}{_{k}^t})\right) \Vert^2 \right)^{\frac{1}{2}}, \\
    RE_{pos_{k_{axis}}}^i &= \left(\frac{1}{N}\sum_{t=0}^{N-1} \vert \left((\tensor*[^{v_k}]{\mathbf{\hat R}}{_{k}^t}^T (\tensor*[^{v_k}]{\mathbf{\hat T}}{_{i}^t}- \tensor*[^{v_k}]{\mathbf{\hat T}}{_{k}^t})\right)_{axis} \vert^2 \right)^{\frac{1}{2}},
    \end{aligned}
\end{equation}
where $RE_{rot_k}^i$ and  $RE_{pos_k}^i$ are the rotation and translation relative error, respectively, of drone $i$ estimated by drone $k$, $N$ is the total length of the trajectory, and $\angle(\cdot)$ is the angle of the rotation error, which is defined in \cite{zhang2018tutorial}.
$axis\in\{x,y,z\}$,
$RE_{pos_{k_x}}^i, RE_{pos_{k_y}}^i, RE_{pos_{k_z}}^i$ are the translation relative errors on the $x, y$ and $z$ axis, which helps the reader to establish a more precise perception of the error.
Due to the space limitation, we use $RE_{pos_k}^i$ in the dataset comparison and only use per-axis errors for inter-drone collision avoidance in real-world environments.
The ATE and RE in this paper are the averaged results over the dataset and the units are in meters and degrees, respectively.

\subsubsection{Accuracy Comparison with Ground Truth}\label{subsect:accuracy_analysis}
\begin{table*}[]
    \centering
    \caption{\small{
            Comparison of swarm state estimation methods on the indoor datasets. Init. time. is the average time for successful initialization of state estimation. 
            The overall best results in ATE and RE are in bold.
        }}
    \label{tab:comp1}
    \setlength{\tabcolsep}{5.5pt}
    \resizebox{\linewidth}{!}{
    \begin{tabular}{c|c|cc|cc|cc|cc|cc|cc|cc|cc}
    \hline
    \multirow{2}{*}{Dataset}             & \multirow{2}{*}{Metrics} & \multicolumn{2}{c|}{\textit{Xu2020}} & \multicolumn{2}{c|}{\textit{PGO}}  & \multicolumn{2}{c|}{\textit{VIO Only}} & \multicolumn{2}{c|}{\textbf{Proposed}} & \multicolumn{2}{c|}{\begin{tabular}[c]{@{}c@{}}\textit{Without}\\ \textit{UWB}\end{tabular}} & \multicolumn{2}{c|}{\begin{tabular}[c]{@{}c@{}}\textit{Without}\\ \textit{Tracking}\end{tabular}} & \multicolumn{2}{c|}{\begin{tabular}[c]{@{}c@{}}\textit{Without}\\ \textit{Map-based}\end{tabular}}& \multicolumn{2}{c}{\begin{tabular}[c]{@{}c@{}}\textit{Without}\\\textit{Outlier Rej.}\end{tabular}} \\ \cline{3-18} 
                                         &                          & Pos           & Rot         & Pos          & Rot        & Pos            & Rot          & Pos                & Rot               & Pos                                  & Rot                                 & Pos                                     & Rot                                   & Pos                                     & Rot                                   & Pos                                     & Rot                                    \\ \hline\hline
    \multirow{2}{*}{\textbf{Parallel1}}  & ATE                      & 0.600         & 8.6$^\circ$        & 0.149        & 2.5$^\circ$       & 0.228          & 2.5$^\circ$         & 0.127              & 2.2$^\circ$     & \textbf{0.126}                       & \textbf{2.2}$^\circ$                                & 0.149                                   & 2.3$^\circ$                                  & 0.160                                   & 2.4$^\circ$                                  & 0.360                                   &2.6$^\circ$                          \\ \cline{2-18} 
                                         & RE                       & 0.959         &17.7$^\circ$        & 0.114        & 3.3$^\circ$       & 0.261          & 2.3$^\circ$         & \textbf{0.062}     & \textbf{2.3$^\circ$}     & 0.063                                & 2.7$^\circ$                                & 0.100                                   & 3.7$^\circ$                                  & 0.062                                   & 2.6$^\circ$                                  &0.064                                    &      3.2$^\circ$                   \\ \cline{2-18} \hline
    \multirow{2}{*}{\textbf{Parallel2}}  & ATE                      & 0.351         & 3.2$^\circ$        & 0.113        & 3.2$^\circ$       & 0.225          & 3.0$^\circ$         & \textbf{0.086}     & \textbf{3.0$^\circ$}     & 0.094                                & 3.1$^\circ$                                & 0.103                                   & 3.0$^\circ$                                  & 0.225                                   & 3.1$^\circ$                                  & 0.119                                   &3.2$^\circ$                          \\ \cline{2-18} 
                                         & RE                       & 0.241         & 4.7$^\circ$        & 0.124        & 4.3$^\circ$       & 0.230          & 4.3$^\circ$         & \textbf{0.063}     & \textbf{4.3}$^\circ$              & 0.083                                & 4.0$^\circ$                                & 0.096                                   & 4.3$^\circ$                                  & 0.069                                   & 4.2$^\circ$                                  &0.069                                    &      4.3$^\circ$                   \\ \cline{2-18} \hline

    \multirow{2}{*}{\textbf{Parallel3}}  & ATE                      & 0.470         & 2.6$^\circ$        & 0.153        & 1.8$^\circ$       & 0.281          & 3.1$^\circ$         & \textbf{0.119}     & \textbf{1.8}$^\circ$              & 0.128                                & 1.8$^\circ$                                & 0.134                                   & 1.7$^\circ$                                  & 0.242                                   & 3.0$^\circ$                                  &       0.535                             &5.1$^\circ$                 \\ \cline{2-18} 
                                         & RE                       & 0.566         & 1.8$^\circ$        & 0.130        & 1.7$^\circ$       & 0.225          & 4.7$^\circ$         & \textbf{0.072}     & \textbf{1.8}$^\circ$     & 0.082                                & 1.8$^\circ$                                & 0.081                                   & 1.8$^\circ$                                  & 0.075                                   & 1.9$^\circ$                                  & 0.143                                   &4.0$^\circ$                     \\ \cline{2-18} \hline
    \multirow{2}{*}{\textbf{RandFlight}} & ATE                      & 0.197         & 2.7$^\circ$        & 0.153        & 1.8$^\circ$       & 0.125          & 2.6$^\circ$         & 0.088              & 2.7$^\circ$     & 0.092                                & 2.7$^\circ$                                & 0.101                                   & 2.7$^\circ$                                  & \textbf{0.079}                                   & \textbf{2.7}$^\circ$                                  & 0.165                                   & 5.9$^\circ$                    \\ \cline{2-18} 
                                         & RE                       & 0.191         & 2.1$^\circ$        & 0.130        & 1.7$^\circ$       & 0.160          & 1.1$^\circ$         & \textbf{0.069}     & \textbf{1.3}$^\circ$     & 0.071                                & 1.2$^\circ$                                & 0.078                                   & 1.4$^\circ$                                  & 0.074                                   & 1.4$^\circ$                                  &0.147                                    &9.8$^\circ$       \\ \cline{2-18} \hline
    \end{tabular}}
    \vspace{-0.2cm}    
\end{table*}

\begin{table*}[]
    \centering
    \caption{
        \small{Comparison of initialization and observability of different methods.
                In the table, Init. Time is the time when the system is initialized after boot-up, 
                Init. Trials is the number of times the system attempted the initialization optimization, Succ. Rate is the success rate of the initialization, 
                and RE is the relative measurement error.
                '-' in the table means the method is not applicable to the dataset.
                }}
    \label{tab:init}
    \resizebox{\linewidth}{!}{
    \begin{tabular}{c|ccccc|ccccc|ccccc}
    \hline
    \multirow{2}{*}{Dataset}                                               & \multicolumn{5}{c|}{\textit{Xu2020}}                                                                                                                                                                                                                                                              & \multicolumn{5}{c|}{\textit{PGO}}                                                                                                                                                                                                                                                                 & \multicolumn{5}{c}{\textbf{Proposed}}                                                                                                                                                                                                                                                    \\ \cline{2-16} 
                                                                           & \multicolumn{1}{c|}{\begin{tabular}[c]{@{}c@{}}Init. \\ Time (s)\end{tabular}} & \multicolumn{1}{c|}{\begin{tabular}[c]{@{}c@{}}Init. \\ Trials\end{tabular}} & \multicolumn{1}{c|}{\begin{tabular}[c]{@{}c@{}}Succ.\\ Rate\end{tabular}} & \multicolumn{1}{c|}{$RE_{pos}$} & $RE_{rot}$ & \multicolumn{1}{c|}{\begin{tabular}[c]{@{}c@{}}Init. \\ Time (s)\end{tabular}} & \multicolumn{1}{c|}{\begin{tabular}[c]{@{}c@{}}Init. \\ Trials\end{tabular}} & \multicolumn{1}{c|}{\begin{tabular}[c]{@{}c@{}}Succ.\\ Rate\end{tabular}} & \multicolumn{1}{c|}{$RE_{pos}$} & $RE_{rot}$ & \multicolumn{1}{c|}{\begin{tabular}[c]{@{}c@{}}Init. \\ Time (s)\end{tabular}} & \multicolumn{1}{c|}{\begin{tabular}[c]{@{}c@{}}Init. \\ Trials\end{tabular}} & \multicolumn{1}{c|}{\begin{tabular}[c]{@{}c@{}}Succ.\\ Rate\end{tabular}} & \multicolumn{1}{c|}{$RE_{pos}$} & $RE_{rot}$ \\ \hline\hline
    \textbf{Parallel1}                                                     & \multicolumn{1}{c|}{52.1}                                                      & \multicolumn{1}{c|}{3}                                                       & \multicolumn{1}{c|}{0\%}                                                  & \multicolumn{1}{c|}{0.959}      & 17.7$^\circ$       & \multicolumn{1}{c|}{1.36}                                                      & \multicolumn{1}{c|}{1}                                                       & \multicolumn{1}{c|}{100\%}                                                & \multicolumn{1}{c|}{0.114}      & 3.3$^\circ$        & \multicolumn{1}{c|}{1.29}                                                      & \multicolumn{1}{c|}{1}                                                       & \multicolumn{1}{c|}{100\%}                                                & \multicolumn{1}{c|}{0.062}      & 2.3$^\circ$        \\ \hline
    \textbf{RandFlight1}                                                   & \multicolumn{1}{c|}{84.4}                                                      & \multicolumn{1}{c|}{3}                                                       & \multicolumn{1}{c|}{100\%}                                                & \multicolumn{1}{c|}{0.191}      & 2.1$^\circ$        & \multicolumn{1}{c|}{1.36}                                                      & \multicolumn{1}{c|}{1}                                                       & \multicolumn{1}{c|}{100\%}                                                & \multicolumn{1}{c|}{0.130}      & 1.7$^\circ$        & \multicolumn{1}{c|}{1.43}                                                      & \multicolumn{1}{c|}{1}                                                       & \multicolumn{1}{c|}{100\%}                                                & \multicolumn{1}{c|}{0.069}      & 1.3$^\circ$        \\ \hline
    \textbf{\begin{tabular}[c]{@{}c@{}}Parallel1\\ Outdoor\end{tabular}}   & \multicolumn{1}{c|}{52.3}                                                      & \multicolumn{1}{c|}{3}                                                       & \multicolumn{1}{c|}{0\%}                                                  & \multicolumn{1}{c|}{1.047}      & 3.2$^\circ$       & \multicolumn{1}{c|}{-}                                                         & \multicolumn{1}{c|}{-}                                                       & \multicolumn{1}{c|}{-}                                                    & \multicolumn{1}{c|}{-}          & -          & \multicolumn{1}{c|}{1.31}                                                      & \multicolumn{1}{c|}{1}                                                       & \multicolumn{1}{c|}{100\%}                                                & \multicolumn{1}{c|}{0.057}      & 1.1$^\circ$        \\ \hline
    \textbf{\begin{tabular}[c]{@{}c@{}}RandFlight1\\ Outdoor\end{tabular}} & \multicolumn{1}{c|}{83.1}                                                      & \multicolumn{1}{c|}{3}                                                       & \multicolumn{1}{c|}{100\%}                                                & \multicolumn{1}{c|}{0.260}      & 4.5$^\circ$       & \multicolumn{1}{c|}{-}                                                         & \multicolumn{1}{c|}{-}                                                       & \multicolumn{1}{c|}{-}                                                    & \multicolumn{1}{c|}{-}          & -          & \multicolumn{1}{c|}{1.38}                                                      & \multicolumn{1}{c|}{1}                                                       & \multicolumn{1}{c|}{100\%}                                                & \multicolumn{1}{c|}{0.111}      & 3.0$^\circ$       \\ \hline
    \end{tabular}}
    \vspace{-0.4cm}
\end{table*}

Table \ref{tab:comp1} shows the accuracy comparison on recorded datasets (\textbf{Parallel1, Parallel2, RandFlight}).
\textbf{Parallel1} and \textbf{Parallel2} are the parallel flight tasks and \textbf{RandFlight} is the double-drone-random-target flight datasets task.
The table contains a comparison of four methods: 
1) our previous method in \cite{xu2020decentralized} (named \textit{Xu2020}).
2) An approach with only map-based measurements with VIO (named \textit{PGO}), namely the pose graph optimization method for CSLAM \cite{cunningham2010ddf,cunningham2013ddf,choudhary2017distributed,lajoie2020door}.
3) A \textit{VIO-only} method \cite{lusk2020distributed}. VIO does not have the ability to perform relative positioning, so we use the ground truth to align the coordinate system of different drones. 
4) Our proposed method.
In addition, the table also consists of four ablation studies:
1) remove UWB (named \textit{Without UWB}),
2) remove visual drone tracking (named \textit{Without Tracking}),
3) remove map-based localization (named \textit{Without Map-based}),
and 4) remove the outlier rejection and Huber norm (named \textit{Without Outlier Rej.}).

The proposed method has the best ATE and RE overall, where the estimated trajectories of the proposed method on \textbf{Parallel1} are shown in Fig. \ref{fig:ex1}.
Our method also shows the best performance on the simulated outdoor dataset in Table \ref{tab:init} (the details will be discussed in Sect. \ref{sect:expr_init}), which proves that Omni-swarm has the best adaptation to the environment.
\textit{Xu2020} nearly fails on the \textbf{Parallel1} and \textbf{Parallel3} datasets because it lacks observability in the parallel flight task.
The reason that relative state estimation of \textit{Xu2020} does not fail on \textbf{Parallel2} is that the drone position control errors in \textbf{Parallel2} cause 15 cm of relative motion between the drones, which causes the \textit{Xu2020} approach to obtain an inaccurate estimate with relative motion.
However, the visual detection of \textit{Xu2020} does not function on this dataset due to the FoV limitation. 
This makes the final relative localization accuracy only reach 0.241 m.
The same situation applies to the \textbf{RandFlight} dataset, where UWB-odometry in \textit{Xu2020} establishes a rough relative estimate, but the method does not yield accurate relative measurements due to the FoV limitation.
As a comparison, \textit{Without Map-based} is similar to \textit{Xu2020} but has the new omnidirectional VDT module and initialization, and it has much better relative localization accuracy than \textit{Xu2020}.
This illustrates the significance of the omnidirectional front-end and initialization in this paper.

On all datasets, \textbf{PGO} obtains accuracy in the order of decimeters, but we will show the drawbacks of PGO in outdoor environments later.
The \textit{VIO-only} method is not as accurate as PGO and the proposed methods, and for this method we need to manually align the coordinate system of different drones at the very beginning. 

From the ablation studies \textit{Without UWB} and \textit{Without Tracking}, we can see the contribution of these two types of factors (distance and visual detection factors) to system accuracy. 
In addition, good relative localization accuracy can be obtained without using the map-based localization (\textit{Without Map-based}), which is especially important for feature-poor outdoor environments.
However, the global consistency (ATE) deteriorates to \textit{VIO-Only} in this case.
This comparison demonstrates the importance of map-based localization for global consistency.
Finally, \textit{Without Outlier Rej.} is two to four times less accurate than the proposed method in some datasets, demonstrating the benefits of introducing the outlier rejection module and Huber norm.

\subsubsection{Observablility analysis and initialization}\label{sect:expr_init}

We conduct comparative experiments to verify the improvement in the observability of Omni-swarm relative to previous works.
The experiments give a comparison of Omni-swarm with the \textit{PGO} and \textit{Xu2020} methods in terms of initialization performance under different environments.
The comparison shows the initialization of the indoor environment and the outdoor environment under parallel and random flight.
The comparison results are shown in Table \ref{tab:init}.

We argue that open outdoor scenes do not have sufficient feature points for map-based localization. 
While VIO still works in this case, it is not as accurate as it is in indoor environments. 
On this basis, we use the \textbf{Paralell1} and \textbf{RandFlight} datasets to simulate outdoor datasets with ground truth, named \textbf{Paralell1Out} and \textbf{RandFlightOut} in Table \ref{tab:init}.
We restrict VINS-Fisheye to use ground features and limit surrounding features (no more than 20), and turn off the map-based localization to simulate feature-poor outdoor environments. 

As shown in Table \ref{tab:init}, our proposed method has accurate relative localization on all indoor and outdoor datasets, which means that it is observable in all cases. 
In indoor environments, map-based localization can be used for initialization (Sect. \ref{sect:init_map}), while visual detection initialization (Sect. \ref{sect:init_det}) is used for indoor and outdoor environments.
The \textit{PGO} method also has fair relative localization accuracy indoors but is not applicable to open outdoor environments.
The \textit{Xu2020} method has a fair relative localization accuracy on the dataset with relative motion (\textbf{RandFlight}) in indoor and outdoor cases due to its use of UWB-odometry fusion, which is observable in this case.
However, on the parallel flight dataset (neither indoor or outdoor), the relative localization of this method is considered as having failed with a huge relative localization error.

Table \ref{tab:init} also shows the initialization quantitatively.
The proposed method can be initialized quickly before the aerial swarm takes off in all cases and requires only one optimization.
The \textit{PGO} method can also be quickly initialized in indoor environments with the help of surrounding feature points, but it does not work in outdoor environments.
The \textit{Xu2020} method, on the other hand, does not succeed in initializing the correct state estimate on the parallel flight dataset. 
On the \textbf{RandFlight} dataset, \textit{Xu2020} is able to initialize successfully, but it requires a period after the aerial swarm has flown to complete the process.
This period brings the issue of flight safety.
As a reference, the take-off time of these datasets can be found in Table \ref{tab:datasets}.

\subsubsection{Impact of the random deletion mechanism}\label{sect:randdel}
% Please add the following required packages to your document preamble:
% \usepackage{multirow}
\begin{table}[]
    \centering
    \caption{\small{
            Accuracy comparison of different deletion mechanisms.
            Avg. Time is the averaged solving time of the graph-based optimization.
            }}
    \label{tab:randdel}
    \begin{tabular}{c|c|cc|cc|cc}
    \hline
    \multirow{2}{*}{Dataset}   & \multirow{2}{*}{Metrics} & \multicolumn{2}{c|}{NoDel}   & \multicolumn{2}{c|}{SldWin}  & \multicolumn{2}{c}{\textbf{RandDel}} \\ \cline{3-8} 
                               &                          & Pos           & Rot          & Pos            & Rot         & Pos               & Rot              \\ \hline\hline
    \multirow{3}{*}{Parallel1} & ATE                      & 0.091         & 3.2$^\circ$         & 0.153          & 3.3$^\circ$        & 0.086             & 3.0$^\circ$             \\ \cline{2-8} 
                               & RE         & 0.066         & 4.4$^\circ$         & 0.067          & 4.4$^\circ$        & 0.063             & 4.3$^\circ$             \\ \cline{2-8} 
                               & Avg Time.                & \multicolumn{2}{c|}{21.00ms} & \multicolumn{2}{c|}{12.11ms} & \multicolumn{2}{c}{14.36ms}          \\ \hline
    \multirow{3}{*}{Parallel2} & ATE                      & 0.115         & 1.8$^\circ$         & 0.175          & 2.3$^\circ$        & 0.119             & 1.8$^\circ$             \\ \cline{2-8} 
                               & RE         & 0.072         & 1.8$^\circ$         & 0.073          & 1.8$^\circ$        & 0.072             & 1.8$^\circ$             \\ \cline{2-8} 
                               & Avg Time.                & \multicolumn{2}{c|}{21.51ms} & \multicolumn{2}{c|}{11.62ms} & \multicolumn{2}{c}{13.86ms}          \\ \hline
    \end{tabular}
    \vspace{-0.2cm}
\end{table}

Table \ref{tab:randdel} shows the comparison between the random deletion proposed in Sect. \ref{sect:branddel} (\textbf{RandDel} in the table) and the other two deletion mechanisms, an approach without any deletion mechanism (\textit{NoDel} in the table), and the sliding window mechanism used in \cite{xu2020decentralized} (\textit{SldWin} in the table),  which deletes the first frame when swarm keyframes in $\mathcal{G}$ are more than $m_{max}$. 
In the comparison, $m_{max}$ is set to 100.
The \textbf{Parallel1} and \textbf{Parallel2} dataset generates 255 and 321 swarm keyframes, respectively.

In the table, all three approaches have similar relative measurement accuracy (measured using RE), showing the deletion mechanism has no effect on the relative state estimates.
Nevertheless, the \textit{NoDel} approach has the best global consistency (measured using ATE), \textbf{RandDel} is slightly worse than \textit{NoDel}, and the worst approach, \textit{SldWin}, is close to the level of VIO-only in Table \ref{tab:comp1}. 
Concluding, \textbf{RandDel} has the same optimization speed compared to \textit{SldWin} but is 50\% faster than \textit{NoDel}.

\subsection{Evaluation on Outdoor Dataset} 
\begin{figure}[t]

    \centering
    \settowidth\limage{\includegraphics[height=3cm]{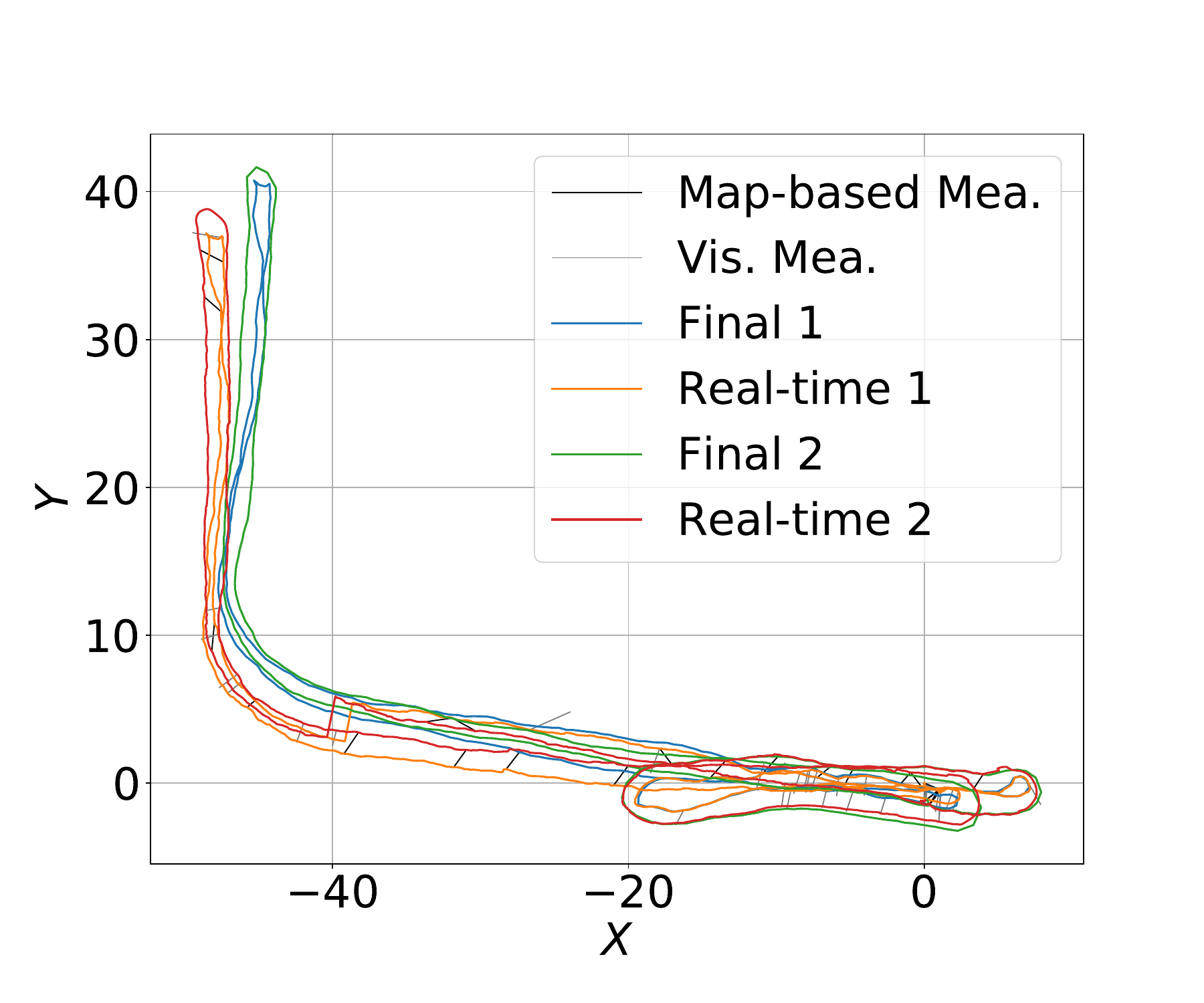}}
    \settowidth\mimage{\includegraphics[height=3cm]{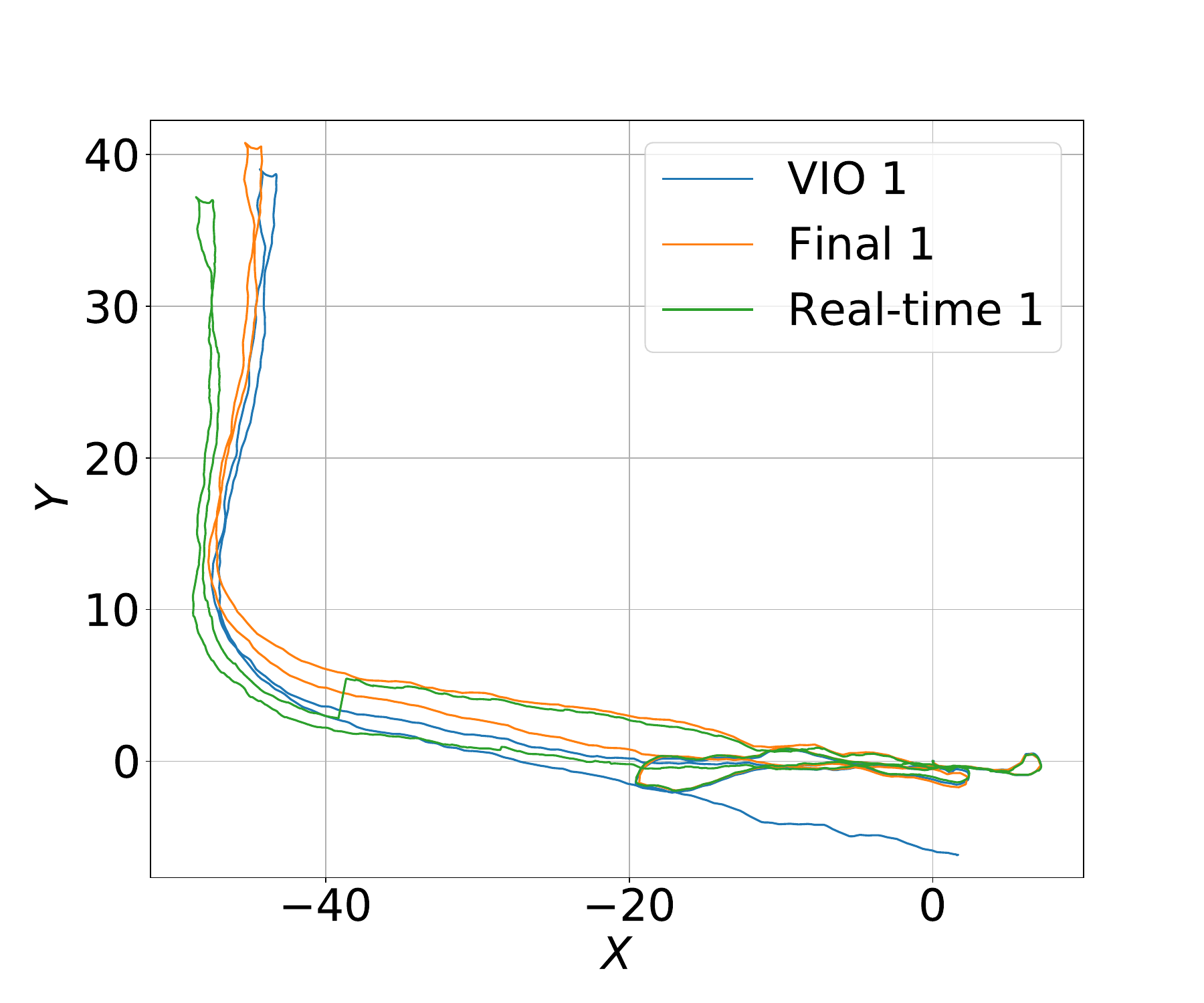}}
    \settowidth\rimage{\includegraphics[height=3cm]{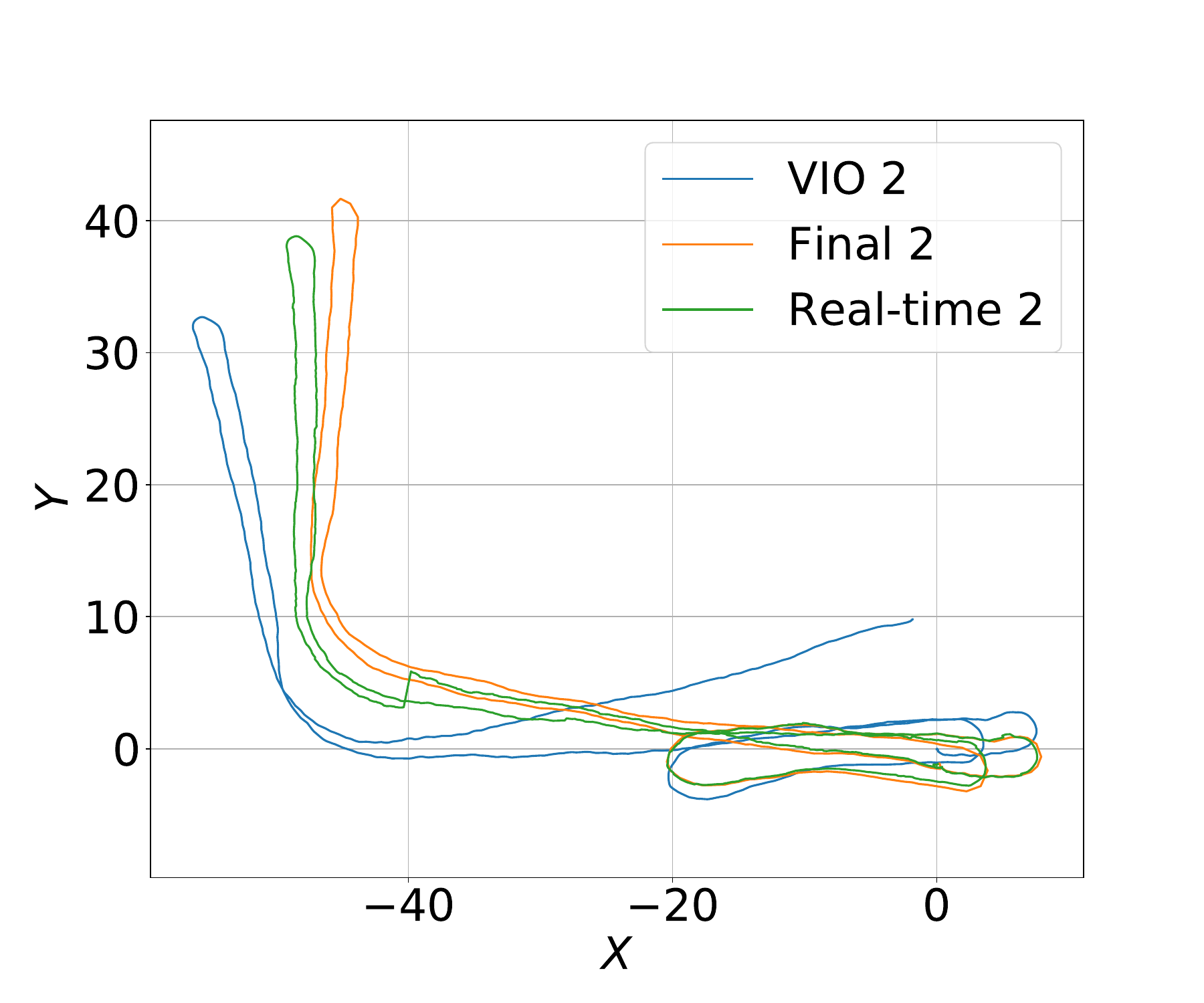}}
    \resizebox{1.0\linewidth}{!}{
        \begin{tabular}{p{\limage}p{\mimage}p{\rimage}}
    \includegraphics[height=3cm]{outdoor-fused2d}\newline
    \vspace{-0.5cm}
    \subcaption{}\label{fig:ex3_traj}
        &   \includegraphics[height=3cm]{outdoor-fusedvsgt2d_1}\newline
    \vspace{-0.5cm}
    \subcaption{}\label{fig:ex3_com1}
    &   \includegraphics[height=3cm]{outdoor-fusedvsgt2d_2}\newline
    \vspace{-0.5cm}
    \subcaption{}\label{fig:ex3_com2}
    \end{tabular}
    }
    \vspace{-0.4cm}
    \caption{\small{The estimated trajectories and the VIO trajectories of the two drones in an outdoor dataset. 
    For better visualization of the global consistency, the real-time estimated trajectories and final estimated trajectories are shown in the figures. }
    (a) The estimated trajectories of two drones with the map-based measurements and visual detection measurements. 
    The black lines are the map-based measurements. The gray lines are the detection measurements. 
    (b)-(c) The detailed comparison of the estimated trajectories (blue) and the VIO trajectories (orange).
    }\label{fig:outdoor}
    \vspace{-0.6cm}
\end{figure}

We also evaluate Omni-swarm on an outdoor dataset to verify that Omni-swarm is flexible in many scenarios.
Fig. \ref{fig:outdoor} shows the estimation trajectories on a double-drone outdoor dataset. 
The real-time trajectory in the figure is the output of state estimation in real-time with forward propagation (Eq. \ref{eq:forward_pro}).
The Final trajectory is the final estimate poses of keyframes in the  $\mathcal{G}$ at the end of the task.

As a result of a 235-meter-run in the outdoor environment, the estimated states drift only 1.9 m from the start point, which is 0.8\% of the trajectory length, showing that the global consistency is ensured. 
As a comparison, the original VIO trajectories' drifts averaged 6.35 m from the start point, which is 3.3 times bigger compared to our proposed method.
The real-time trajectories in the figure have a noticeable jump, while the final trajectories do not. This is because Omni-swarm eliminates the drifting error of VIO at this instant after detecting a map-based measurement, so the real-time estimation jumps to a more global consistency estimate. 
On the other hand, the final trajectory is smoother without this jump because it is a whole trajectory estimated result.

\subsection{Inter-drone collision avoidance experiments}\label{subsec:formation}
\begin{figure*}[ht!]
    \centering
    \settowidth\aimage{\includegraphics[height=4cm]{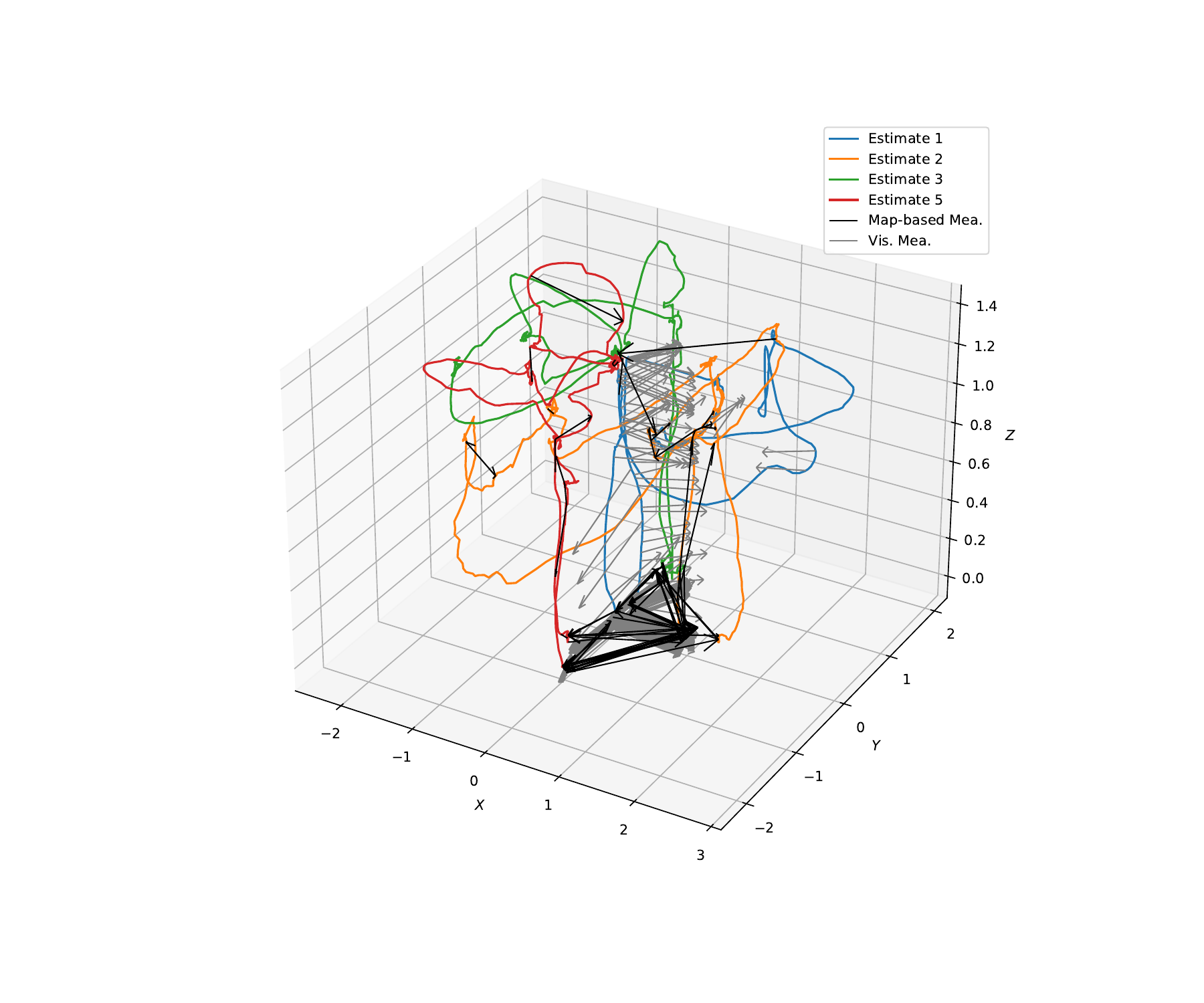}}
    \settowidth\bimage{\includegraphics[height=4cm]{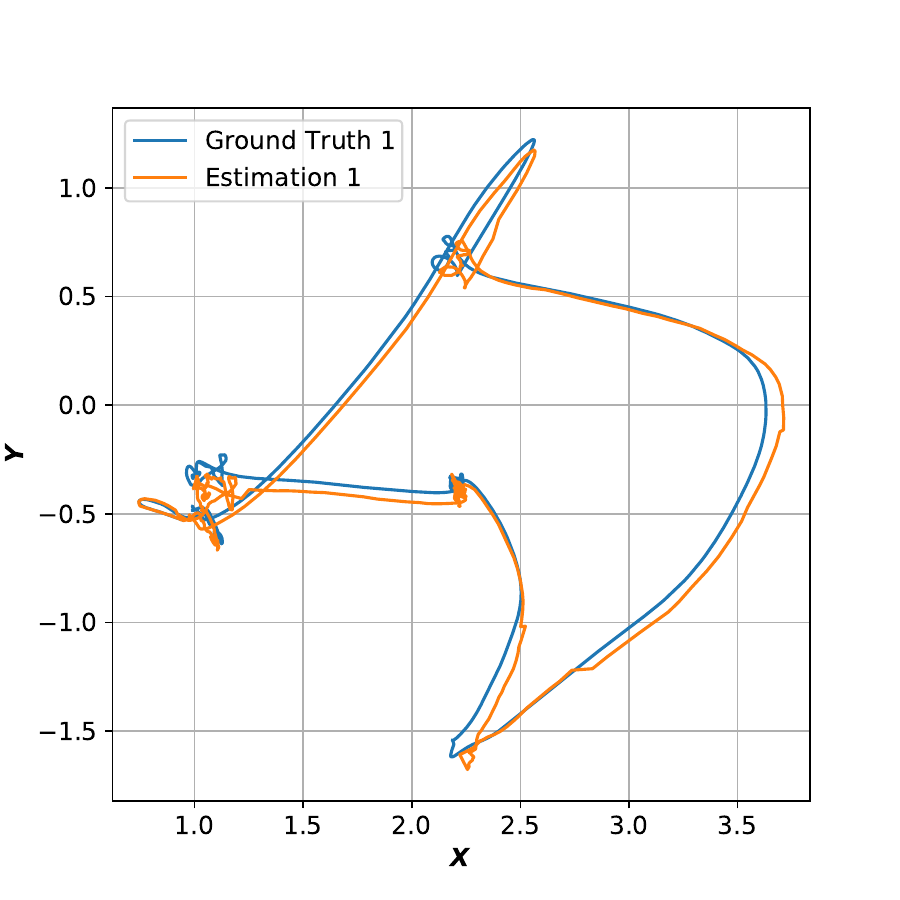}}
    \settowidth\cimage{\includegraphics[height=4cm]{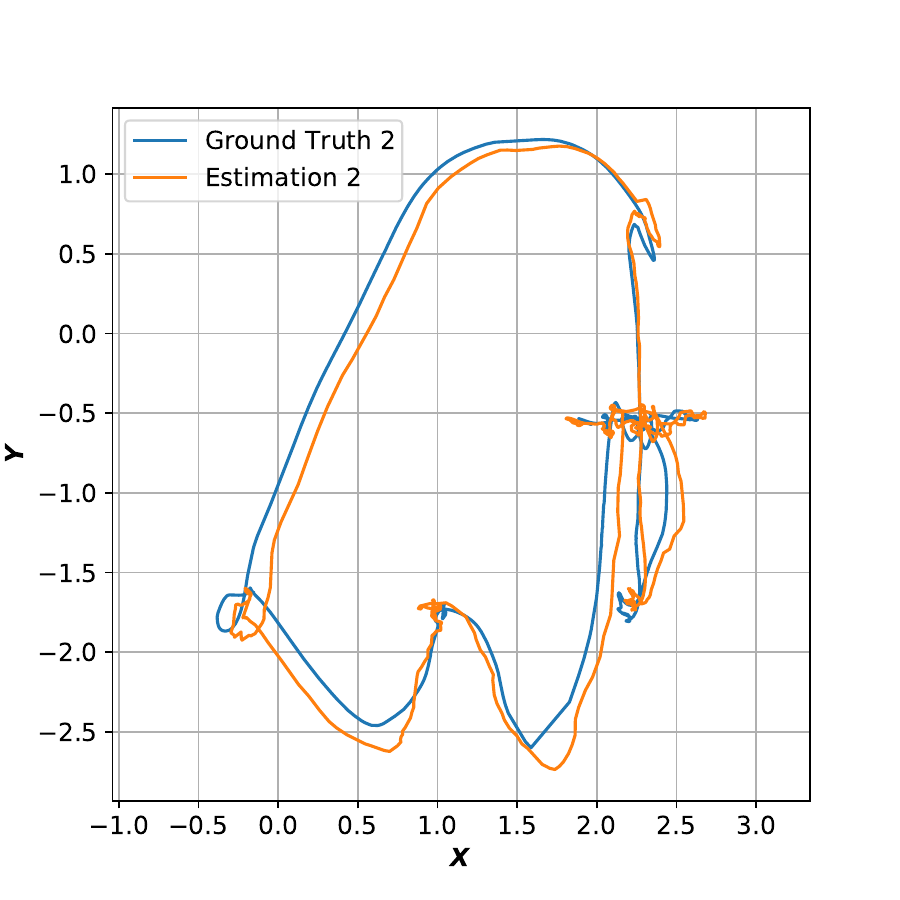}}
    \settowidth\dimage{\includegraphics[height=4cm]{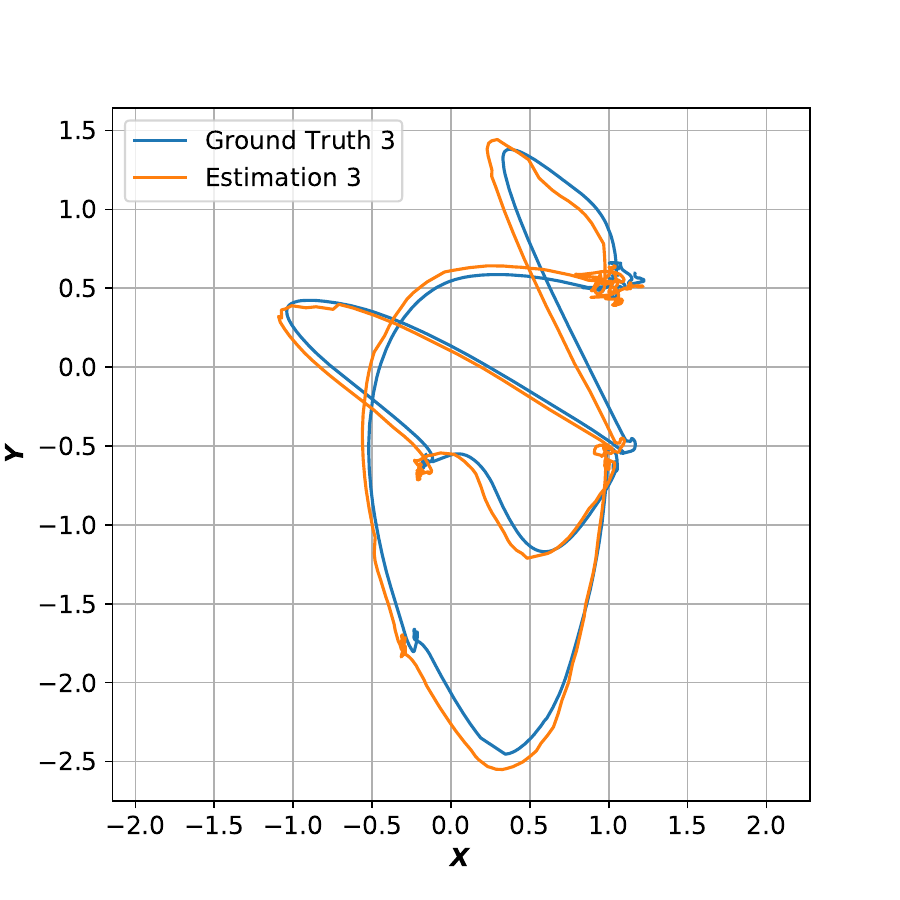}}
    \settowidth\eimage{\includegraphics[height=4cm]{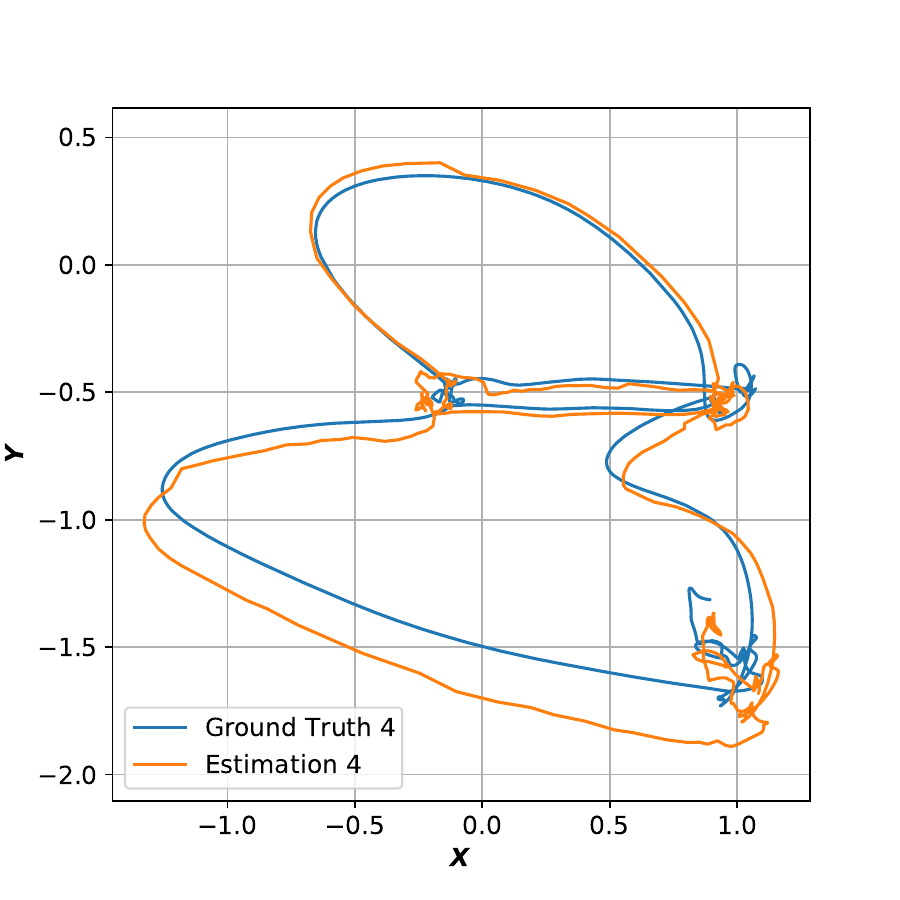}}
    \resizebox{0.9\textwidth}{!}{
        \begin{tabular}{p{\aimage}p{\bimage}p{\cimage}p{\dimage}p{\eimage}}
    \includegraphics[height=4cm]{ob-4-Traj2}\newline
    \vspace{-0.5cm}
    \subcaption{}\label{fig:ex5_traj}
        &   \includegraphics[height=4cm]{ob-4-fusedvsgt2d_1}\newline
    \vspace{-0.5cm}
    \subcaption{}\label{fig:ex5_com1}
    &   \includegraphics[height=4cm]{ob-4-fusedvsgt2d_2}\newline
    \vspace{-0.5cm}
    \subcaption{}\label{fig:ex5_com2}
    &   \includegraphics[height=4cm]{ob-4-fusedvsgt2d_3}\newline
    \vspace{-0.5cm}
    \subcaption{}\label{fig:ex5_com3}
    &   \includegraphics[height=4cm]{ob-4-fusedvsgt2d_5}\newline
    \vspace{-0.5cm}
    \subcaption{}\label{fig:ex5_com4}
\end{tabular}
    }
    \vspace{-0.4cm}

    \caption{
        \small{These figures show the estimated trajectories,and ground truth trajectories a four-drone inter-drone collision avoidance experiment. 
    (a): The estimated trajectories of the drones.
    The black lines are the map-based measurements. 
    The gray lines are the detection measurements. 
    Only part of the measurements is shown in this figure for better visualization.
    (b)-(e): The detailed comparison of the estimated trajectories (blue) and the ground truth trajectories (orange). 
    }
    }\label{fig:ex4}
    \vspace{-0.6cm}
\end{figure*}

% Please add the following required packages to your document preamble:
% \usepackage{multirow}
\begin{table}[]
    \centering
    \caption{\small{
    Accuracy comparison on three-drone and four-drone inter-drone collision avoidance experiments.
    Traj. Len. is the averaged trajectory length of the experiment.
    }}
    \label{tab:inter-collision}
    \resizebox{\linewidth}{!}{
    \begin{tabular}{cc|cccccc}
    \hline
    \multicolumn{2}{c|}{\multirow{2}{*}{Task}}                                                                 & \multicolumn{3}{c|}{\textbf{Proposed Method}}                                        & \multicolumn{3}{c}{VIO-Only}                                    \\ \cline{3-8} 
    \multicolumn{2}{c|}{}                                                                                      & \multicolumn{1}{c|}{x}     & \multicolumn{1}{c|}{y}     & \multicolumn{1}{c|}{z}     & \multicolumn{1}{c|}{x}     & \multicolumn{1}{c|}{y}     & z     \\ \hline\hline
    \multicolumn{1}{c|}{\multirow{5}{*}{\begin{tabular}[c]{@{}c@{}}Three\\ Drone\end{tabular}}} & Traj. Len.   & \multicolumn{6}{c}{27.8}                                                                                                                               \\ \cline{2-8} 
    \multicolumn{1}{c|}{}                                                                       & $RE_{x,y,z}$ & \multicolumn{1}{c|}{0.076} & \multicolumn{1}{c|}{0.074} & \multicolumn{1}{c|}{0.083} & \multicolumn{1}{c|}{0.064} & \multicolumn{1}{c|}{0.061} & 0.030 \\ \cline{2-8} 
    \multicolumn{1}{c|}{}                                                                       & $RE_{rot}$   & \multicolumn{3}{c|}{1.20°}                                                           & \multicolumn{3}{c}{0.8°}                                        \\ \cline{2-8} 
    \multicolumn{1}{c|}{}                                                                       & $ATE_{pos}$  & \multicolumn{3}{c|}{0.103}                                                           & \multicolumn{3}{c}{0.078}                                       \\ \cline{2-8} 
    \multicolumn{1}{c|}{}                                                                       & $ATE_{rot}$  & \multicolumn{3}{c|}{0.427°}                                                          & \multicolumn{3}{c}{0.096°}                                      \\ \hline
    \multicolumn{1}{c|}{\multirow{5}{*}{\begin{tabular}[c]{@{}c@{}}Four\\ Drone\end{tabular}}}  & Traj. Len.   & \multicolumn{6}{c}{19.9}                                                                                                                               \\ \cline{2-8} 
    \multicolumn{1}{c|}{}                                                                       & $RE_{x,y,z}$ & \multicolumn{1}{c|}{0.065} & \multicolumn{1}{c|}{0.056} & \multicolumn{1}{c|}{0.088} & \multicolumn{1}{c|}{0.070} & \multicolumn{1}{c|}{0.131} & 0.064 \\ \cline{2-8} 
    \multicolumn{1}{c|}{}                                                                       & $RE_{rot}$   & \multicolumn{3}{c|}{1.46°}                                                           & \multicolumn{3}{c}{1.2°}                                        \\ \cline{2-8} 
    \multicolumn{1}{c|}{}                                                                       & $ATE_{pos}$    & \multicolumn{3}{c|}{0.103}                                                           & \multicolumn{3}{c}{0.091}                                       \\ \cline{2-8} 
    \multicolumn{1}{c|}{}                                                                       & $ATE_{rot}$    & \multicolumn{3}{c|}{0.399°}                                                          & \multicolumn{3}{c}{0.160°}                                      \\ \hline
    \end{tabular}}
    \vspace{-0.5cm}
\end{table}

In the inter-drone collision avoidance experiments, each drone performs independent swarm state estimation in real-time and uses the estimated results for inter-drone collision avoidance and planning, which follows Sect. \ref{sect:planning}.
Table \ref{tab:inter-collision} shows the accuracy of real-time estimated states in the inter-drone collision avoidance experiments, and Fig. \ref{fig:ex4} shows the estimated trajectories of a four-drone inter-drone collision avoidance experiment.
This experiment successfully validates the practical value of Omni-swarm in a real-world environment.
One may notice that VIO has better performance on some data. 
This is because in short flight tasks (19-30 m as listed in the table), VIO drifts less and therefore has higher accuracy.  
However, our proposed method outperforms VIO on all datasets in the previous comparison, where drones have longer trajectories.
Additionally, VIO-only requires the aerial swarm to be initialized using ground truth or in using predefined starting points, and its error increases with flight range, limiting the practical value of this approach.

\subsection{Computation Time}\label{sect:timing}

Table \ref{tab:timing_expr} shows the  computation time of the main components of Omni-swarm in experiments.
The first P\&P scenario is the four drone plug-and-play experiment in Sect. \ref{sect:plugandplay}.
The second C\&3 scenario is the three drone inter-drone collision avoidance experiment in Sect. \ref{subsec:formation}.
The C\&3 scenario's back-end computation time is much longer than that of P\&P because it is a more complex task.
The table shows that Omni-swarm is able to achieve real-time performance with the computational frequency shown in Table \ref{tab:freq}.

\begin{table}[h!]
    \centering
    \caption{\small{The average computation time on each component in real-world experiments on the onboard computer.
     The units in this table are milliseconds.
     The abbreviations VINSF, VINSB, Det, Trk. LoopDet., and Opti. are the same as in Table \ref{tab:freq}.
    PCM is the PCM outlier rejection.
  }}
    \label{tab:timing_expr}
    \setlength{\tabcolsep}{6pt}
    \resizebox{\linewidth}{!}{
    \begin{tabular}{c|c|c|c|c|l|c|c|c|c}
    \hline
    Scenario            & VINSF                                                     & VINSB                                                  & Det \& Trk                                                  & PoseEsti.                                                & Desc.                                                       & \begin{tabular}[c]{@{}c@{}}Loop\\ Det.\end{tabular}      & Init.                  & PCM                    & Opti.               \\ \hline
    P\&P                & 12.1                                                      & 43.8                                                   & 32.8                                                        & 24.1                                                     &          273.7                                              & 48.9                                                     &   8.6                  & 0.8                    & 6.2                 \\ \hline
    C\&3                & 19.5                                                      & 82.1                                                   & 41.4                                                        & 16.6                                                     &          262.6                                              & 28.1                                                     &   23.7                 & 110.5                  & 155.5               \\ \hline
    \end{tabular}
    }
\vspace{-0.4cm}
\end{table}

\section{Conclusion and Future Work}\label{sect:con}
This paper introduced Omni-swarm, a decentralized omnidirectional visual-inertial-UWB state estimation system for aerial swarms.
Compared to previous works, the proposed approach addresses the complicated initialization issue, observability issue caused by a restricted FoV, and global consistency issue.
To demonstrate the feasibility and effectiveness of Omni-swarm, we tested it by extensive aerial swarm flight experiments.
Compared with ground truth data from the motion capture system, the state estimation results reach centimeter-level accuracy on relative estimation while ensuring global consistency.
With Omni-swarm, formation flights in various complex environments are no longer impossible.
Inter-drone collision avoidance is successfully demonstrated based on Omni-swarm.
We believe that Omni-swarm can be widely adopted in a variety of scenarios and on multiple scales.

However, our system is also inevitably characterized by a number of shortcomings that need to be improved in future work, including:
1) The dependence on camera intrinsic and extrinsic calibration, which is a common problem for all visual SLAM systems. In the future, we will try to develop online fault detection and calibration to further improve the robustness.
2) Scalability. Our current back-end algorithm is $O(n^2)$ for the growth of the scale of the swarm, which makes it difficult to apply Omni-swarm to larger-scale swarms (more than 100 drones).
In the future, we will try to extend Omni-swarm to large aerial swarms.
3) Communication range. 
Current work is focused on state estimation, and the network setup currently limits our aerial swarm to work within a short distance (22.4 m) from each other.
In the future, we will extend the communication distance by deploying routing algorithms, e.g., Batman-adv\cite{seither2011routing}, and upgrading communication devices.

In the future, we will deploy Omni-swarm to real-world applications and also use it to build the dense global map to exploit the advantages of Omni-swarm fully.

\newlength{\bibitemsep}\setlength{\bibitemsep}{.03\baselineskip}
\newlength{\bibparskip}\setlength{\bibparskip}{0pt}
\let\oldthebibliography\thebibliography
\renewcommand\thebibliography[1]{%
 \oldthebibliography{#1}%
 \setlength{\parskip}{\bibitemsep}%
 \setlength{\itemsep}{\bibparskip}%
}
\bibliographystyle{IEEEtran}
\bibliography{hao2021.bib} 

\begin{IEEEbiography}[{\includegraphics[width=1in,height=1in,clip,keepaspectratio]{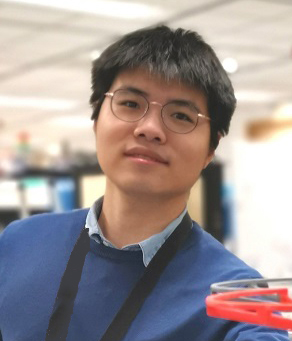}}]{Hao Xu}
    received the B.Sc. degree in Physics from the University of Science and Technology of China, Hefei, China, in 2016. He is currently working toward the
    Ph.D. degree with the Hong Kong University of Science and Technology, Hong Kong, under the supervision of Prof. Shaojie Shen.
    His research interests include unmanned aerial vehicles, aerial swarm, state estimation, sensor fusion, localization and mapping.\end{IEEEbiography}
\vspace{-1.5cm}
\begin{IEEEbiography}[{\includegraphics[width=1in,height=1in,clip,keepaspectratio]{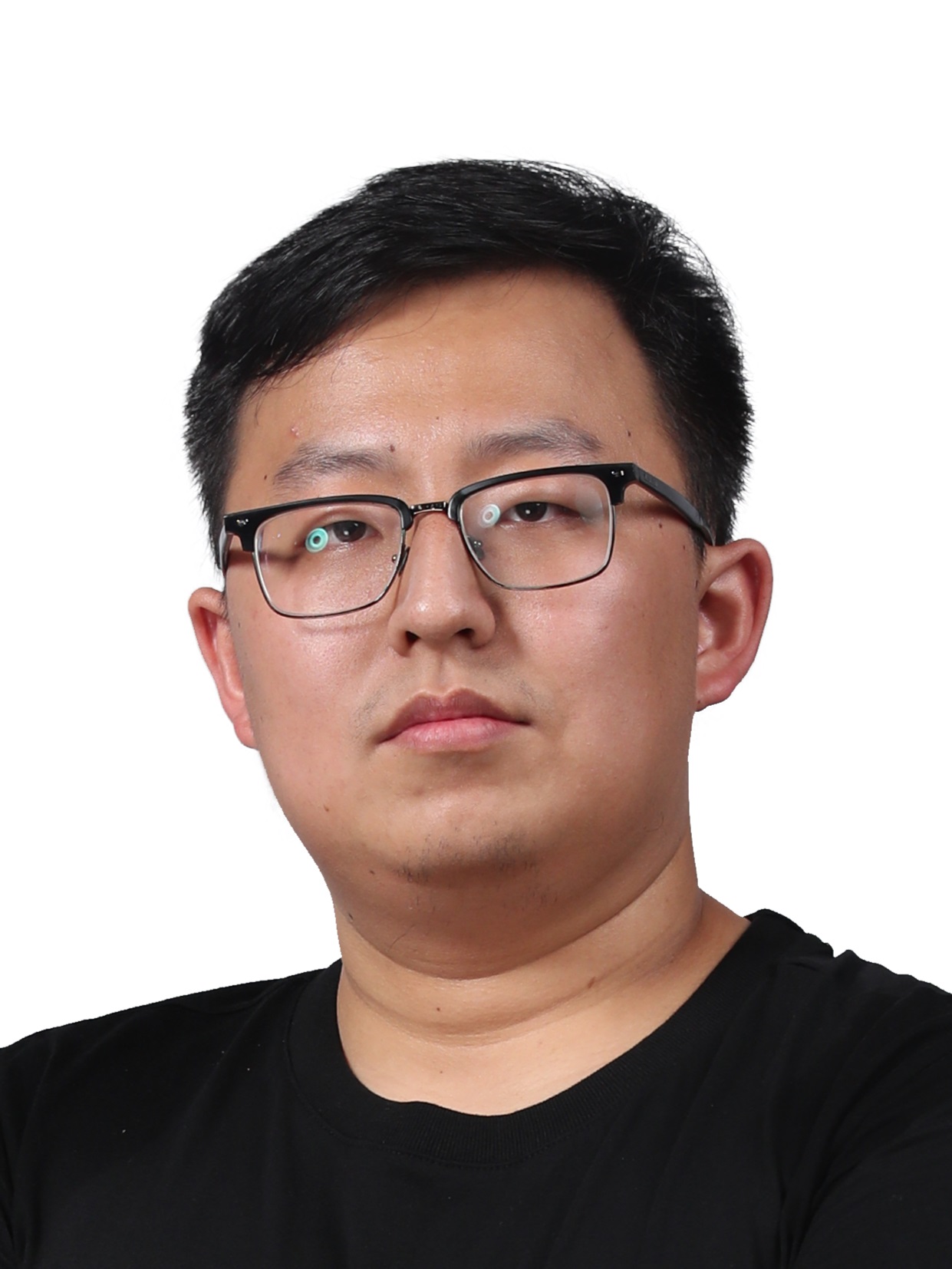}}]{Yichen Zhang}
    received the B.Eng. degree in computer engineering and B.B.A in general business management in 2020 from the Hong Kong University of Science and Technology, Hong Kong, where he is currently working toward the Ph.D. degree in electronic and computer engineering under the supervision of Prof. S. Shen.
    His research interests include motion planning, dense mapping and active SLAM for autonomous robots.\end{IEEEbiography}
\vspace{-1.5cm}

\begin{IEEEbiography}[{\includegraphics[width=0.75in,height=1in,clip,keepaspectratio]{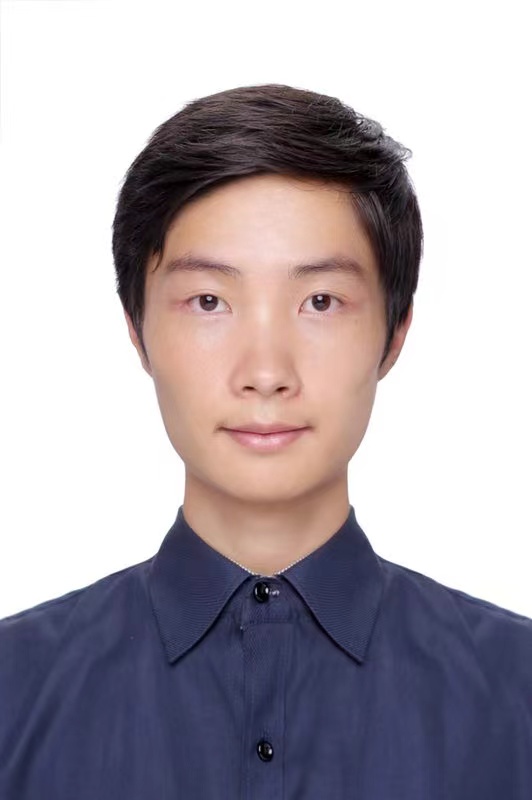}}]{Boyu Zhou}
        received the B.Eng. degree in mechanical engineering from Shanghai Jiao Tong University, Shanghai, China, in 2018. He is currently working toward the Ph.D. degree in electronic and computer engineering with the Hong Kong University of Science and Technology, Hong Kong. 
        His research interests include aerial robots, autonomous navigation, motion planning, dense mapping, exploration and swarm.\end{IEEEbiography}
\vspace{-1.5cm}

\begin{IEEEbiography}[{\includegraphics[width=0.75in,height=1in,clip,keepaspectratio]{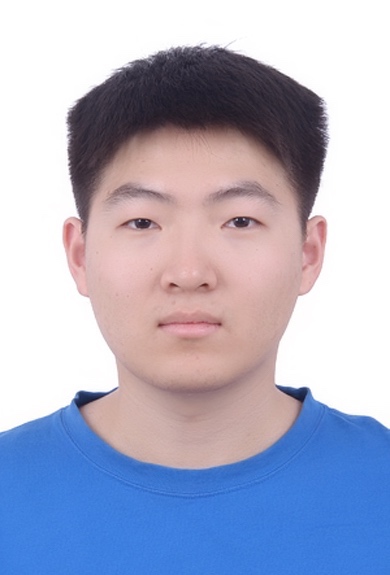}}]{Luqi Wang}
    received the B.Eng. degree in computer engineering and aerospace engineering in 2018 from the Hong Kong University of Science and  Technology, Hong Kong, where he is currently working toward the Ph.D. degree in electronic and computer engineering.
    His research interests include control, navigation, and path planning for autonomous robots.
    \end{IEEEbiography}
    \vspace{-1.5cm}

\begin{IEEEbiography}[{\includegraphics[width=0.75in,height=1in,clip,keepaspectratio]{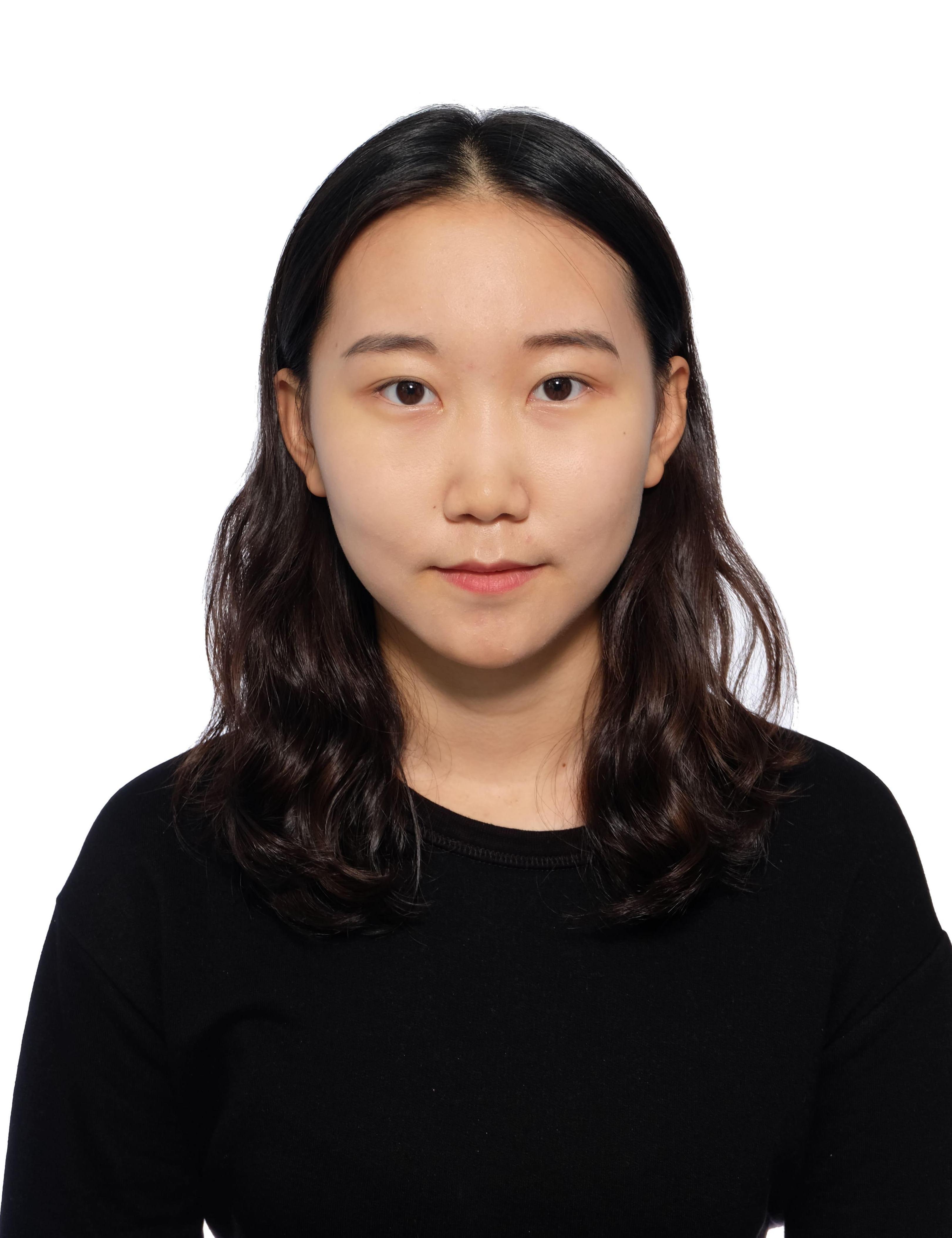}}]{Xinjie Yao}
    received her B.Eng in Computer Engineering from the Hong Kong University of Science and Technology in 2020. She received her M.S. in Robotics in 2022 from Carnegie Mellon University. Her research interests are in robotics, focusing on navigation, autonomy, and human-robot interaction.\end{IEEEbiography}
\vspace{-1.5cm}

\begin{IEEEbiography}[{\includegraphics[width=0.75in,height=1in,clip,keepaspectratio]{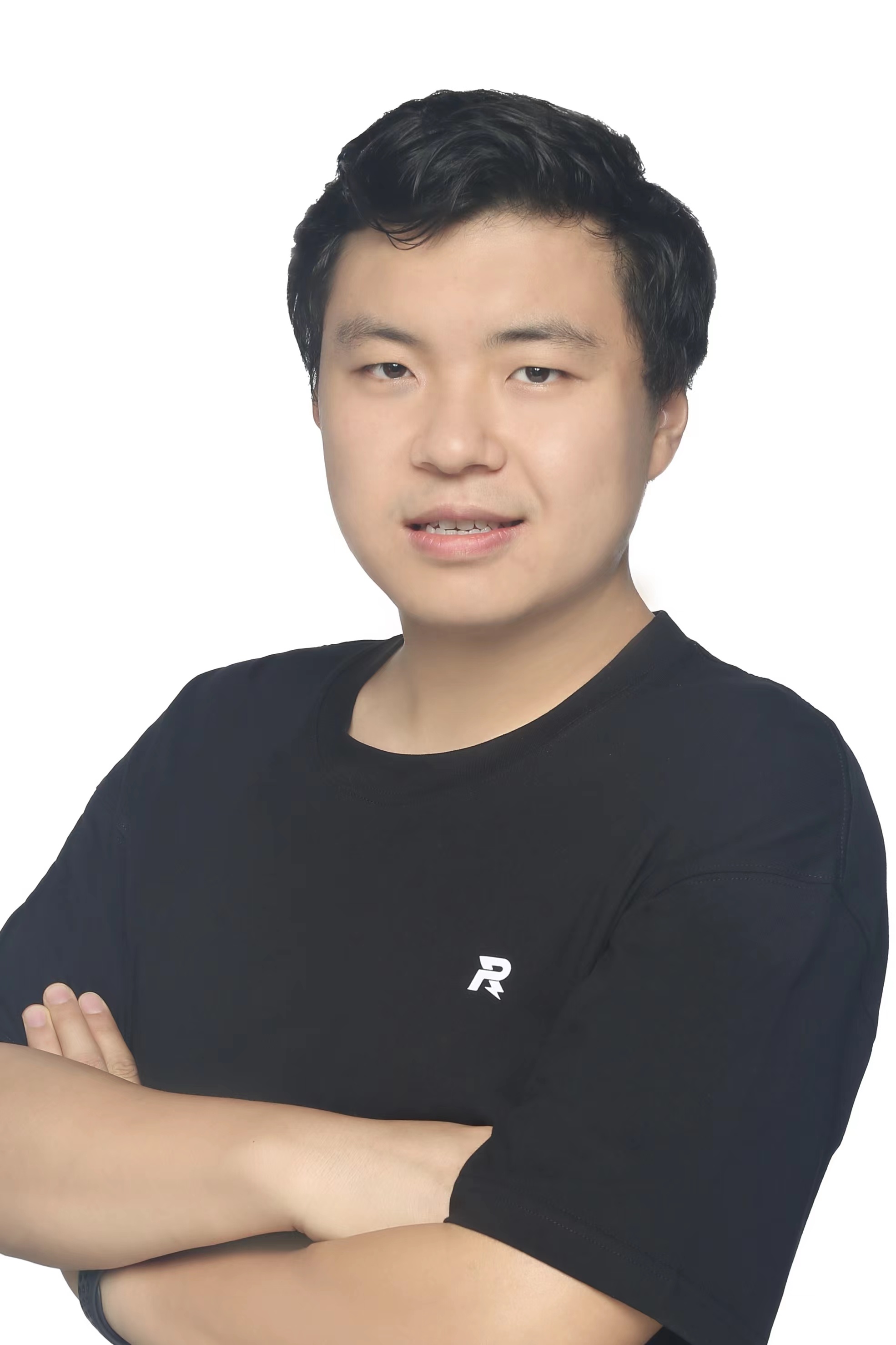}}]{Guotao Meng}
    received his B.Eng. degree in Automation from the School of Electronic and Information Engineering of Xi'an Jiaotong University in 2018. He is working towards his Ph.D. degree at the Visual Intelligence Laboratory, the Department of Electronic and Computer Engineering, the Hong Kong University of Science and Technology. His research interests include image processing, deep learning, computer vision and robotics.\end{IEEEbiography}
\vspace{-1cm}

\begin{IEEEbiography}[{\includegraphics[width=0.75in,height=1in,clip,keepaspectratio]{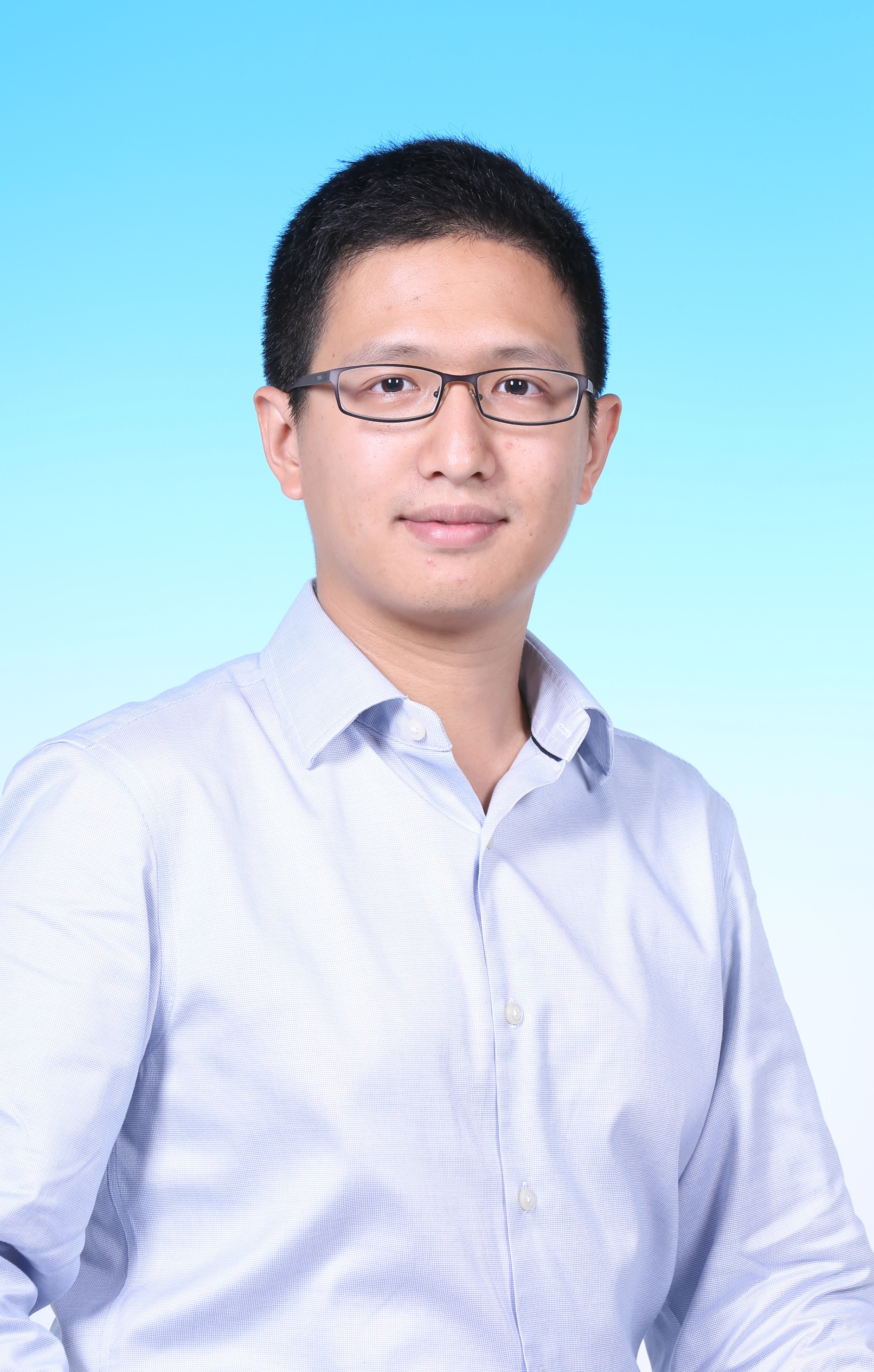}}]{Shaojie Shen} received the B.Eng. degree in electronic
    engineering from the Hong Kong University of Science and Technology, Hong Kong, in 2009, and the
    M.S. degree in robotics and the Ph.D. degree in electrical and systems engineering from the University
    of Pennsylvania, Philadelphia, PA, USA, in 2011 and
    2014, respectively.

    He was with the Department of Electronic and
    Computer Engineering, Hong Kong University of
    Science and Technology in September 2014 as an
    Assistant Professor, and was promoted to an Associate Professor in 2020. His research interests include the areas of robotics
    and unmanned aerial vehicles, with focus on state estimation, sensor fusion,
    computer vision, localization and mapping, and autonomous navigation in
    complex environments. \end{IEEEbiography}

\end{document}